\documentclass[11pt]{article}

\usepackage[margin=1.05in]{geometry}
\usepackage{amsmath,amssymb,amsthm,mathtools}
\usepackage{microtype}
\usepackage{enumitem}
\usepackage{float}
\floatstyle{ruled}
\newfloat{algorithm}{t}{loa}
\floatname{algorithm}{Algorithm}
\usepackage{xcolor}
\usepackage{tikz}
\usetikzlibrary{arrows.meta,positioning,calc,decorations.pathreplacing}
\usepackage{natbib}
\usepackage[colorlinks=true,linkcolor=blue,citecolor=blue,urlcolor=blue]{hyperref}
\hypersetup{
  pdftitle={Calibrated Multiaccuracy: Deterministic and Randomized Sample Complexity},
  pdfauthor={Jiuyao Lu, Georgy Noarov, and Aaron Roth}
}

\newtheorem{theorem}{Theorem}[section]
\newtheorem{lemma}[theorem]{Lemma}
\newtheorem{proposition}[theorem]{Proposition}
\newtheorem{corollary}[theorem]{Corollary}

\newcommand{\E}{\mathbb E}
\newcommand{\Pp}{\mathbb P}
\newcommand{\R}{\mathbb R}
\newcommand{\one}{\mathbf 1}
\newcommand{\eps}{\varepsilon}
\newcommand{\calA}{\mathcal A}
\newcommand{\calF}{\mathcal F}
\newcommand{\calG}{\mathcal G}
\newcommand{\calH}{\mathcal H}

\newcommand{\calX}{\mathcal X}
\newcommand{\MA}{\operatorname{MA}}
\newcommand{\ECE}{\operatorname{ECE}}
\newcommand{\CMA}{\operatorname{CMA}}

\newcommand{\Cov}{\operatorname{Cov}}

\newcommand{\Ber}{\operatorname{Ber}}
\newcommand{\Unif}{\operatorname{Unif}}
\newcommand{\poly}{\operatorname{poly}}
\newcommand{\sgn}{\operatorname{sign}}
\newcommand{\dist}{\operatorname{dist}}

\title{Calibrated Multiaccuracy:\\
Deterministic and Randomized Sample Complexity}
\author{
  Jiuyao Lu\\
  {\small University of Pennsylvania}
  \and
  Georgy Noarov\\
  {\small Institute for Advanced Study}
  \and
  Aaron Roth\\
  {\small Resolution}\\
  {\small University of Pennsylvania}
}
\date{\today}

\begin{document}
\hypersetup{pageanchor=false}
\pagenumbering{gobble}
\maketitle

\begin{abstract}

Calibrated multiaccuracy asks that a predictor be globally calibrated and
multiaccurate with respect to a bounded test family $\calG$; that is, its
residuals should be uncorrelated with every test $g(x)\in\calG$. It has emerged
as a promising weakening of multicalibration that nonetheless is sufficient
for many applications. However, the sample complexity of learning a predictor
satisfying calibrated multiaccuracy has remained poorly understood --- and in
fact was not known to be less than the sample complexity needed for full
multicalibration, which is
$\widetilde\Theta\left(\frac{\log|\calG|}{\eps^2} + \frac{1}{\eps^{3}}\right)$
when $|\calG|$ is at least a sufficiently large power of $\log(1/\eps)$.
We resolve the sample complexity of
$\eps$-calibrated multiaccuracy, showing that it scales as
\[
\widetilde\Theta\left(\frac{\log|\calG|}{\eps^2} + \frac{1}{\eps^{5/2}}\right),
\]
separating it from the complexity of multicalibration. The upper bound is obtained by an algorithm that outputs a randomized predictor.

We also determine the sample complexity for calibrated multiaccuracy when the learning algorithm is restricted to outputting a deterministic predictor.  The rate is
\[
  \widetilde\Theta\left(
    \frac{\log|\calG|}{\eps^2}
    +
    \frac{(\log|\calG|)^{1/3}}{\eps^{8/3}}
  \right)
\]
whenever $\log|\calG|=\Omega(\log(1/\eps))$.  When 
$\log|\calG|\le\eps^{-1+\kappa}$ for any fixed $\kappa\in(0,1)$, this rate is
polynomially larger than the rate for randomized predictors. Thus, unlike for multicalibration, calibrated
multiaccuracy exhibits a genuine sample-complexity separation between
deterministic and randomized predictors.  The matching upper and lower bounds
show that the unusual cube-root dependence on $\log|\calG|$ is intrinsic.  At
$\log|\calG|\asymp\eps^{-1}$, the minimax rates for randomized and
deterministic calibrated multiaccuracy both coincide with the minimax rate for
multicalibration.
\end{abstract}

\clearpage
\begingroup
\small
\tableofcontents
\endgroup
\clearpage
\pagenumbering{arabic}
\hypersetup{pageanchor=true}

\section{Introduction}

\paragraph{Calibration and multiaccuracy.} Calibration and multiaccuracy are both statistical consistency measures that we can ask of a model $p:\mathcal{X}\rightarrow [0,1]$ mapping features to predictions of bounded labels $Y\in[0,1]$.
Calibration asks that, among examples assigned prediction value $p$, the
average label should be close to $p$.  Multiaccuracy asks that for every test (or ``group'') function $g:\mathcal{X}\rightarrow [-1,1]$  in a prescribed
family, the residual $Y-p(X)$ should be nearly uncorrelated with $g(X)$
\citep{HebertJohnsonKimReingoldRothblum2018,KimGhorbaniZou2019}.  Multicalibration combines these two ideas in a
strong way, asking for calibration not just overall but also simultaneously
within the feature subsets or test-weighted populations specified by the group
family
\citep{HebertJohnsonKimReingoldRothblum2018}.

This paper studies an intermediate guarantee, weaker than multicalibration, obtained by imposing
\emph{global} calibration together with multiaccuracy.  We study both randomized and deterministic predictors. In both regimes, the
training algorithm may use randomness.  The important distinction is the
type of predictor it outputs.  A randomized predictor
$Q=(Q_x)_{x\in\calX}$ assigns to each feature value $x$ a distribution $Q_x$
over predictions in $[0,1]$.  On a fresh
example $(X,Y)$, let $P\sim Q_X$ be the realized prediction, drawn independently
of $Y$ conditional on $X$.  A deterministic predictor is instead a fixed
function $p:\calX\to[0,1]$, corresponding to the special case in which $Q_x$
is a point mass at $p(x)$, so that $P=p(X)$.  For a finite class
$\calG$ of tests $g:\calX\to[-1,1]$,  define the multiaccuracy (MA) error and expected calibration
error (ECE) in terms of the realized prediction:
\[
  \MA(Q;\calG)=
  \sup_{g\in\calG}\left|\E[g(X)(Y-P)]\right|,
  \qquad
  \ECE(Q)=\E\left|\E[Y-P\mid P]\right|.
\]
The calibrated multiaccuracy error is
\[
  \CMA(Q;\calG)=\max\{\MA(Q;\calG),\ECE(Q)\}.
\]

\paragraph{Why calibrated multiaccuracy is interesting.}
Over the past several years, multicalibration has been the subject of intense study in part because it has found so many different applications, from omniprediction \citep{GopalanKalaiReingoldSharanWieder2022} to conditional guarantees in conformal prediction \citep{bastani2022practical,jung2023batch} to complexity-theoretic constructions \citep{casacuberta2024complexity} to multiparty information aggregation \citep{collina2025tractable,CollinaGlobusHarrisGoelGuptaRothShi2025Collaborative,KearnsRothRyu2025Networked}.
A growing line of work, however, has shown that calibrated multiaccuracy is enough for
tasks where multicalibration had previously been the standard tool.  In
omniprediction, the outcome-indistinguishability viewpoint decomposes loss
minimization into calibration plus multiaccuracy for a loss-derived class of
tests~\citep{GopalanHuKimReingoldWieder2023}.  \citet{CasacubertaGopalanKanadeReingold2025} show that adding
global calibration to multiaccuracy recovers several implications previously
known from multicalibration, including 
optimal-density hardcore measures.  In computational complexity, the
multicalibration-based characterization of product-distribution
indistinguishability of \citet{MarcussenPuttermanVadhan2025Product} can also be
obtained from calibrated
multiaccuracy~\citep{DworkTankala2025Supersimulators}.

These results suggest that calibrated multiaccuracy is a useful relaxation of
multicalibration: strong enough for many applications, but structurally weaker. However, prior to this paper, it had not been established that calibrated multiaccuracy error could actually be driven to $0$ any faster than multicalibration error, as a function of the number of samples needed for learning.  The best known sample complexity upper bounds for calibrated multiaccuracy in
the regime $|\calG|=\poly(1/\eps)$ were
$\widetilde O(\eps^{-3})$~\citep{GibbsTibshirani2025,NoarovRoth2026Deterministic}, no better than
the minimax optimal rates for multicalibration~\citep{CollinaLuNoarovRoth2026}. The only non-trivial lower bound for calibrated multiaccuracy was the 
$\Omega(\eps^{-5/2})$ bound of
\citet{GibbsTibshirani2025}, which applied only when the output after training
was a deterministic predictor $p:\calX\to[0,1]$.  In particular, no
super-$\eps^{-2}$  lower bound was known for algorithms
outputting randomized predictors.

\paragraph{Randomized versus deterministic predictors.}
One might naturally expect allowing the output predictor to randomize not to matter
for a batch learning problem.  For many familiar distributional learning
objectives, randomization is not needed.  It is never needed for risk minimization of a single expected loss, for example.  The closely related multicalibration problem reinforces
this expectation: \citet{CollinaLuNoarovRoth2026} identified the minimax optimal
sample complexity, initially with upper bounds that were realized by randomized predictors. However subsequently, \citet{NoarovRoth2026Deterministic} showed that
deterministic predictors suffice to achieve the minimax optimal sample complexity.  Thus
one might reasonably expect the same for calibrated multiaccuracy --- that there should be no worst-case advantage to outputting a randomized predictor. Surprisingly (to us, at least), we show that the story is different for calibrated multiaccuracy: for polynomial-size test families, randomized predictors attain a strictly smaller minimax sample complexity!

\subsection{Our results}

Write $A=\max\{1,\log|\calG|\}$ for the group budget.  We prove the following, where the lower bounds take the worst case over distributions and bounded test
families of the specified cardinality.

\begin{enumerate}[leftmargin=2em,label=\textbf{\arabic*.}]
\item \emph{Tight randomized sample complexity.}
We give an efficient algorithm which, given $n$ samples and a  
test family $\calG$, outputs a
randomized predictor $Q$ satisfying, with probability $1-\delta$:
\[
  \CMA(Q;\calG)
  \le
  C\left\{
    \sqrt{\frac{\log(|\calG|/\delta)}{n}}
    +
    \left(\frac{\log(n/\delta)}{n}\right)^{2/5}
  \right\}.
\]
Equivalently, error $\eps$ is
achievable with
  $\widetilde O\left(\frac{A}{\eps^2}+\frac1{\eps^{5/2}}\right)$
samples.  We prove a matching lower bound for arbitrary randomized predictors that is tight for every value of $A$.
For polynomial-size test families, this rate is
$\widetilde\Theta(\eps^{-5/2})$, compared with
$\widetilde\Theta(\eps^{-3})$ for full multicalibration. 
The formal upper bound is 
Theorem~\ref{thm:main}; the
matching randomized lower bound is Theorem~\ref{thm:lower}.  
\item \emph{Tight deterministic sample complexity and a separation.}
For algorithms that must output a deterministic predictor $p$, we prove the
information-theoretic sample-complexity upper bound
\[
  \widetilde O\left(
    \frac{A}{\eps^2}
    +
    \frac{A^{1/3}}{\eps^{8/3}}
  \right).
\]
We also prove that, for
$\log(1/\eps)\lesssim A\lesssim\eps^{-1}$, some bounded test family with
$\log|\calG|=\Theta(A)$ requires
$\Omega(A^{1/3}\eps^{-8/3})$ samples from every algorithm whose output is a
deterministic predictor.  The randomized
$\Omega(A/\eps^2)$ multiaccuracy lower bound applies a fortiori to
deterministic predictors and supplies the matching term.
Thus the upper bound is tight up to logarithmic factors for every
$A=\Omega(\log(1/\eps))$.  In particular, the matching bounds show that the
unusual $A^{1/3}$ dependence is a genuine feature of deterministic prediction,
not an artifact of the upper-bound analysis.  Unlike the efficient algorithm
for randomized predictors in item~1, this is an information-theoretic
construction whose direct implementation can require exponential time.   The formal upper and lower bounds are Theorems~\ref{thm:det-upper}
and~\ref{thm:det-budget-lower}, respectively.

In the matching lower-bound range, this deterministic rate is polynomially
larger than the randomized rate whenever $A\le\eps^{-1+\kappa}$ for a fixed
$\kappa\in(0,1)$.  For example, at
$A=\Theta(\log(1/\eps))$, deterministic predictors require
$\widetilde\Theta(\eps^{-8/3})$ samples, while randomized predictors need
only $\widetilde\Theta(\eps^{-5/2})$.  At $A\asymp\eps^{-1}$, both
deterministic terms are of order $\eps^{-3}$, matching the deterministic
multicalibration rate of \citet{NoarovRoth2026Deterministic}.
\end{enumerate}

\begin{figure}[H]
\centering
\begin{tikzpicture}[x=9.8cm,y=8.0cm,>=Stealth,font=\small]
  \def\ybase{2.38}
  \draw[->,thick] (0,0) -- (1.08,0)
    node[below right,font=\scriptsize] {$\beta$ in $A=\eps^{-\beta}$};
  \draw[->,thick] (0,0) -- (0,0.82)
    node[above,font=\scriptsize] {sample exponent};

  \foreach \x/\lab in {0/{0},0.5/{1/2},1/{1}} {
    \draw[gray!45] (\x,0) -- (\x,0.66);
    \draw (\x,0.006) -- (\x,-0.006) node[below,font=\scriptsize] {$\lab$};
  }
  \foreach \yy/\lab in {2.5/{5/2},2.6667/{8/3},2.8333/{17/6},3/{3}} {
    \pgfmathsetmacro{\py}{\yy-\ybase}
    \draw[gray!30] (0,\py) -- (1.03,\py);
    \draw (0.006,\py) -- (-0.006,\py) node[left,font=\scriptsize] {$\lab$};
  }

  \draw[blue!70!black,line width=1.2pt] (0,2.5-\ybase) -- (0.5,2.5-\ybase);
  \draw[blue!70!black,line width=1.2pt] (0.5,2.5-\ybase) -- (1,3-\ybase);
  \node[blue!70!black,anchor=north west,font=\scriptsize,align=left,inner sep=1pt]
    at (0.55,2.52-\ybase) {randomized tight rate\\$\max\{5/2,2+\beta\}$};

  \draw[teal!70!black,line width=1.2pt] (0,3-\ybase) -- (1,3-\ybase);
  \draw[teal!70!black,line width=1pt,fill=white]
    (0,3-\ybase) circle (1.5pt);
  \node[teal!70!black,anchor=south west,font=\scriptsize,align=left,
        inner sep=1pt]
    at (0.08,3.005-\ybase) {multicalibration tight rate $3$};

  \draw[red!75!black,line width=1.2pt] (0,2.6667-\ybase) -- (1,3-\ybase);
  \draw[red!75!black,line width=1pt,fill=white]
    (0,2.6667-\ybase) circle (1.5pt);
  \node[red!70!black,anchor=south west,font=\scriptsize,align=left,inner sep=1pt]
    at (0.07,2.85-\ybase) {deterministic tight rate\\$8/3+\beta/3$};

  \fill[black] (1,3-\ybase) circle (1.2pt);
  \node[black,anchor=south east,font=\scriptsize,align=right,inner sep=1pt]
    at (0.98,3.07-\ybase) {all three meet\\at $A\asymp\eps^{-1}$};
  \draw[gray!65,thin] (0.98,3.055-\ybase) -- (1,3.01-\ybase);

  \draw[decorate,decoration={brace,amplitude=4pt},gray!65]
    (0.02,2.45-\ybase) -- (0.48,2.45-\ybase)
    node[midway,below=5pt,font=\scriptsize,gray!60!black] {small test budget};
\end{tikzpicture}
\caption{The group-budget rate landscape, with $A=\eps^{-\beta}$ and vertical
axis the sample-complexity exponent of $1/\eps$, suppressing logarithmic
factors.  Blue and red give the tight randomized and deterministic calibrated
multiaccuracy rates; teal gives the tight multicalibration rate.  The open
endpoints represent subpolynomial budgets.}
\label{fig:rate-landscape}
\end{figure}
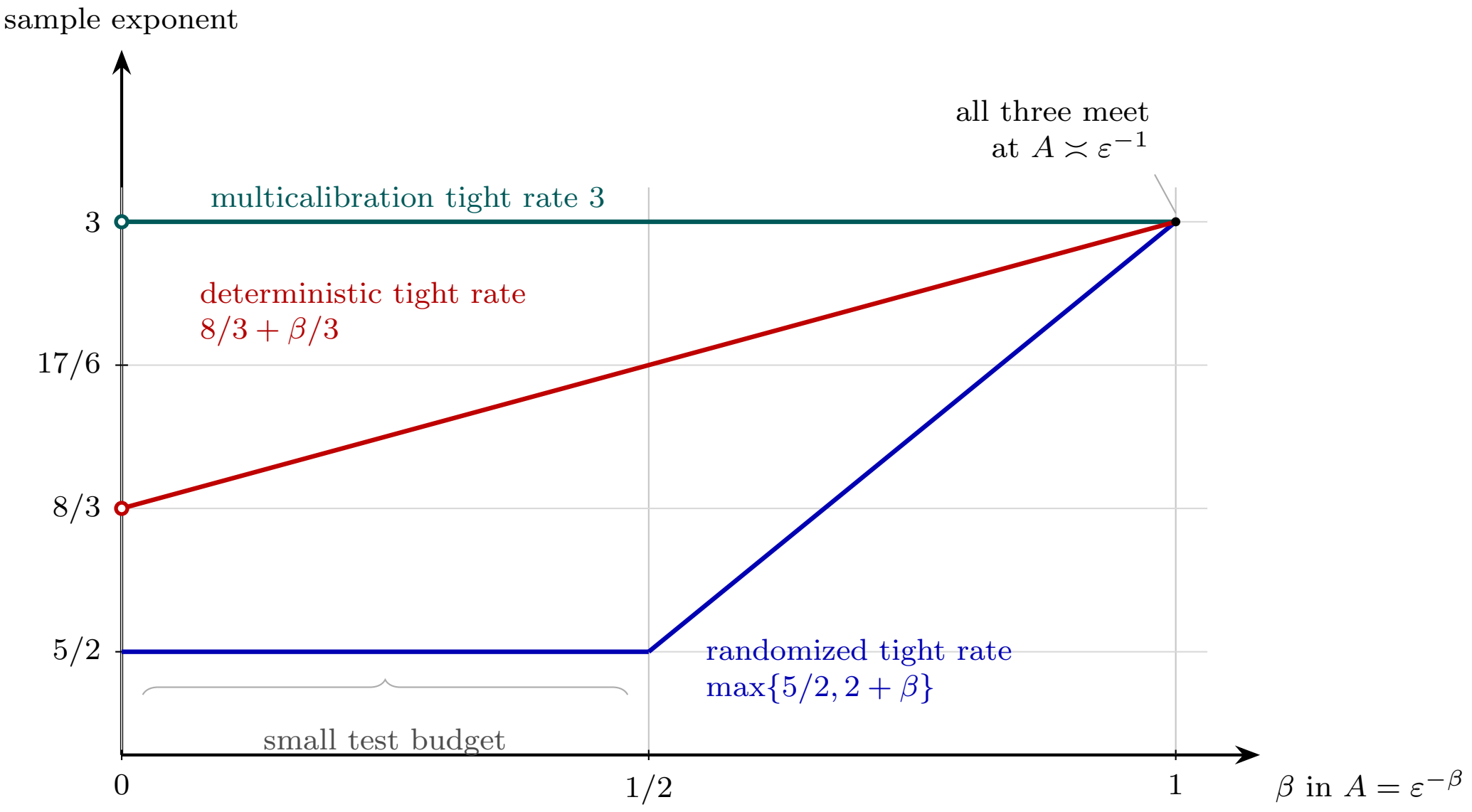

It is useful to visualize the group-budget dependence by writing
$A=\eps^{-\beta}$.  Figure~\ref{fig:rate-landscape}
summarizes the resulting sample-complexity landscape.

\subsection{Technical overview}

\paragraph{A shared upper-bound template.}
Both upper bounds first learn a real valued intermediate score $S\in[0,1]$ and then transform it
into the final discrete prediction $P$.  We call this second stage \emph{score
coarsening}: it replaces $S$ by a predictor with finitely many 
prediction values.

The final objective involves two kinds of tests.  Each $g\in\calG$ defines the
multiaccuracy test $\phi_g(x,p)=g(x)$.  Calibration also has a test-based form:
writing $Q$ for the final predictor and $P\sim Q_X$ for its realized
prediction,
\[
  \ECE(Q)
  =
  \sup_{h:[0,1]\to\{-1,1\}}
  \E[h(P)(Y-P)].
\]
Thus each sign-valued function $h$ of the prediction defines a calibration
test $\phi_h(x,p)=h(p)$.

Now fix a coarsening rule.  For any bounded test $\phi(x,p)$ of the context
and final prediction, define its \emph{lift} to the intermediate score by
\[
  \phi^\uparrow(x,s)
  =
  \E[\phi(x,P)\mid X=x,S=s],
\]
where the expectation is over any randomness in the coarsening rule.
The rule uses no information about $Y$ once $(X,S)$ is fixed, and so
Lemma~\ref{lem:postprocessing-decomposition} will give:
\begin{equation}
  \E[\phi(X,P)(Y-P)]
  =
  \E[\phi^\uparrow(X,S)(Y-S)]
  +
  \E[\phi(X,P)(S-P)].
  \label{eq:overview-residual-decomposition}
\end{equation}
For a multiaccuracy test $\phi_g$, the lift is simply $g$.  For a calibration
test $\phi_h$, write
\[
  a_h(x,s)=\E[h(P)\mid X=x,S=s],
\]
for its lift.  The score learner  controls residuals of $S$ against
both the original tests $g\in\calG$ and these additional lifted calibration
tests.  The analysis of the coarsening rule controls the second term in the
decomposition, and we choose the coarsening rule as a function of $S$ to minimize it.

There is an apparent circularity: the chosen coarsening rule depends on the
learned score, but its lifted tests must be known when the score is learned.
Both algorithms resolve this by specifying a finite menu of candidate rules
in advance and learning $S$ against every induced lifted test.  The randomized
algorithm then chooses one rule from the menu.  The deterministic algorithm
will use different rules on different parts of the context space.

The randomized algorithm coarsens using mean-preserving rounding to a grid.  The
deterministic algorithm coarsens using category averaging: it partitions contexts into
categories and replaces the scores in each category by their average.  We
next explain the main idea in each construction, followed by the two lower
bounds.

\subsubsection{Randomized upper bound}

We process the i.i.d. training examples
$(X_1,Y_1),\ldots,(X_n,Y_n)$ sequentially, learning score rules
$s_1,\ldots,s_n$.  Rule $s_t$ depends only on the preceding labeled examples
$(X_i,Y_i)_{i<t}$.  On observing $X_t$, it produces the score
$S_t=s_t(X_t)$ before observing $Y_t$.  On a new input $x$, the intermediate predictor
$Q^{\mathrm{pre}}$ draws an index $J$ uniformly from $[n]$ and outputs
the score $S=s_J(x)$.

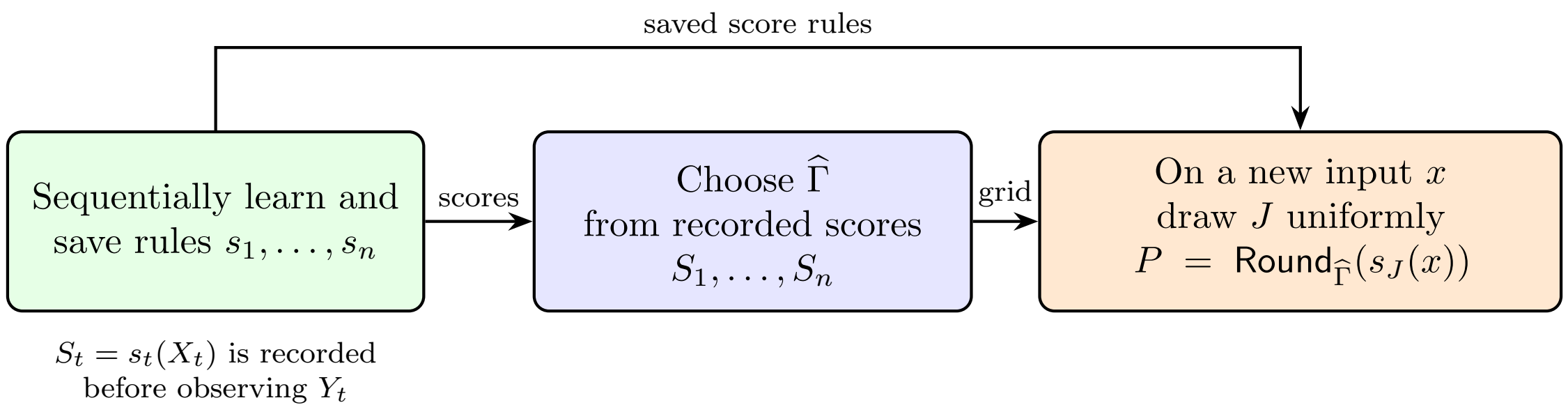
\begin{figure}[ht]
\centering
\begin{tikzpicture}[
  >=Stealth,font=\small,
  stage/.style={draw,rounded corners,align=center,inner sep=5pt,minimum height=17mm},
  every path/.style={thick}]
\node[stage,fill=green!10,text width=36mm] (learn) at (1.9,0)
  {Sequentially learn and \\save rules $s_1,\ldots,s_n$};
\node[stage,fill=blue!10,text width=38mm] (partition) at (7,0)
  {Choose $\widehat\Gamma$\\from recorded scores\\$S_1,\ldots,S_n$};
\node[stage,fill=orange!18,text width=46mm] (output) at (12.2,0)
  {On a new input $x$\\draw $J$ uniformly\\
   $P=\mathsf{Round}_{\widehat\Gamma}(s_J(x))$};
\draw[->] (learn.east)--node[above,font=\scriptsize] {scores} (partition.west);
\draw[->] (partition.east)--node[above,font=\scriptsize] {grid} (output.west);
\draw[->] (learn.north)--(1.9,1.65)--node[above,font=\scriptsize]
  {saved score rules} (12.2,1.65)--(output.north);
\node[font=\scriptsize,align=center,anchor=north] at ($(learn.south)+(0,-.12)$)
  {$S_t=s_t(X_t)$ is recorded\\before observing $Y_t$};
\end{tikzpicture}
\caption{Rule $s_t$ is learned from the preceding examples and produces the
recorded score $S_t=s_t(X_t)$ before observing $Y_t$.  These $n$ scores
determine a grid from the fixed candidate menu.  On a new input, the final
predictor evaluates a uniformly chosen saved score rule and then rounds its
score to the selected grid while preserving its mean.}
\label{fig:pipeline}
\end{figure}

In addition to the original multiaccuracy tests $\calG$, the learner
controls a finite class $\calA$ of lifted calibration tests, corresponding to the linear interpolations of binary signs at points on a fixed candidate collection of grids.  For every
candidate grid and sign assignment to its points, $\calA$ contains the
corresponding lift from \eqref{eq:overview-residual-decomposition}.
After training, we choose a grid $\widehat\Gamma$ with $O(m)$ intervals
using the recorded scores $S_1,\ldots,S_n$.  The parameter $m$ trades off
learning a larger auxiliary class against more accurate rounding.
The final predictor rounds $S$ to this grid while preserving its
conditional mean.  Figure~\ref{fig:pipeline} summarizes the construction.

\paragraph{Learning the pre-rounded scores.}
We now bound the size of the auxiliary class $\calA$, which is separate
from the original multiaccuracy test class $\calG$.
Each candidate grid consists of $O(m)$ partition points selected from a fixed fine mesh
with polynomially many points in $n$, and each selected point receives a sign
in $\{-1,1\}$ in each calibration test defined on that grid. For every collection of $O(m)$ grid points from the mesh and assignment of signs to the grid points, we include in $\calA$ the function that linearly interpolates the signs between the grid points. The choices of points and signs give at most
$\exp(O(m\log n))$ lifted tests, so $\log|\calA|=O(m\log n)$.
Standard multiobjective learning methods
(e.g. \citep{lee2022online,noarov2025high,farina2026efficient}) let us learn score rules promising:
\[
  \max\left\{
    \sup_{g\in\calG}\left|\frac1n\sum_{t=1}^n g(X_t)(Y_t-S_t)\right|,
    \sup_{a\in\calA}\left|\frac1n\sum_{t=1}^n a(S_t)(Y_t-S_t)\right|
  \right\}
  \lesssim\sqrt{\frac{\log|\calG|+m\log n}{n}}.
\]

\paragraph{Rounding for calibration.}
The original tests ensure multiaccuracy of the scores, but the auxiliary tests do not directly ensure calibration (since they correspond to lifts of calibration tests for scores that are restricted to discrete grids, but our learned scores are not confined to a grid).  Deterministically
rounding scores to $m$ grid values can introduce $\Theta(1/m)$ error; balancing this
with the learning error $\sqrt{m/n}$ gives only the  $n^{-1/3}$ rate from prior work.
We instead randomly round $S$ to one of its two neighboring grid points so that
\[
  \E[P\mid S]=S.
\]
This preserves
every prediction-independent multiaccuracy constraint without loss.

\begin{figure}[H]
\centering
\begin{tikzpicture}[x=1cm,y=1cm,>=Stealth,font=\footnotesize]
\def\base{3.05}
\draw[thick] (-0.2,\base) -- (10.6,\base);
\node[below=1pt] at (-0.05,\base) {\scriptsize$0$};
\node[below=1pt] at (10.4,\base) {\scriptsize$1$};
\node[anchor=west,font=\scriptsize] at (0.1,\base+1.75) {density of $S$};
\draw[blue!70!black,thick,smooth] plot coordinates
  {(0,\base+0.05)(1,\base+0.18)(2,\base+0.45)(3,\base+0.85)(4,\base+1.18)
   (5,\base+1.30)(6,\base+1.16)(7,\base+0.82)(8,\base+0.45)(9,\base+0.18)(10,\base+0.05)};
\foreach \x in {0,2.2,3.4,4.2,5.0,5.8,6.6,7.8,10}{
  \draw[blue!70!black,line width=1pt] (\x,\base+0.16)--(\x,\base-0.16);
}
\node[font=\scriptsize,anchor=east,align=right] at (10.3,\base+1.75)
  {partition $\widehat\Gamma$:\\denser where mass is high};
\def\bb{0.35}
\draw[thick] (-0.2,\bb) -- (10.6,\bb);
\node[anchor=east,font=\scriptsize] at (-0.15,\bb+0.55) {$q_j$};
\draw[gray!60!black,densely dashed] (-0.2,\bb+0.8)--(10.6,\bb+0.8);
\node[gray!60!black,anchor=west,font=\scriptsize,fill=white,inner sep=1pt] at (9.15,\bb+1.02) {$\tau\asymp m^{-2}$};
\foreach \mx/\h/\lab in {1.1/0.81/{$q_0$},2.8/0.82/{},3.8/0.80/{},4.6/0.83/{},5.4/0.81/{},6.2/0.80/{},7.2/0.82/{},8.9/0.50/{}}{
  \fill[orange!80!black,opacity=0.85] (\mx-0.32,\bb) rectangle (\mx+0.32,\bb+\h);
  \node[font=\scriptsize,above=0pt] at (\mx,\bb+\h) {\lab};
}
\node[font=\scriptsize,anchor=west] at (9.25,\bb+0.25) {$q_{M-1}$};
\node[align=center,font=\scriptsize,text width=10.3cm] at (5.2,\bb-0.65)
  {the interior bars stay nearly flat, so the total variation with endpoints
  $q_0+q_{M-1}+\sum_j|q_j-q_{j-1}|$ is small};
\end{tikzpicture}
\caption{The adaptive partition makes the coefficients $(q_j)$ nearly equal.
\emph{Top:} the greedy rule places partition points $\gamma_j$ from the
marginal distribution of $S$, more densely where $S$ has more mass, so that
each full interval has coefficient close to
$\tau\asymp m^{-2}$.  \emph{Bottom:} the resulting sequence is nearly flat.
The terms $q_0$ and $q_{M-1}$ are the boundary jumps from
$q_{-1}=q_M=0$; only the final interval may fall below the threshold.  The
endpoint-inclusive total variation, and hence the calibration error from
rounding, is $O(m^{-2})$.}
\label{fig:variance}
\end{figure}
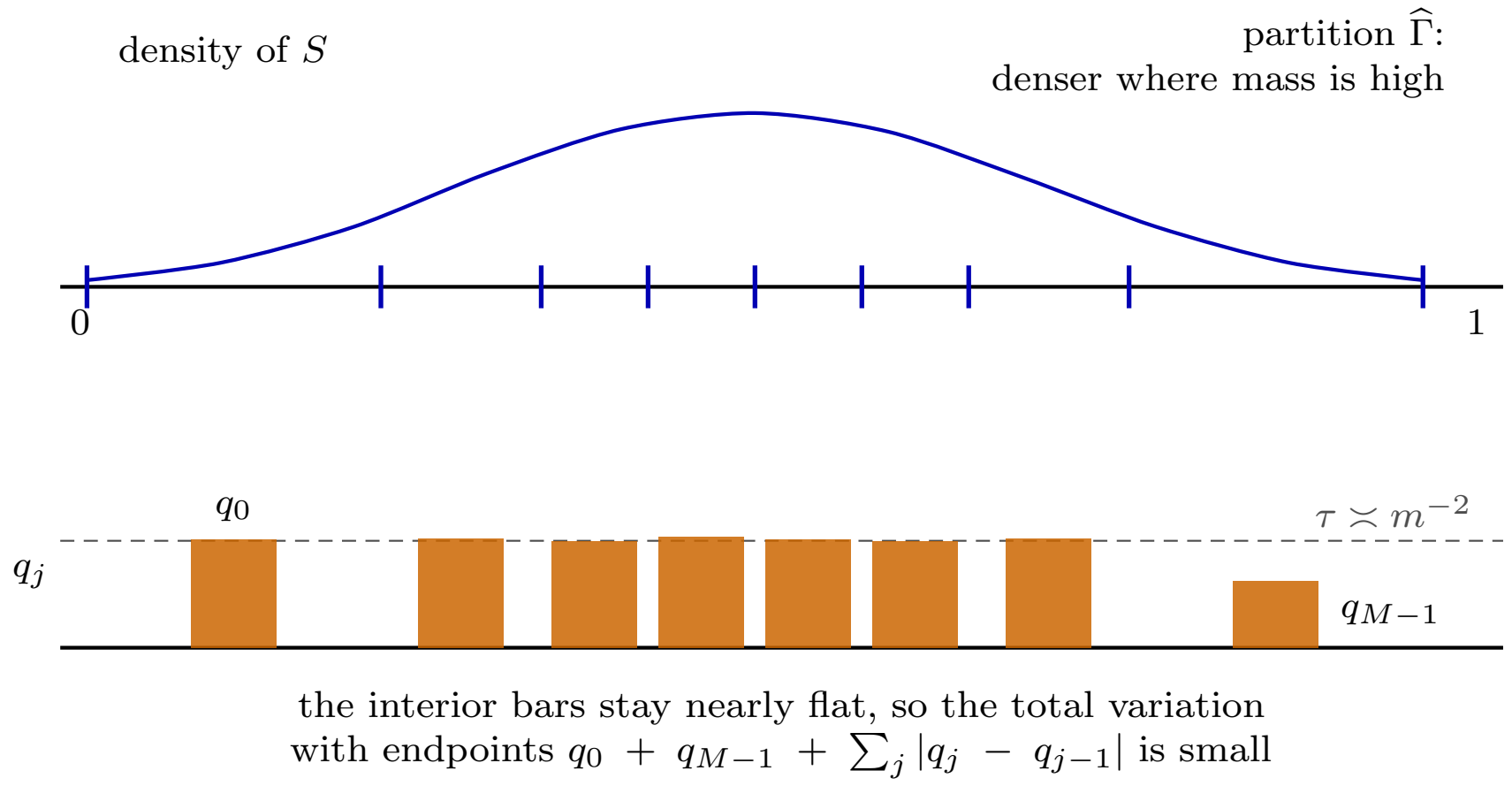

For calibration, fix a grid
$\Gamma=(\gamma_0,\ldots,\gamma_M)$ and a sign test $h$ on its points.  This
rounding rule depends on $S$ but not on $X$, so its lift is the function
\[
  a_h(s)=\E[h(P)\mid S=s].
\]
This function is the linear interpolant of the signs $h(\gamma_j)$ and  belongs to
$\calA$.  It remains to control the error created by rounding.  Write
$h_j=h(\gamma_j)$ and let $q_j$ denote the coefficient measuring the expected
rounding contribution from the $j$th interval:
\[
  q_j
  =
  \E\left[
    \frac{(S-\gamma_j)(\gamma_{j+1}-S)}
         {\gamma_{j+1}-\gamma_j}
    \one\{\gamma_j<S<\gamma_{j+1}\}
  \right].
\]
A direct calculation from the two-point rounding rule gives
\[
  \E[h(P)(S-P)]
  =
  \sum_{j=0}^{M-1}(h_j-h_{j+1})q_j .
\]
The factor $h_j-h_{j+1}$ creates the potential for cancellation between adjacent prediction
values.  If all the coefficients $q_j$ were to equal a common value $q$, then the sum would 
telescope to
\[
  \sum_{j=0}^{M-1}(h_j-h_{j+1})q
  =
  q(h_0-h_M),
\]
whose magnitude is at most $2q$, independently of the number of intervals.
This is the cancellation the construction will approximate.  Set
$q_{-1}=q_M=0$.  Taking the supremum over $h$ and applying summation by parts shows that the
worst rounding contribution is the endpoint-inclusive total variation of the $q_j$ coefficients:
\[
  q_0+q_{M-1}+\sum_{j=1}^{M-1}|q_j-q_{j-1}|
  =
  \sum_{j=0}^{M}|q_j-q_{j-1}|.
\]
Thus we want a partition for which the coefficients $(q_j)$ are nearly
equal and small.  We construct it directly from the recorded scores
$S_1,\ldots,S_n$.  Starting with a left endpoint $a=0$, move the right
endpoint $b$ to the right along a fixed fine grid, tracking
\[
  \widehat q(a,b)
  =
  \frac1n\sum_{t:\,a<S_t<b}
  \frac{(S_t-a)(b-S_t)}{b-a}.
\]
This is the coefficient $q_j$ for the candidate interval $(a,b)$, with the
expectation replaced by the empirical average of the recorded scores.  It
is nondecreasing as $b$ increases.  End the interval at the first grid point
where $\widehat q(a,b)$ reaches a common threshold $\tau\asymp m^{-2}$,
then start the next interval there.  If the threshold is never reached,
end the final interval at $1$.  This rule produces $O(m)$ intervals whose
empirical coefficients have total variation $O(m^{-2})$.
Figure~\ref{fig:variance} illustrates the cancellation.

Each score rule $s_t$ was chosen using only preceding examples.  For each
fixed candidate grid and test, a martingale bound controls the average
residual from rounding the recorded scores by the average population
residual of the saved score rules after rounding.  The bound holds
simultaneously for all candidate grids and tests, so it remains valid for
the grid chosen using the recorded scores.  Its statistical error is of
order $\sqrt{(\log|\calG|+m\log n)/n}$.

Applying \eqref{eq:overview-residual-decomposition} to a sign test $h$, the lifted residual is
controlled because $a_h\in\calA$, while the rounding contribution is
$O(m^{-2})$.  The resulting error is
\[
  \sqrt{\frac{\log|\calG|+m\log n}{n}} + m^{-2}.
\]
Choosing $m\asymp (n/\log n)^{1/5}$ gives the theorem.

\subsubsection{Deterministic upper bound}
Our randomized construction uses randomness in two places: the score function $S$ it learns is already randomized, and the coarsening procedure adds additional randomness. To give a deterministic upper bound we need to remove both sources of randomness. 
Recall that $A=\max\{1,\log|\calG|\}$ denotes the group budget. To learn a deterministic score function, we 
 use two components of the deterministic finite-test construction of
\citet{NoarovRoth2026Deterministic}: an algorithm that learns a randomized
score satisfying a finite family of residual tests, followed by a
procedure that fixes one random seed on each part of a finite context
partition, making that score deterministic.  Given a
finite family of bounded tests $a:\calX\times[0,1]\to[-1,1]$ the resulting deterministic score $S$ has
small residual
$\E[a(X,S(X))(Y-S(X))]$ for every test $a$.  With $N$ labeled examples
and test family $\Phi$, the statistical error is
$\widetilde O(\sqrt{\log|\Phi|/N})$.  We include both the original
multiaccuracy tests and auxiliary tests that anticipate the later
coarsening.  Lemma~\ref{lem:det-finite-test-score} gives the precise
guarantee.

The main difficulty is in removing the randomness of the coarsening rule. Why does deterministic prediction need a different coarsening rule, rather than there being a simple derandomization of our randomized rounding procedure?  If the
context distribution were discrete and every context were rare, we could fix
one independent rounding coin per context: each test would average many
low-influence choices, and the realized calibrated multiaccuracy error would be close to its expectation, resulting in a simple derandomization of our randomized construction. This fails if certain contexts have substantial mass, since now fixing the randomness on each context might be a high variance operation. On the other hand, very frequent contexts can  be identified in
the training sample and their conditional label means estimated. Thus very frequent contexts could be handled as a special case.  At the target sample size,
however, a context can have enough probability mass that fixing its coin
causes substantial error, yet appear too few times to estimate its
conditional label mean to the required accuracy.
Our construction does give each very frequent context its own category and keeps
its score unchanged.  But rather than attempting to fix the randomness of other contexts it handles all remaining contexts collectively by
assigning them to a small number of categories and replacing the scores in each
category by their average.  This is the deterministic score-coarsening step.

\paragraph{Category averaging.}
Assign each sufficiently frequent context its own singleton category, and assign all
remaining contexts to $K+1$ shared categories.  On every category $C_j$, set
the final prediction to $p(x)=\E[S(X)\mid X\in C_j]$; on a singleton this
simply retains the original score. This results in a predictor that is calibrated to the score $S$ (but without further argument, not yet to the labels $Y$).  To obtain calibration relative to $Y$, it suffices to make every
signed combination of the category residuals
$e_j=\E[\one\{X\in C_j\}(Y-S)]$ small.  Maximizing
$\sum_j\sigma_j e_j$ over signs gives $\sum_j|e_j|$, and category averaging preserves
each residual.  Indeed, applying
Lemma~\ref{lem:postprocessing-decomposition} with
$\phi_j(x,p)=\one\{x\in C_j\}$ and using
$\E[S-p\mid X\in C_j]=0$ gives
\[
  \E[\one\{X\in C_j\}(Y-p)]
  =
  \E[\one\{X\in C_j\}(Y-S)].
\]
We establish this control using the auxiliary tests and the assignment
analysis below.

Writing $P_X$ for the marginal distribution of $X$, for any test $h$
the change in the multiaccuracy residual of $p$ relative to $S$ is a sum of
within-category covariances (singleton categories contribute zero):
\[
  \E[h(X)(S(X)-p(X))]
  =
  \sum_{j=0}^K P_X(C_j)\Cov(h(X),S(X)\mid X\in C_j).
\]
It remains to choose categories for which all these covariance sums are
small.

\paragraph{Choosing the categories.}
Let $\calH=\calG\cup(-\calG)$, so controlling the largest residual over
$\calH$ controls absolute error for every test in $\calG$.  Endow the tests in
$\calH$ temporarily with a distribution $\pi$.  Leave the sufficiently
frequent contexts in their singleton categories, where their scores remain
unchanged.  Partition all other contexts into low-mass parts.  For each part
$U$, independently sample $m_{\rm sp}$ tests from $\pi$ and average them:
\[
  f_U(x)=\frac1{m_{\rm sp}}\sum_{\ell=1}^{m_{\rm sp}} h_{U,\ell}(x).
\]
Here $h_{U,1},\ldots,h_{U,m_{\rm sp}}$ are the sampled tests, and we write
$U(x)$ for the part containing $x$.  We assign
$x\in U$ to a category by rounding the scalar
$f_U(x)+\lambda_{\rm cat}S(x)$, where $\lambda_{\rm cat}>0$ controls the
relative weight on the score, to an equally spaced grid of $K+1$ values; write
$J(x)$ for the resulting category index.
The score learner includes the signed category tests for every rule that uses
one tuple from $\calH^{m_{\rm sp}}$ throughout the context space.  The expected
residual under the cellwise choices is an average of these finite-menu
residuals; we later use concentration to fix the choices.

\paragraph{Why the assignment rule preserves multiaccuracy.}
Fix the sampled tests and consider a shared category $C_j$.
By construction, $f_{U(x)}(x)+\lambda_{\rm cat}S(x)$ is close to the grid
point $u_j$ defining this category.
Because $p$ is the average of $S$ within $C_j$, the difference $S-p$ has
mean zero conditional on $X\in C_j$.  Since $p$ is constant on $C_j$, we also have
\begin{align*}
  \E[S(S-p)\mid X\in C_j]
  &=\E[(S-p)^2\mid X\in C_j]+p\,\E[S-p\mid X\in C_j] \\
  &=\E[(S-p)^2\mid X\in C_j].
\end{align*}
Since $u_j$ is constant within $C_j$, its contribution is exactly zero:
\[
  \E[u_j(S-p)\mid X\in C_j]
  =u_j\E[S-p\mid X\in C_j]=0.
\]
Subtracting $u_j$ and using the identity for $S(S-p)$ therefore gives
\begin{multline*}
  \E[f_{U(X)}(X)(S-p)\mid X\in C_j]
  =-\lambda_{\rm cat}\E[(S-p)^2\mid X\in C_j] \\
  +\E[(f_{U(X)}(X)+\lambda_{\rm cat}S-u_j)(S-p)\mid X\in C_j].
\end{multline*}
The first term is nonpositive: it subtracts $\lambda_{\rm cat}$ times the
mean squared change in predictions within this category.  The second term
is the grid approximation error, which we bound below.

We now average over the sampled tests as well.  Write
$V=\E[(S-p)^2]$ for the expected squared change caused by category averaging,
where the expectation includes both the context and the randomly sampled
tests.  Let $d_h(\pi)$ be the corresponding expected change in the residual for test
$h$ from category averaging.

For each context assigned to a shared category, $f_U(x)$ averages
$m_{\rm sp}$ independent tests in $[-1,1]$.  It therefore has mean
$f_\pi(x)=\E_{h\sim\pi}[h(x)]$ and mean-squared error at most
$1/m_{\rm sp}$.  Replacing $f_U$ by $f_\pi$ in the residual multiplies
this error by $S-p$.  Cauchy--Schwarz, averaging over both contexts and
sampled tests, bounds the resulting contribution by
$\sqrt{V/m_{\rm sp}}$.  This does not require independence between
$S-p$ and the sampling error.

For the grid error, let $w$ denote the grid spacing.  For $x\in C_j$, the assignment score
$f_{U(x)}(x)+\lambda_{\rm cat}S(x)$ is within $w/2$ of $u_j$.
Averaging over contexts and sampled tests, the absolute contribution of
the grid approximation error is therefore at most
\[
  \frac w2\E|S-p|\le\frac w2\sqrt V.
\]
Contexts whose scores remain unchanged contribute zero to both errors.
Adding the bounds and using that $f_\pi$ is the $\pi$-weighted average of
the tests gives
\[
  \sum_{h\in\calH}\pi(h)d_h(\pi)
  \le
  e_{\rm app}\sqrt V-\lambda_{\rm cat}V,
  \qquad
  e_{\rm app}=\frac w2+\frac1{\sqrt{m_{\rm sp}}}.
\]
Finally recall that the distribution $\pi$ is not intrinsic to the problem, and that we are free to choose it. A fixed-point argument chooses
$\pi$ self-consistently so that it places weight only on tests maximizing
$d_h(\pi)$.  Hence
\[
  \max_h d_h(\pi)
  =
  \sum_h\pi(h)d_h(\pi).
\]
The negative quadratic term now bounds both the largest residual increase and
$V$.  Lemma~\ref{lem:det-population-assignment} makes this argument precise.  This is
the conceptual core of the deterministic upper bound.  The argument applies
to any context distribution, including an empirical one.  The training
algorithm therefore searches for a mixture with small expected empirical
residuals and then fixes the random category assignments.

\paragraph{Fixing the category assignments.}
The preceding argument selects a distribution over category assignments.  The
training algorithm now draws and stores one test set for every low-mass part
$U$.  This produces a fixed category map and hence a deterministic predictor.
A variance-sensitive concentration bound shows that the multiaccuracy
residuals remain small with high probability over the training randomness.
We apply this argument to the empirical context distribution, using $S$ as
the label.  Calibration relative to $S$ is then exact for every assignment:
each category predicts its empirical average score.  It remains to transfer
both guarantees to the population and the true labels $Y$.

\paragraph{Implementing the population argument from samples.}
A preprocessing sample is used to identify frequent contexts, build confidence
intervals for their conditional means, and partition the remaining space
into low-mass parts.  A second labeled sample is used to learn $S$.  On a fresh
context-only sample, the algorithm then chooses the assignment distribution
and computes the final category averages.  A transfer argument compares
the final predictor's population residuals against $Y$ with its empirical
residuals against $S$, for both group tests and signs on categories.  The
finite-menu score guarantees and sample concentration control the expected
difference; the confidence
radii and squared part masses control fluctuations across the independent
assignments.  This comparison is uniform over the category averages.
Thus the same context sample can both be used to select the assignments and supply
their predictions.  Section~\ref{sec:det-upper} gives the details.

The error beyond the score-learning term is
$O(\sqrt{A/K^3})$.  Balancing it with the target error $\eps$ gives
$K\asymp(A/\eps^2)^{1/3}$.  Learning the resulting finite test class uses
$\widetilde O((A+K)/\eps^2)$ samples, and hence
\[
  \widetilde O\left(
    \frac{A}{\eps^2}+
    \frac{A^{1/3}}{\eps^{8/3}}
  \right).
\]
The matching lower bound shows that this unusual cube-root dependence on
$A$ is intrinsic.  Section~\ref{sec:det-upper} gives the formal
construction and proof.

\subsubsection{Lower bounds}

\paragraph{A shared hard family with different decoding mechanisms.}
Both lower bounds use the same basic construction.  On an ordered finite
domain, the conditional mean at point $i$ is a known increasing baseline value plus
a small perturbation determined by a hidden randomly chosen sign vector $\theta$.  We show that a
low-error predictor could be used to decode many coordinates of $\theta$.  The
shared recovery bound in Lemma~\ref{lem:sign-recovery-information} then
shows that too few samples cannot produce such a predictor.

What differs in the two bounds is how the predictor reveals the hidden vector.
For an arbitrary randomized predictor, calibration and multiaccuracy with
respect to the single test $g(x)=x$ force the realized prediction close to
the public baseline in squared error.  This controls distance at the
relatively coarse scale $\sqrt\eps$, enough to distinguish a grid of
$k\asymp\eps^{-1/2}$ points and give the $\eps^{-5/2}$ lower bound.  For a
deterministic predictor, we use the test-family budget
$A$ to resolve more points.  On a domain of $R$ ordered points, a
family of $\exp(\Theta(A))$ sign tests forces the average distance between
$p(i)$ and the public baseline to be at most
$O(\eps\sqrt{R/A})$.  The baseline values are spaced $1/R$ apart, so balancing
these two scales gives
$R\asymp A^{1/3}\eps^{-2/3}$.  For this choice, most prediction values are
distinct and hence form singleton level sets.  Calibration on these
singletons reveals $\Omega(R)$ coordinates of $\theta$, and the resulting
information bound gives the $A^{1/3}\eps^{-8/3}$ lower-bound term.

\paragraph{Randomized predictors.}
The randomized lower bound uses the finite-domain construction of
\citet{GibbsTibshirani2025} for deterministic predictors, with the single
bounded test $g(x)=x$.  A sample draws a uniform grid index
$I\in\{1,\ldots,k\}$ and sets $X=x_I$, where $x_i=i/k$.  The label $Y$ is
Bernoulli with conditional mean
\[
  \E[Y\mid X=x_i]=b_i+\Delta\theta_i,
  \qquad b_i=\frac14+\frac{x_i}{2},
  \qquad \Delta\asymp k^{-2},
\]
where the hidden bits $\theta_1,\ldots,\theta_k$ are chosen independently
and uniformly from $\{0,1\}$.  

Calibration controls $\E[Y-P]$ and $\E[P(Y-P)]$, and multiaccuracy controls
$\E[X(Y-P)]$.  Since the baseline is linear in $X$, combining these moments
bounds $\E[(P-b_I)^2]$ by $O(\eps+\Delta^2)$.  The baselines are separated
by $\Theta(1/k)$, so at the target accuracy a prediction is closest to its
input's baseline most of the time.

To recover the hidden bits, group prediction values by their nearest
baseline value.  Calibration controls the total label residual in each such bin.
Subtracting the contribution of $b_I$, the original input's baseline value, from
the predictions in a bin therefore estimates its hidden offset.   Predictions assigned to incorrect
bins affect only a small fraction of the mass, so thresholding these
estimates recovers most of the bits.  Section~\ref{sec:lower} gives
the quantitative argument.

\paragraph{Deterministic predictors.}
The deterministic lower bound uses $R$ ordered points with public baseline
values $t_1<\cdots<t_R$ separated by $\Theta(1/R)$.  The conditional mean at
point $i$ is $t_i+\Theta(\eps)\theta_i$ for an unknown sign $\theta_i$.

We construct a family of sign tests with a detection property: unless the
average distance between $p_i=p(i)$ and $t_i$ is small, some test has
noticeable correlation with $(t_i-p_i)_{i=1}^R$.  The strength of this
property grows with the test budget.  Multiaccuracy therefore places $p_i$
within a small fraction of the baseline spacing for most indices.

This is where determinism enters.  A deterministic predictor partitions the
domain into level sets $\{i:p_i=v\}$.  No two indices whose predictions are
close to their respective baselines can share a value, because the baselines
are separated.  Moreover, any such index that does share its value must share
it with one of the few poorly approximated indices.  Hence most points form
singleton prediction level sets.

On a singleton level set $\{i\}$, calibration compares $p_i$ directly with
$t_i+\Theta(\eps)\theta_i$.  The sign of $p_i-t_i$ must therefore recover
$\theta_i$ on most such points; otherwise their calibration errors add up.
Thus low multiaccuracy error and low ECE decode a constant fraction of the
hidden signs.  Flipping one sign changes the label mean by $O(\eps)$ at
only one of the $R$ equally likely contexts, so the shared recovery lemma
requires $\Omega(R/\eps^2)$ samples.
With $R\asymp A^{1/3}\eps^{-2/3}$, this is the claimed
$\Omega(A^{1/3}\eps^{-8/3})$ bound.

This argument does not extend to randomized predictors because the sign tests
force only the mean prediction $\bar p(i)=\E[P\mid X=i]$ close to $t_i$,
whereas calibration conditions on the realized $P$.  For a deterministic
predictor, $P=\bar p(X)=p(X)$, so both constraints concern the same level sets.

\subsection{Related work}

Calibration goes back at least to \citet{Dawid1982}.
\citet{HebertJohnsonKimReingoldRothblum2018} introduced multicalibration and
also studied, under the name \emph{multi-AE}, the groupwise mean-accuracy
condition now called multiaccuracy (cf.~\citealp{KimGhorbaniZou2019}).  These conditions were initially motivated
as fairness guarantees that prevent systematic bias on algorithmically
identifiable subpopulations.  \citet{dwork2021outcome} later placed them in a
hierarchy of outcome-indistinguishability conditions: the lower levels recover
multiaccuracy and multicalibration, while the stronger levels permit tests with
greater access to the predictor.  Omniprediction supplied a compelling
downstream motivation by connecting these conditions to simultaneous loss
minimization~\citep{
GopalanKalaiReingoldSharanWieder2022,GopalanHuKimReingoldWieder2023}.
Subsequent work showed that full multicalibration is stronger than necessary
for several of these applications, including
omniprediction~\citep{GopalanHuKimReingoldWieder2023,garg2024oracle,okoroafor2025near}.
Sharp lower bounds for multicalibration in both batch and online settings
further motivate the study of weaker
conditions~\citep{CollinaLuNoarovRoth2026,collina2026optimal}.

\citet{GopalanHuKimReingoldWieder2023} show that calibrated multiaccuracy
suffices for omniprediction for important classes of losses.
\citet{CasacubertaGopalanKanadeReingold2025} study calibrated multiaccuracy
more generally and show that it recovers strong agnostic learning and
optimal-density hardcore measures, consequences that had previously been
obtained using stronger multicalibration-based tools.
\citet{DworkTankala2025Supersimulators} similarly show that the
multicalibration-based characterization of product-distribution
indistinguishability due to
\citet{MarcussenPuttermanVadhan2025Product} can be obtained from calibrated
multiaccuracy.

These applications motivate understanding the statistical complexity of
calibrated multiaccuracy, a study initiated by \citet{GibbsTibshirani2025}.
They prove an $\Omega(\eps^{-5/2})$ sample lower bound for algorithms whose
output is a deterministic predictor.  They also prove a
multiaccuracy lower bound that gives the $\log|\calG|/\eps^2$ term in our
randomized lower bound after specializing to a finite bounded test family.  Our
$\eps^{-5/2}$ lower bound uses their hard family, with a decoder that also
applies to randomized predictors.

Several other papers compare randomized and deterministic prediction in
batch learning.  \citet{NoarovRoth2026Deterministic} show that randomized
predictors are \emph{not} needed for minimax-optimal learning of
multicalibrated predictors, in contrast to what we find for calibrated
multiaccuracy.  Agnostic multi-distribution learning is another problem where
randomization is needed.  In that problem, the learning algorithm seeks one predictor
whose worst-case risk is small across several data distributions.  Recent work gives
near-optimal sample complexity for VC classes~\citep{Peng2024SampleMDL,ZhangZhanChenDuLee2024OptimalMDL}; in particular,
\citet{ZhangZhanChenDuLee2024OptimalMDL} obtain the optimal order with
randomized hypotheses and prove that deterministic hypotheses can require more
samples.  \citet{LarsenMontasserZhivotovskiy2024DerandomizingMDL} show further
that, in the fully general multi-distribution setting, derandomizing such
predictors can be computationally hard even when ERM over the benchmark class
is efficient.  Our setting is different: calibrated multiaccuracy is evaluated
under a single joint distribution, not by worst-case risk across several
distributions.  The connection is that both lines of work identify natural
batch learning objectives for which randomized prediction can reduce
sample complexity.

\section{Setting and calibration definitions}

Let $(X,Y)\sim\mathcal D$, where $X\in\calX$ and $Y\in[0,1]$, and write $P_X$
for the marginal distribution of $X$.  A randomized
predictor $Q=(Q_x)_{x\in\calX}$ assigns to each input $x$ a distribution $Q_x$
on $[0,1]$.  When
evaluating the predictor, we draw $P\sim Q_X$
using randomness independent of $Y$ conditional on $X$.  All expectations below
include this prediction randomness.  We write the error as a function of the
predictor $Q$, with the realized prediction $P$ appearing inside the defining
expectations.

For a finite class $\calG$ of maps $g:\calX\to[-1,1]$, define the multiaccuracy error of a predictor as:
\[
  \MA(Q;\calG)
  =
  \sup_{g\in\calG}\left|\E[g(X)(Y-P)]\right|.
\]
The population expected calibration error is
\[
  \ECE(Q)
  =
  \E\left|\E[Y-P\mid P]\right|.
\]
Equivalently,
\begin{equation}
  \ECE(Q)
  =
  \sup_{h:[0,1]\to\{-1,1\}}
  \E[h(P)(Y-P)],
  \label{eq:ece-dual}
\end{equation}
where $h$ ranges over sign-valued one-dimensional tests of the prediction
value.  This is equivalent to the absolute-value definition because the
supremum is attained by choosing $h(P)$ to be the sign of
$\E[Y-P\mid P]$, with either sign chosen when this conditional expectation is
zero.

The calibrated multiaccuracy error of a predictor is
\[
  \CMA(Q;\calG)=\max\{\MA(Q;\calG),\ECE(Q)\}.
\]

\subsection{A score coarsening identity}
\label{sec:postprocessing-identity}

Both upper bounds learn an intermediate score and then coarsen it.  The
following simple identity separates the residual already controlled while learning
the score from the error introduced by this coarsening.

\begin{lemma}[Residual decomposition under postprocessing]
\label{lem:postprocessing-decomposition}
Let $(X,Y,S)$ be jointly distributed, with $S\in[0,1]$.  For each input-score
pair $(x,s)$, let $\mathsf R_{x,s}$ be a distribution on $[0,1]$, and draw the
postprocessed prediction $P\sim\mathsf R_{X,S}$ using randomness independent
of $Y$ conditional on $(X,S)$.  For a bounded test
$\phi:\calX\times[0,1]\to[-1,1]$, define its lift to the intermediate score by
\[
  \phi^\uparrow(x,s)
  =
  \E_{P\sim\mathsf R_{x,s}}[\phi(x,P)].
\]
Then
\begin{equation}
\begin{aligned}
  \E[\phi(X,P)(Y-P)]
  &={}
  \E[\phi^\uparrow(X,S)(Y-S)]
  +
  \E[\phi(X,P)(S-P)].
\end{aligned}
\label{eq:postprocessing-decomposition}
\end{equation}
\end{lemma}

\begin{proof}
Split $Y-P=(Y-S)+(S-P)$.  Conditional independence gives
\[
  \E[\phi(X,P)(Y-S)\mid X,S]
  =
  \phi^\uparrow(X,S)\E[Y-S\mid X,S].
\]
Taking expectations and applying the tower property to the right-hand side
proves the identity.
\end{proof}

The first term in \eqref{eq:postprocessing-decomposition} explains why the
score learner will anticipate a finite menu of rules: it controls
the corresponding lifted tests before the data are used to select a rule or
a distribution over rules.
The two upper bounds differ in how they control the second term.

\section{Ingredients for the randomized upper bound}
\label{sec:upper-ingredients}

\subsection{Learning against finitely many residual tests}
\label{sec:finite-prediction-tests}

The first ingredient is a standard online algorithm that controls finitely
many residual tests.  Such multiobjective learning arguments have been used
for online multiaccuracy and multicalibration
\citep{gupta2021online,lee2022online,noarov2025high,farina2026efficient}.
We use its predictions to choose the rounding partition, then convert the
rounded predictions into a batch predictor.  Appendix~\ref{app:online-proof}
gives the standard analysis for completeness.

\begin{lemma}[Online residual control for finite tests]
\label{lem:online-finite-tests}
Let $N\ge1$, and let $\Phi$ be a finite nonempty class of tests
$\phi:\calX\times[0,1]\to[-1,1]$ that are continuous in their second argument.
There is an online algorithm that, before round $t$, determines a prediction
rule $s_t:\calX\to[0,1]$ from the preceding examples.  It observes $X_t$,
predicts $S_t=s_t(X_t)$, and then observes $Y_t$.  For every sequence of
examples in $\calX\times[0,1]$,
\[
  \sup_{\phi\in\Phi}
  \left|\frac1N\sum_{t=1}^N\phi(X_t,S_t)(Y_t-S_t)\right|
  \le C\sqrt{\frac{\log(2|\Phi|)}N}.
\]
\end{lemma}

This lemma controls the empirical residuals of the sequential predictions.
For i.i.d. examples, a martingale argument will transfer the final rounded
residuals to the population in the proof of Theorem~\ref{thm:main}.
The deterministic upper bound uses an  analogous finite-test learner in
Lemma~\ref{lem:det-finite-test-score}, with additional structure to make the
intermediate predictor deterministic.

\subsection{Mean-preserving randomized rounding}
\label{sec:quantization}

The second ingredient is a one-dimensional randomized rounding scheme.  It
rounds an intermediate prediction to finitely many values while preserving its
conditional mean.

Fix a grid increment $\rho>0$ such that $1/\rho$ is an integer, and write
\[
  \Lambda_\rho=\{0,\rho,2\rho,\ldots,1\}.
\]
A grid partition is a sequence
\[
  \Gamma=(\gamma_0,\ldots,\gamma_M),
  \qquad
  0=\gamma_0<\gamma_1<\cdots<\gamma_M=1,
  \qquad
  \gamma_j\in\Lambda_\rho.
\]
We call $\gamma_0,\ldots,\gamma_M$ the partition points.  Given $\Gamma$,
define the adjacent mean-preserving randomized rounding rule
$\mathsf{Round}_\Gamma$ as follows.  If
$s\in[\gamma_j,\gamma_{j+1}]$, let
$\Delta_j=\gamma_{j+1}-\gamma_j$ and set
\[
  \Pp(\mathsf{Round}_\Gamma(s)=\gamma_j)=\frac{\gamma_{j+1}-s}{\Delta_j},
  \qquad
  \Pp(\mathsf{Round}_\Gamma(s)=\gamma_{j+1})=\frac{s-\gamma_j}{\Delta_j}.
\]
At a partition point either adjacent interval gives the same prediction $P=s$.
The rule preserves the conditional mean: if $S=s\in[\gamma_j,\gamma_{j+1}]$,
then
\begin{align*}
  \E[P\mid S=s]
  &=
  \gamma_j\frac{\gamma_{j+1}-s}{\Delta_j}
  +
  \gamma_{j+1}\frac{s-\gamma_j}{\Delta_j} \\
  &=
  \frac{\gamma_j\gamma_{j+1}-\gamma_js+\gamma_{j+1}s-\gamma_j\gamma_{j+1}}
       {\Delta_j}
  =
  s.
\end{align*}
Thus, for $P=\mathsf{Round}_\Gamma(S)$,
\begin{equation}
  \E[P\mid S]=S.
  \label{eq:mean-preserving-rounding}
\end{equation}
We next observe that the lift of a test defined on the grid under this randomized rounding scheme is simply a linear interpolation of the original test values on the grid points: 
\begin{lemma}[Lifted rounded tests are linear interpolants]
\label{lem:lifted-rounded-tests}
Fix a grid partition $\Gamma=(\gamma_0,\ldots,\gamma_M)$ and a test
$h:\{\gamma_0,\ldots,\gamma_M\}\to\mathbb R$.  Let
$P=\mathsf{Round}_\Gamma(S)$ and define
\[
  a_h(s)=\E[h(P)\mid S=s].
\]
Then, for $s\in[\gamma_j,\gamma_{j+1}]$,
\[
  a_h(s)
  =
  h(\gamma_j)\frac{\gamma_{j+1}-s}{\gamma_{j+1}-\gamma_j}
  +
  h(\gamma_{j+1})\frac{s-\gamma_j}{\gamma_{j+1}-\gamma_j}.
\]
Thus $a_h$ is the continuous piecewise-linear interpolation of the values
$h(\gamma_j)$ on the partition points.
\end{lemma}

\begin{proof}
The formula follows by averaging the two endpoint values of $h$ with the
rounding probabilities.  It is affine on each interval and equals
$h(\gamma_j)$ at each partition point, so the pieces agree at their endpoints.
\end{proof}

We next identify the rounding error that our choice of partition must control.  Fix a bounded calibration
test $h$ on the rounded prediction values and define
$a_h(s)=\E[h(P)\mid S=s]$.  Applying
Lemma~\ref{lem:postprocessing-decomposition} with $\phi(x,p)=h(p)$ gives
\[
  \E[h(P)(Y-P)]
  =
  \E[a_h(S)(Y-S)]
  +
  \E[h(P)(S-P)].
\]
The score-learning step controls the first term.  We now analyze the second,
which is the error introduced by mean-preserving rounding.

For an interval $I_j=(\gamma_j,\gamma_{j+1})$, define
\[
  r_j(s)
  =
  \frac{(s-\gamma_j)(\gamma_{j+1}-s)}
       {\gamma_{j+1}-\gamma_j}
  \one\{s\in I_j\}.
\]
For a distribution $\nu$ on $[0,1]$, write
\[
  q_j(\nu,\Gamma)=\E_{S\sim \nu}[r_j(S)].
\]
Define the following quantity from the sequence $q_j(\nu,\Gamma)$:
\[
  \operatorname{RoundErr}(\nu,\Gamma)
  =
  \sup_{h:\{\gamma_0,\ldots,\gamma_M\}\to\{-1,1\}}
  \left|
    \sum_{j=0}^{M-1}
    \bigl(h(\gamma_j)-h(\gamma_{j+1})\bigr)q_j(\nu,\Gamma)
  \right|.
\]
The supremum ranges over the sign-valued calibration tests on the finite set of possible
rounded predictions; $h(\gamma_j)$ is the value assigned to prediction
$\gamma_j$.  The next lemma shows that
$\operatorname{RoundErr}(\nu,\Gamma)$ is exactly the largest
calibration residual caused by rounding and rewrites it as the total variation
of the sequence $(q_j)$.

\begin{lemma}[Rounding error as total variation]
\label{lem:rounding-contribution-identity}
Let $S\sim \nu$, let $P=\mathsf{Round}_\Gamma(S)$, and set
$q_{-1}=q_M=0$.  Then
\begin{align*}
  \operatorname{RoundErr}(\nu,\Gamma)
  =
  \sup_{h:\{\gamma_0,\ldots,\gamma_M\}\to[-1,1]}
  \left|\E[h(P)(S-P)]\right|
  &=
  q_0+q_{M-1}+
  \sum_{j=1}^{M-1}|q_j-q_{j-1}| \\
  &=
  \sum_{j=0}^{M}|q_j-q_{j-1}|,
\end{align*}
where $q_j=q_j(\nu,\Gamma)$.
\end{lemma}

\begin{proof}
Fix a test $h:\{\gamma_0,\ldots,\gamma_M\}\to[-1,1]$ and write
$h_j=h(\gamma_j)$.
Condition on $S=s\in(\gamma_j,\gamma_{j+1})$.  Then
\begin{align*}
  \E[h(P)(S-P)\mid S=s]
  &=
  h_j\frac{\gamma_{j+1}-s}{\Delta_j}(s-\gamma_j)
  +
  h_{j+1}\frac{s-\gamma_j}{\Delta_j}(s-\gamma_{j+1}) \\
  &=
  (h_j-h_{j+1})
  \frac{(s-\gamma_j)(\gamma_{j+1}-s)}{\Delta_j}.
\end{align*}
At partition points the same conditional expectation is zero.  Taking
expectations gives
\[
  \E[h(P)(S-P)]
  =
  \sum_{j=0}^{M-1}
  \bigl(h(\gamma_j)-h(\gamma_{j+1})\bigr)q_j(\nu,\Gamma).
\]
Summation by parts makes the cancellation between adjacent intervals
explicit:
\begin{align*}
  \sum_{j=0}^{M-1}(h_j-h_{j+1})q_j
  &=
  h_0q_0-h_Mq_{M-1}
  +
  \sum_{j=1}^{M-1}h_j(q_j-q_{j-1}) \\
  &=
  \sum_{j=0}^{M}h_j(q_j-q_{j-1}),
\end{align*}
where $q_{-1}=q_M=0$.  The values $h_j$ can be chosen independently, so the
supremum over $h_j\in[-1,1]$ is attained by choosing each $h_j$ to have the
sign of $q_j-q_{j-1}$, with either sign when the difference is zero.  Its value
is therefore
\[
  \sum_{j=0}^{M}|q_j-q_{j-1}|
  =
  q_0+q_{M-1}+
  \sum_{j=1}^{M-1}|q_j-q_{j-1}|.
\]
The maximizing choices are sign-valued, so this supremum is also the quantity
$\operatorname{RoundErr}(\nu,\Gamma)$ defined above.
\end{proof}

The scale of the coefficients explains why an $m^{-2}$ bound is possible.
Writing $\Delta_j=\gamma_{j+1}-\gamma_j$ and
$s=\gamma_j+u\Delta_j$ for $u\in[0,1]$ gives
\[
  r_j(s)=\Delta_j u(1-u)\le\frac{\Delta_j}{4},
  \qquad
  q_j\le\frac{\Delta_j}{4}\Pr(S\in I_j).
\]
Thus $m^{-2}$ is the natural target when both the interval width and its
probability mass are of order $1/m$.  For example, if $S$ is uniform and the
partition consists of $m$ equal-width intervals, then every interval has
\[
  q_j
  =
  \int_{\gamma_j}^{\gamma_{j+1}}r_j(s)\,ds
  =
  \frac{1}{6m^2}.
\]
Summing the interval contributions separately would give only an $m^{-1}$
bound.  The useful cancellation  occurs across adjacent intervals.  If
$q_0=\cdots=q_{M-1}=q$, then
\[
  \sum_{j=0}^{M-1}(h_j-h_{j+1})q
  =q(h_0-h_M).
\]
Thus all interior values of $h$ cancel, regardless of how rapidly $h$ changes
across the grid, and the rounding error is at most $2q$.  In the uniform
example this is $1/(3m^2)$.  More generally,
Lemma~\ref{lem:rounding-contribution-identity} shows that only the endpoint
terms and the total variation of $(q_j)$ matter.  The construction below chooses the grid to 
make this small. 

\subsection{Choosing a low-variation rounding partition}
\label{sec:rounding-partition}

The next lemma introduces the two parameters used to choose the partition.  The
parameter $m$ is the target number of rounding intervals, up to a constant
factor; later it will be optimized against the statistical error.  The
parameter $\rho$ is a fine-grid increment used to make the candidate
partitions finite, and it will be chosen small enough that its effect is
absorbed into the $m^{-2}$ rounding term.

\begin{lemma}[Population grid rounding rule]
\label{lem:population-quantizer}
Let $\nu$ be any distribution on $[0,1]$, let $m\ge2$, and let
$\rho>0$ with $1/\rho$ an integer and $\rho\le m^{-2}$.  There is a
partition $\Gamma$ whose partition points belong to
$\Lambda_\rho=\{0,\rho,2\rho,\ldots,1\}$, with at most $4m+2$ intervals,
such that
\[
  \operatorname{RoundErr}(\nu,\Gamma)\le \frac{C}{m^2}.
\]
\end{lemma}

\begin{proof}
Let $S\sim\nu$; all expectations in this proof are with respect to this
score distribution.
By Lemma~\ref{lem:rounding-contribution-identity}, it suffices to make the
interval contributions $q_j$ nearly equal while keeping their common scale
small.  We do this constructively by iteratively growing the next interval, and then ending it and beginning the next when its contribution first reaches
a common threshold.

\paragraph{Equalizing the interval contributions.}
Let
$\tau=\frac{1}{64m^2}.$
Starting from $\gamma_0=0$, greedily choose $\gamma_{j+1}$ to be the smallest
gridpoint larger than $\gamma_j$ such that
\[
  \E\left[
    \frac{(S-\gamma_j)(\gamma_{j+1}-S)}
         {\gamma_{j+1}-\gamma_j}
    \one\{\gamma_j<S<\gamma_{j+1}\}
  \right]
  \ge \tau,
\]
if such a gridpoint exists.  Otherwise set $\gamma_{j+1}=1$.
Stop as soon as the right endpoint reaches $1$.

Call an interval \emph{full} if the search finds a right endpoint whose
contribution is at least $\tau$.  If the search never reaches the threshold,
the algorithm adds one final leftover interval ending at $1$.

The fine grid limits the total overshoot.  Fix a left endpoint $a$.
Scores $s\le a$ contribute zero.  For $s>a$, the contribution is
nondecreasing in the right endpoint $b$; where $b>s$ its derivative is
\[
  \frac{d}{db}\frac{(s-a)(b-s)}{b-a}
  =\frac{(s-a)^2}{(b-a)^2}\in[0,1].
\]
It is zero for $b\le s$ and continuous at $b=s$.  Increasing the right
endpoint by one grid step therefore changes the contribution by at most
$\rho$, and only for scores inside the new interval.
Write $I_j=(\gamma_j,\gamma_{j+1})$ and let
$\nu(I_j)=\Pr_{S\sim\nu}(S\in I_j)$ denote its score probability mass.
At the first threshold crossing, a full interval consequently has
\[
  q_j=\tau+e_j,\qquad
  0\le e_j\le\rho\,\nu(I_j).
\]
The intervals are disjoint, so $\sum_{\text{full }j}e_j\le\rho$.
A final leftover interval has contribution less than $\tau$.

\paragraph{Bounding the number of intervals.}
Every full interval has $q_j\ge\tau$.  Since
\[
  q_j
  \le
  \frac{\Delta_j}{4}\nu((\gamma_j,\gamma_{j+1})),
\]
each full interval satisfies
\[
  \Delta_j\nu((\gamma_j,\gamma_{j+1}))\ge4\tau.
\]
If $J$ is the number of full intervals, Cauchy--Schwarz and disjointness give
\[
  2J\sqrt{\tau}
  \le
  \sum_{j=0}^{J-1}\sqrt{\Delta_j\nu((\gamma_j,\gamma_{j+1}))}
  \le
  \left(\sum_j\Delta_j\right)^{1/2}
  \left(\sum_j\nu((\gamma_j,\gamma_{j+1}))\right)^{1/2}
  \le1.
\]
Thus $J\le(2\sqrt\tau)^{-1}=4m$.  There is at most one leftover interval, so
$M\le4m+1$.

\paragraph{Bounding the rounding error.}
Replace every full coefficient $q_j$ by $\tau$ and retain the final
leftover coefficient, if there is one.  This reference sequence is constant
until a possible final drop to a value below $\tau$, so its variation,
including both endpoints at zero, is at most $2\tau$.
Restoring the overshoots increases variation by at most
$2\sum_{\text{full }j}e_j\le2\rho$, since each coefficient occurs in two
adjacent differences.  Lemma~\ref{lem:rounding-contribution-identity} gives
\[
  \operatorname{RoundErr}(\nu,\Gamma)\le2\tau+2\rho\le C/m^2.
\]
\end{proof}

We will apply this construction to the empirical distribution of the
sequential predictions.  Its guarantee holds for any distribution, so this
step requires no independence among those predictions.

\subsection{Prediction-value tests induced by rounding}
\label{sec:test-class}

Lemma~\ref{lem:lifted-rounded-tests} says that every sign-valued test of a
rounded predictor lifts to a piecewise-linear test of the pre-rounded score.
The next finite class collects exactly these lifted tests for grid partitions
with at most $M$ intervals.  For an integer $M\ge1$ and grid $\Lambda_\rho$,
let $\calA_{M,\rho}$ be the class of functions obtained as follows: choose a
grid partition $\Gamma=(\gamma_0,\ldots,\gamma_L)$ with $L\le M$ intervals,
choose a sign-valued test
$h:\{\gamma_0,\ldots,\gamma_L\}\to\{-1,1\}$ on the partition points, and
include the lifted function
\[
  a_h(s)=\E[h(\mathsf{Round}_\Gamma(s))],
\]
where the expectation is only over the rounding randomness.  Equivalently, if
$P=\mathsf{Round}_\Gamma(S)$, then $a_h(s)=\E[h(P)\mid S=s]$; by
Lemma~\ref{lem:lifted-rounded-tests}, $a_h$ is the continuous piecewise-linear
interpolation of the values $a_h(\gamma_j)=h(\gamma_j)$.

We can bound the size of $\calA_{M,\rho}$ by counting.  Write
$K_\rho=|\Lambda_\rho|=1/\rho+1\le 2/\rho$.  For a fixed number $L$ of
intervals, there are at most $\binom{K_\rho}{L+1}$ choices of partition points
and $2^{L+1}$ choices of signs.  Therefore
\[
  |\calA_{M,\rho}|
  \le
  \sum_{L=1}^{M}\binom{K_\rho}{L+1}2^{L+1}
  \le
  (M+1)(2K_\rho)^{M+1}.
\]
Taking logarithms gives
\[
  \log|\calA_{M,\rho}|
  \le
  \log(M+1)+(M+1)\log(2K_\rho).
\]
Here $K_\rho\le2/\rho$, so $\log(2K_\rho)\le C\log(e/\rho)$.  The
$\log(M+1)$ term is also absorbed: since $M\ge1$ and $\log(e/\rho)\ge1$, we
have $\log(M+1)\le M\le M\log(e/\rho)$.  Hence, for a universal
constant $C$,
\begin{equation}
  \log|\calA_{M,\rho}|
  \le
  C M\log(e/\rho).
  \label{eq:A-cardinality}
\end{equation}

Thus $\calA_{M,\rho}$ is the auxiliary finite class needed for the learning
step: controlling the pre-rounded score against all functions in
$\calA_{M,\rho}$ controls every sign-valued calibration test after rounding to
any grid partition with at most $M$ intervals.

\section{Upper bound for randomized predictors}
\label{sec:main-proof}

We now combine the residual identity \eqref{eq:postprocessing-decomposition}
with the ingredients from
Section~\ref{sec:upper-ingredients}.  We learn score rules sequentially and
choose the rounding partition from the scores recorded during training.
The output rounds a uniformly chosen score rule.

\begin{algorithm}[t]
\caption{Learning a randomized calibrated-multiaccurate predictor}
\label{alg:randomized-cma}
\begin{enumerate}[leftmargin=*,label=\textbf{\arabic*.}]
\item \textbf{Input:} $n$ labeled examples, a finite test class $\calG$, and a
failure probability $\delta$.
\item Set
\[
  m=\max\left\{
    2,\left\lceil
      \left(\frac{n}{\log(2n/\delta)}\right)^{1/5}
    \right\rceil
  \right\},
  \qquad
  \rho=m^{-2},
  \qquad
  M_*=4m+2,
\]
and let $\calA=\calA_{M_*,\rho}$ be the lifted test class from
Subsection~\ref{sec:test-class}.
\item Run Lemma~\ref{lem:online-finite-tests} on all $n$ examples with
\[
  \Phi_{\calG,\calA}
  =
  \{(x,s)\mapsto g(x):g\in\calG\}
  \cup
  \{(x,s)\mapsto a(s):a\in\calA\}.
\]
Save the prediction rules $s_1,\ldots,s_n$ and the scores
$S_t=s_t(X_t)$ made before observing $Y_t$.
\item Apply the greedy construction from
Lemma~\ref{lem:population-quantizer} to
$\widehat\nu=n^{-1}\sum_{t=1}^n\delta_{S_t}$, with parameters $m,\rho$,
to obtain a grid partition $\widehat\Gamma$.
\item \textbf{Output:} the randomized predictor $Q$ that, on a new input $x$,
draws $J$ uniformly from $\{1,\ldots,n\}$ and then draws
\[
  P=\mathsf{Round}_{\widehat\Gamma}(s_J(x))
\]
using fresh independent rounding randomness.
\end{enumerate}
\end{algorithm}

\begin{theorem}[Randomized upper bound for calibrated multiaccuracy]
\label{thm:main}
There is a universal constant $C<\infty$ such that the following holds.  Let
$\calG$ be a finite class of functions $g:\calX\to[-1,1]$.  Given
$n\ge 2$ i.i.d. samples from a distribution on $\calX\times[0,1]$ and
$\delta\in(0,1/2)$, Algorithm~\ref{alg:randomized-cma} returns a randomized
predictor $Q=(Q_x)_{x\in\calX}$ such that, with probability at least $1-\delta$ over the
training sample and the construction randomness,
\[
  \CMA(Q;\calG)
  \le
  C\left\{
    \sqrt{\frac{\log(2|\calG|/\delta)}{n}}
    +
    \left(\frac{\log(2n/\delta)}{n}\right)^{2/5}
  \right\}.
\]
\end{theorem}

The computational issue is that the auxiliary prediction-value test class
$\calA$ is too large to enumerate directly.  There is an efficient implementation which represents
each piecewise-linear test by the fine-grid breakpoints where its sign is
specified.  With this representation, the exponential-potential update factors
over paths in a directed acyclic graph on the fine grid, so the online algorithm
can compute its next prediction by dynamic programming rather than by summing
over all tests.  The rounding partition can also be found from sorted
predictions and prefix sums.  Appendix~\ref{sec:algorithmic} gives the details of this efficient implementation; in the body we focus on sample complexity.

The theorem immediately gives the following sample-complexity form.

\begin{corollary}[Sample complexity for randomized predictors]
\label{cor:sample-complexity}
There is a universal constant $C<\infty$ such that, for every
$\eps,\delta\in(0,1/2)$, the following holds.  Write
\[
  A_\delta=\log(8|\calG|/\delta)+1.
\]
If the algorithm receives
\[
  n
  \ge
  C\left(
    \frac{A_\delta}{\eps^2}
    +
    \frac{\log(2(A_\delta+2)/(\eps\delta))}{\eps^{5/2}}
  \right)
\]
samples, it outputs a randomized predictor $Q$ satisfying
$\CMA(Q;\calG)\le \eps$ with probability at least $1-\delta$.
\end{corollary}

\begin{proof}[Proof of Theorem~\ref{thm:main}]
Keep $m\ge2$ and $\rho\le m^{-2}$ symbolic until the final optimization,
and write $\calA=\calA_{4m+2,\rho}$.

\paragraph{Controlling the rounded residuals on the sample.}
Lemma~\ref{lem:online-finite-tests} and \eqref{eq:A-cardinality} give, for
every realized sample,
\[
  \sup_{\phi\in\Phi_{\calG,\calA}}
  \left|\frac1n\sum_{t=1}^n\phi(X_t,S_t)(Y_t-S_t)\right|
  \le B_n
  :=C\sqrt{\frac{\log(2|\calG|)+m\log(e/\rho)}n}.
\]
Lemma~\ref{lem:population-quantizer}, applied to the empirical distribution
$\widehat\nu$ of these scores, gives
$\operatorname{RoundErr}(\widehat\nu,\widehat\Gamma)\le C/m^2$.

For a candidate partition $\Gamma$ with at most $4m+2$ intervals and a
sign assignment $h$ on its points, define
\[
  Z_t(\Gamma,h)
  =\E_{P\sim\mathsf{Round}_\Gamma(S_t)}[h(P)(Y_t-P)].
\]
Here the example and score are fixed; only the rounding draw is averaged.
The lift $a_{\Gamma,h}(s)=\E[h(\mathsf{Round}_\Gamma(s))]$ belongs to
$\calA$.  Applying Lemma~\ref{lem:postprocessing-decomposition} to the
uniform empirical distribution on $(X_t,Y_t,S_t)$ gives
\[
  \left|\frac1n\sum_{t=1}^n Z_t(\widehat\Gamma,h)\right|
  \le B_n+\operatorname{RoundErr}(\widehat\nu,\widehat\Gamma)
  \le B_n+C/m^2.
\]
This holds for every sign assignment on the selected partition.
For $g\in\calG$, mean preservation gives
\[
  \E_{P\sim\mathsf{Round}_\Gamma(S_t)}[g(X_t)(Y_t-P)]
  =g(X_t)(Y_t-S_t),
\]
so the empirical multiaccuracy residuals are bounded by $B_n$ for every
partition.

\paragraph{Transferring the final residuals to the population.}
Let $\mathcal F_{t-1}$ be the history before example $t$.  For each fixed
candidate pair $(\Gamma,h)$, define
\[
  \overline Z_t(\Gamma,h)
  =\E[Z_t(\Gamma,h)\mid\mathcal F_{t-1}].
\]
The rule $s_t$ is determined by this history, and example $t$ is independent
of it.  Thus $\overline Z_t(\Gamma,h)$ is the population residual of the
rule $s_t$ rounded using $\Gamma$.
Since $|Z_t(\Gamma,h)|\le1$, the differences
$Z_t-\overline Z_t$ are martingale differences bounded by $2$.
The number of candidate partition-sign pairs is at most
\[
  \sum_{L\le4m+2}\binom{|\Lambda_\rho|}{L+1}2^{L+1}
  \le \exp\{Cm\log(e/\rho)\}.
\]
Azuma--Hoeffding and a union bound over these pairs and the $|\calG|$
multiaccuracy tests therefore give, with probability at least $1-\delta$,
simultaneous deviations of the averages $n^{-1}\sum_t Z_t(\Gamma,h)$ and
$n^{-1}\sum_t g(X_t)(Y_t-S_t)$ from their conditional population averages
of at most
\[
  D_n=C\sqrt{\frac{\log(2|\calG|/\delta)+m\log(e/\rho)}n}.
\]

Fix a complete training transcript on this event.  In what follows, all
population expectations are over an independent test example and fresh
prediction randomness, conditional on this transcript.  The final
predictor draws $J$ uniformly from $[n]$ and rounds $s_J(X)$ with
$\widehat\Gamma$, so for every sign test on its support,
\[
  \E[h(P)(Y-P)]
  =\frac1n\sum_{t=1}^n\overline Z_t(\widehat\Gamma,h).
\]
The functions $\overline Z_t(\Gamma,h)$ were defined for fixed candidates;
we substitute $\widehat\Gamma$ only after obtaining the uniform event.
Consequently, the preceding empirical bounds imply
\[
  \MA(Q;\calG)\le B_n+D_n,
  \qquad
  \ECE(Q)\le B_n+D_n+C/m^2.
\]
Here the calibration bound uses the finite-support dual form
\eqref{eq:ece-dual}.  In particular,
\begin{equation}
  \CMA(Q;\calG)
  \le C\left\{
    \sqrt{\frac{\log(2|\calG|/\delta)+m\log(e/\rho)}n}
    +\frac1{m^2}
  \right\}.
  \label{eq:preopt}
\end{equation}

Choose $m$ and $\rho=m^{-2}$ as in Algorithm~\ref{alg:randomized-cma}.
When $n/\log(2n/\delta)\ge32$, the ceiling changes $m$ by at most a
constant factor, and
\[
  \sqrt{\frac{m\log(e/\rho)}n}+m^{-2}
  \le C\left(\frac{\log(2n/\delta)}n\right)^{2/5}.
\]
In the remaining case the theorem follows from $\CMA(Q;\calG)\le1$
after increasing $C$.  Substituting into \eqref{eq:preopt} proves the claim.
\end{proof}

\begin{proof}[Proof of Corollary~\ref{cor:sample-complexity}]
Let
\[
  A_\delta=\log(8|\calG|/\delta)+1,
  \qquad
  L=\log(2(A_\delta+2)/(\eps\delta)).
\]
Set
\[
  n_0=\left\lceil C\left(A_\delta\eps^{-2}+L\eps^{-5/2}\right)\right\rceil
\]
for a sufficiently large universal constant $C$.  Then
$C\sqrt{A_\delta/n_0}\le \eps/2$.  Also
$\log(2n_0/\delta)\le C' L$ after increasing constants, because
$\eps,\delta<1/2$ and $L$ already dominates
$\log(A_\delta+2)$, $\log(1/\eps)$, and $\log(1/\delta)$.  Hence
\[
  C\left(\frac{\log(2n_0/\delta)}{n_0}\right)^{2/5}
  \le
  \eps/2.
\]
Theorem~\ref{thm:main}, applied to any subset of $n_0$ examples if more are
available, gives the claim.
\end{proof}

\section{Upper bound for deterministic predictors}
\label{sec:det-upper}

This construction follows the same strategy of choosing a coarsening rule from a finite menu, as in the randomized
upper bound.  Before learning an intermediate score $S$, we enumerate a finite
family of rules that assign each context-score pair $(x,s)$ to a category.
For each rule, we include auxiliary tests that assign an arbitrary sign to
each category and measure the corresponding signed residual of $S$.
Combining the finite-test learning and cellwise rounding procedures of
\citet{NoarovRoth2026Deterministic}, we learn a deterministic score function
$S$ against the original multiaccuracy tests and these auxiliary tests.
After $S$ is known, we choose the category assignments and replace
the scores in each category by their average.

The residual identity \eqref{eq:postprocessing-decomposition} specializes
here to the identities in Lemma~\ref{lem:det-category-projection}.
The auxiliary tests control the residuals inherited from $S$, while the
assignment rule controls the error caused by replacing $S$ with the category
averages.  First suppose we could replace the scores in each category $C_j$
by their true average under the context distribution.  The resulting
predictor $p$ satisfies $p(x)=\E[S(X)\mid X\in C_j]$ for $x\in C_j$,
so $S-p$ has mean zero within every category.  Consequently, category
averaging adds no error to any signed sum of category residuals.
The same cancellation applies to every calibration test $h(p(X))$, since
$p$, and hence $h(p(X))$, is constant within each category.
For a multiaccuracy test $g(X)$, which can vary within a category,
the additional residual is instead
\[
  \E[g(X)(S(X)-p(X))]
  =\sum_j\Pp(X\in C_j)\,
    \Cov(g(X),S(X)\mid X\in C_j).
\]
The algorithm estimates the category averages from samples; we later bound
the resulting estimation error.

We first choose a distribution over assignments for which the additional
multiaccuracy residuals are small in expectation.  We then fix the random
choices during training.  Concentration, using the small probability mass of each part receiving
an independent choice, keeps these residuals small.  Calibration requires a
different argument: category averaging is exactly calibrated relative to
$S$, but we need calibration relative to $Y$.  The auxiliary tests control
the signed sums of $Y-S$ over the categories in expectation over the
assignment choices.  Concentration then controls these sums for the fixed
assignment, allowing us to transfer calibration from $S$ to $Y$.

  The training algorithm below may randomize,
but after training it outputs a fixed function of the context.  Unlike the
randomized upper bound in Section~\ref{sec:main-proof}, the construction is
information-theoretic: its direct implementation can require exponential
time.

\begin{theorem}[Deterministic upper bound]
\label{thm:det-upper}
Suppose $\calX\subseteq[0,1]^d$ is a Borel set, and let $\calG$ be a finite
nonempty family of measurable tests $g:\calX\to[-1,1]$.  There is a learning
algorithm that, given accuracy and confidence parameters
$\eps,\delta\in(0,1/10)$, outputs a deterministic predictor
$p:\calX\to[0,1]$ satisfying
\[
  \CMA(p;\calG)\le\eps
\]
with probability at least $1-\delta$ using a set of i.i.d.\ samples of size
\[
  \widetilde O\left(
    \frac{A_\delta}{\eps^2}
    +
    \frac{A_\delta^{1/3}}{\eps^{8/3}}
  \right),
  \qquad
  A_\delta=\log\frac{8|\calG|}{\delta}+1.
\]
The probability is over the sample and the training algorithm's randomness;
the output is the fixed function $p$.
\end{theorem}

\subsection{What category averaging accomplishes}

Category averaging has a direct interpretation as an $L_2$ projection: it replaces the score by its
conditional expectation given the category.  We will use the following elementary
identities which follow from this viewpoint. 

For a finite measurable partition $\mathcal C$ of $\calX$, define its
cellwise calibration error by
\[
  \operatorname{Cal}_{\mathcal C}(q)
  =
  \sum_{C\in\mathcal C}
  \left|\E[\one\{X\in C\}(Y-q(X))]\right|.
\]

\begin{lemma}[Category averaging as projection]
\label{lem:det-category-projection}
Fix an intermediate score $S:\calX\to[0,1]$ and a measurable assignment
$J:\calX\to\mathcal J$ of contexts to a finite set of categories $\mathcal J$.
Set
\[
  p_J(X)=\E[S(X)\mid J]
  \qquad\text{and}\qquad
  R_J=S-p_J.
\]
Then
\begin{equation}
  \E[R_J\mid J]=0,
  \qquad
  \E[SR_J]=\E[R_J^2].
  \label{eq:det-ideal-category-average}
\end{equation}
Consequently, for every bounded test $g:\calX\to\mathbb R$,
\begin{equation}
  \E[g(X)R_J(X)]
  =\sum_{j\in\mathcal J}\Pp(J=j)
    \Cov(g(X),S(X)\mid J=j).
  \label{eq:det-covariance-target}
\end{equation}
Moreover, if $\mathcal C_J=\{J^{-1}(j):j\in\mathcal J\}$, then
\begin{equation}
  \operatorname{Cal}_{\mathcal C_J}(p_J)
  =
  \max_{\sigma\in\{-1,1\}^{\mathcal J}}
    \left|\E[\sigma_{J(X)}(Y-S(X))]\right|.
  \label{eq:det-category-dual}
\end{equation}
\end{lemma}

\begin{proof}
The first identity is the defining property of conditional expectation.
Since $p_J$ is constant within each category and $\E[R_J\mid J]=0$, we have
$\E[p_JR_J]=\E[p_J\E[R_J\mid J]]=0$.  Therefore,
\[
  \E[SR_J]=\E[(p_J+R_J)R_J]=\E[R_J^2].
\]
Conditioning $\E[gR_J]$ on each category gives
\eqref{eq:det-covariance-target}.  Finally,
$\E[\one\{J=j\}R_J]=0$ for every $j$, so
\[
  \E[\one\{J=j\}(Y-p_J)]
  =
  \E[\one\{J=j\}(Y-S)].
\]
Summing absolute values over $j$ and applying $\ell_1/\ell_\infty$ duality
proves \eqref{eq:det-category-dual}.
\end{proof}

By \eqref{eq:det-covariance-target}, the change in the multiaccuracy residual
for $g$ is the probability-weighted sum of its within-category covariances
with $S$.  Our goal is therefore to choose the categories so that this sum
is small in absolute value for every $g\in\calG$.

If $q$ is constant on every cell of a partition $\mathcal C$, then
\begin{equation}
  \ECE(q)\le\operatorname{Cal}_{\mathcal C}(q).
  \label{eq:det-refinement-dominates-ece}
\end{equation}
ECE groups together all contexts receiving the same prediction.  Since $q$
is constant on each cell, each such group is a union of cells assigned the
same prediction value.  Combining these cells adds their residual
expectations before taking an absolute value.  By the triangle inequality,
this is at most the sum of the absolute residual expectations of the
individual cells.

We call the expectations $\E[\sigma_{J(X)}(Y-S(X))]$, one for each choice
of category signs $\sigma$, the \emph{category-sign residuals}.  The
score-learning algorithm will need to control these residuals for the
assignment rules used in the later stages.

Three different partitions appear later, with different roles.
An \emph{assignment cell} is a set of contexts that shares one random choice
made during training.  Using a finite partition lets the training algorithm
store one seed per cell and thereby define a fixed measurable predictor.  A
\emph{category} is a set $J^{-1}(j)$ on which we
replace $S$ by its average.  A \emph{prediction level set} contains all
contexts receiving a particular prediction value; it may be a union of
several categories.

The proof has two steps.  The population argument chooses a distribution over
category assignments for which the expected signed change in every
multiaccuracy residual is small in absolute value, and the expected squared
change in the score is small.  Both expectations are over the assignments.  The
finite-sample argument applies the same construction to an empirical context
distribution, fixes one assignment per cell, and transfers the resulting
empirical coarsening guarantees to the population.

\subsection{The population assignment argument}

We first control a weighted average of the changes in the multiaccuracy
residuals that come from category averaging.  We will then choose the weights so that this average equals the
largest change among the signed tests.  Including both $g$ and $-g$ lets us
control changes in either direction.  To see how this works, let
$\calH=\calG\cup(-\calG)$, let $\pi$ be a distribution on $\calH$, and write
\[
  f_\pi(x)=\E_{h\sim\pi}[h(x)].
\]
Suppose that, for some $\lambda_{\rm cat}>0$, we could group contexts by the
exact value of $Z=f_\pi(X)+\lambda_{\rm cat}S(X)$ and replace $S$ by
$p^\star=\E[S\mid Z]$.  Fix $\pi$, so $p^\star$ does not depend on which
test $h$ is drawn.  Since $f_\pi(x)=\E_{h\sim\pi}[h(x)]$, averaging the
residual over tests drawn from $\pi$ is the same as testing against $f_\pi$.
Thus
\begin{align*}
  \E_{h\sim\pi}\E[h(X)(S-p^\star)]
  &=\E[f_\pi(X)(S-p^\star)] \\
  &=\E[(Z-\lambda_{\rm cat}S)(S-p^\star)] \\
  &=-\lambda_{\rm cat}\E[(S-p^\star)^2].
\end{align*}
The term containing $Z$ vanishes because
$\E[S-p^\star\mid Z]=0$, while
$\E[S(S-p^\star)]=\E[(S-p^\star)^2]$.  Thus grouping by
$f_\pi+\lambda_{\rm cat}S$ and averaging $S$ creates a negative quadratic term in the
$\pi$-weighted change in the multiaccuracy residuals.

The construction below approximates $f_\pi$ by an average of a small number
of sampled tests and
the exact level sets of $Z$ by $K+1$ categories.  These approximations
introduce an error proportional to the root-mean-square change in the score;
Lemma~\ref{lem:det-population-assignment} makes this tradeoff precise.  We now
define the finite construction.  Fix an intermediate score $S$, a finite
measurable partition $\mathcal U$, and a finite set
$\mathcal B=\{x^{\rm priv}_1,\ldots,x^{\rm priv}_{K_{\rm priv}}\}$ whose
singletons are cells of $\mathcal U$.  Each point in $\mathcal B$ will form a
private category and retain its score; every other cell will independently
choose an assignment rule as described below.  Also fix an integer $K\ge1$, a tuple length
$m_{\rm sp}\ge1$, and a parameter $\lambda_{\rm cat}>0$.

For this finite construction, let $\pi$ again denote a distribution over the
signed tests in $\calH$.  For a tuple
$\tau=(h_1,\ldots,h_{m_{\rm sp}})\in\calH^{m_{\rm sp}}$, define its sparse
average
\[
  f_\tau(x)=\frac1{m_{\rm sp}}\sum_{\ell=1}^{m_{\rm sp}}h_\ell(x).
\]
For a possible score value $s\in[0,1]$, the assignment score
$f_\tau(x)+\lambda_{\rm cat}s$ lies in
$[-1,\lambda_{\rm cat}+1]$.  Let $u_0,\ldots,u_K$ be the uniform grid on this
interval, with spacing
\[
  w=\frac{\lambda_{\rm cat}+2}{K},
\]
and let $j_\tau(x,s)\in\{0,\ldots,K\}$ be the index of the nearest grid point,
with fixed tie-breaking.  We call the categories indexed by $0,\ldots,K$
the \emph{nominal categories}, to distinguish them from the private
singleton categories.  The full category map is
\[
  J_\tau(x,s)
  =
  \begin{cases}
    j_\tau(x,s),&x\notin\mathcal B,\\
    K+i,&x=x^{\rm priv}_i.
  \end{cases}
\]

Scores assigned to the same nominal category, even under different tuples,
are close: if $j_\tau(x,s)=j_{\tau'}(x',s')$, then
\[
  \lambda_{\rm cat}|s-s'|
  \le |f_\tau(x)-f_{\tau'}(x')|+w\le2+w.
\]
Thus every nominal category has score diameter at most
\begin{equation}
  D_{\rm cat}=\frac{2+w}{\lambda_{\rm cat}}.
  \label{eq:det-category-diameter}
\end{equation}
A private category has score diameter zero.

Independently for each nonprivate cell $U\in\mathcal U$, draw
$\tau_U\sim\pi^{m_{\rm sp}}$.  Write $\omega=(\tau_U)_U$ and, for
$x\in U$, set
\[
  f_\omega(x)=f_{\tau_U}(x),
  \qquad
  J_\omega(x)=j_{\tau_U}(x,S(x)).
\]
For $x=x_i^{\rm priv}$, set $J_\omega(x)=K+i$; the value of $f_\omega$ there
is set to zero.  For every category of positive mass, let
$r_{\omega,j}=\E[S(X)\mid J_\omega(X)=j]$.  Set
$r_{\omega,K+i}=S(x_i^{\rm priv})$ for a private category of probability mass
zero, and assign zero to a nominal category of probability mass zero.  Define the category-averaged
predictor globally by
\begin{equation}
  p_\omega(x)=r_{\omega,J_\omega(x)}.
  \label{eq:det-pathwise-category-average}
\end{equation}
This is a version of $\E[S(X)\mid J_\omega]$ and equals
$S(x_i^{\rm priv})$ on a private singleton.  

For a fixed assignment, define the additional residual for test $h$ and the
squared change in the score by
\[
  F_h(\omega)=\E[h(X)(S(X)-p_\omega(X))],
  \qquad
  V(\omega)=\E[(S(X)-p_\omega(X))^2].
\]
Let
\[
  d_h(\pi)=\E_\omega F_h(\omega),
  \qquad
  \overline V(\pi)=\E_\omega V(\omega).
\]
Thus $d_h(\pi)$ is the expected signed change in the multiaccuracy residual
for test $h$, and
$\overline V(\pi)$ is the expected squared change caused by category
averaging.  We call $d$ the
residual-change map.

The next lemma contains the complete population argument.  Its first step
bounds the $\pi$-weighted average of the residual changes.  Its second chooses
$\pi$ supported on tests attaining the largest expected change $d_h(\pi)$,
so this weighted average equals the maximum.

\begin{lemma}[Population category assignment]
\label{lem:det-population-assignment}
There is a distribution $\pi^\star$ on $\calH$ such that
\[
  \max_h d_h(\pi^\star)
  \le
  \frac{e_{\rm app}^2}{4\lambda_{\rm cat}},
  \qquad
  \overline V(\pi^\star)
  \le
  \frac{e_{\rm app}^2}{\lambda_{\rm cat}^2},
  \qquad
  e_{\rm app}=\frac w2+\frac1{\sqrt{m_{\rm sp}}}.
\]
The bounds hold for every context distribution and every fixed choice of
$S$, $\mathcal U$, and $\mathcal B$ in the construction above.
\end{lemma}

\begin{proof}
\smallskip\noindent
Fix any distribution $\pi$ on $\calH$ and let
$f_\pi=\sum_h\pi(h)h$.  On a point in nominal category $j$, write
\[
  e_\omega(x)
  =f_\omega(x)+\lambda_{\rm cat}S(x)-u_j,
  \qquad |e_\omega(x)|\le w/2.
\]
Set $e_\omega=0$ on the private categories, where $S-p_\omega=0$.
The category-projection identities give
$\E[u_{J_\omega}(S-p_\omega)]=0$ (extend $u_j$ by zero on private indices) and
$\E[S(S-p_\omega)]=V(\omega)$.  Since $S-p_\omega=0$ on every private
category,
\begin{align*}
  \E[f_\omega(X)(S-p_\omega)]
  &=
  -\lambda_{\rm cat}V(\omega)
  +\E[e_\omega(X)(S-p_\omega)] \\
  &\le
  \frac w2\sqrt{V(\omega)}-\lambda_{\rm cat}V(\omega).
\end{align*}
For each fixed nonprivate context, $f_\omega(x)$ is the average of
$m_{\rm sp}$ independent draws from $\pi$, so it has mean $f_\pi(x)$ and
variance at most $1/m_{\rm sp}$.  Since $S-p_\omega=0$ on the private
points, Cauchy--Schwarz over the context and assignment randomness gives
\[
  \left|\E_{\omega,X}[(f_\pi(X)-f_\omega(X))
      (S(X)-p_\omega(X))]\right|
  \le m_{\rm sp}^{-1/2}\sqrt{\overline V(\pi)}.
\]
Thus replacing $f_\omega$ by $f_\pi$ in the averaged residual changes it
by at most this amount.  Averaging the preceding residual bound over
$\omega$ and applying Jensen's inequality gives
\begin{equation}
  \sum_h\pi(h)d_h(\pi)
  \le
  e_{\rm app}\sqrt{\overline V(\pi)}
  -\lambda_{\rm cat}\overline V(\pi).
  \label{eq:det-average-residual-change}
\end{equation}

\smallskip\noindent\emph{Worst-test fixed point.}
For each fixed seed vector $\omega$, $F_h(\omega)$ is independent of $\pi$.
Because $\mathcal U$ and $\calH$ are finite, $d_h(\pi)$ is a finite sum of
these fixed residuals weighted by assignment probabilities, each polynomial
in $\pi$.  Hence $d$ is continuous.  Also $d_{-h}(\pi)=-d_h(\pi)$.  Let $\Pi_\Delta$ be Euclidean projection onto the
simplex $\Delta(\calH)$.  The map
\[
  \pi\longmapsto\Pi_\Delta(\pi+d(\pi))
\]
is continuous and maps the simplex to itself, so Brouwer's theorem gives a
fixed point $\pi^\star$.  By the definition of projection, moving from
$\pi^\star$ toward any $q\in\Delta(\calH)$ cannot decrease squared distance
to $\pi^\star+d(\pi^\star)$.  Differentiating along this segment at its
starting point gives
\[
  \langle d(\pi^\star),q-\pi^\star\rangle\le0.
\]
Taking $q$ to be each point mass shows that every coordinate of $d(\pi^\star)$
is at most its $\pi^\star$-weighted average.  Hence
\[
  \max_h d_h(\pi^\star)
  =\sum_h\pi^\star(h)d_h(\pi^\star).
\]
Combining this identity with \eqref{eq:det-average-residual-change} gives
\[
  \max_h d_h(\pi^\star)
  \le
  e_{\rm app}\sqrt{\overline V(\pi^\star)}
  -\lambda_{\rm cat}\overline V(\pi^\star)
  \le
  \frac{e_{\rm app}^2}{4\lambda_{\rm cat}}.
\]
Antisymmetry makes the maximum nonnegative, so
\[
  0
  \le
  e_{\rm app}\sqrt{\overline V(\pi^\star)}
  -\lambda_{\rm cat}\overline V(\pi^\star).
\]
Solving this quadratic inequality in $\sqrt{\overline V(\pi^\star)}$ proves
the variance bound.
\end{proof}

This completes the population argument.  The remaining work is to construct
$S$, $\mathcal U$, and $\mathcal B$ from samples, apply this argument to an
empirical context distribution, and fix the assignment tuples.

We next derive the parameters from the errors they must control.  The next
subsections construct the finite-sample partition and score and then choose
the assignments; here we first record the scales those steps must achieve.
Set
\[
  K=\left\lceil c_K(A_\delta/\eps^2)^{1/3}\right\rceil,
  \qquad
  \alpha=c_\alpha\eps,
  \qquad
  \gamma=\alpha/64,
\]
where $c_K$ is large and $c_\alpha$ is small.  We first consider
$A_\delta\le c/\eps$, which implies $A_\delta\le K$; the complementary range
will follow by applying the finite-test learner directly.  Write
\[
  \xi=\sqrt{\frac{A_\delta}{K^3}}.
\]
The score learner must anticipate every sparse tuple and every sign
assignment to the eventual $O(K)$ categories.  The resulting auxiliary class has
log-cardinality
$m_{\rm sp}\log|\calH|+O(K)$.  Taking
\[
  m_{\rm sp}=\left\lceil\frac K{A_\delta}\right\rceil
\]
keeps this quantity $O(K)$ and gives sparse-average error
$m_{\rm sp}^{-1/2}=O(\sqrt{A_\delta/K})$.  Match the category-grid spacing to
this error as follows.  For a general choice of $\lambda_{\rm cat}$,
\[
  e_{\rm app}
  =O\left(\frac{\lambda_{\rm cat}}K+\sqrt{\frac{A_\delta}K}\right),
  \qquad
  \frac{e_{\rm app}^2}{\lambda_{\rm cat}}
  =O\left(
    \frac{\lambda_{\rm cat}}{K^2}
    +\frac{A_\delta}{K\lambda_{\rm cat}}
  \right).
\]
Balancing these two terms leads to
\[
  \lambda_{\rm cat}=c_{\rm cat}\sqrt{A_\delta K},
  \qquad
  w=\frac{\lambda_{\rm cat}+2}{K}.
\]
Then $e_{\rm app}=w/2+m_{\rm sp}^{-1/2}=O(\sqrt{A_\delta/K})$, and
Lemma~\ref{lem:det-population-assignment} gives a population mixture with
\[
  \frac{e_{\rm app}^2}{\lambda_{\rm cat}}=O(\xi),
  \qquad
  \overline V(\pi^\star)=O(K^{-2}).
\]
The same bounds hold when the context distribution is empirical, which
will let us choose the assignment distribution from samples.

Lemma~\ref{lem:det-category-average-concentration} bounds the deviation of
each additional multiaccuracy residual from its expectation, when fixing
the assignment seeds, by
\[
  O\left(\sqrt{b\overline V(\pi)A_\delta}+bD_{\rm cat}A_\delta\right),
\]
where $b$ is the largest assignment-cell mass.
Since $\overline V(\pi)=O(K^{-2})$ and
$D_{\rm cat}=O((A_\delta K)^{-1/2})$ for the chosen distribution,
both terms are $O(\xi)$ when $b=O(1/K)$.
We therefore give sufficiently frequent contexts private
categories and partition the rest into cells of mass $O(1/K)$.

The finite-sample construction uses three independent samples.  A labeled
preprocessing sample constructs confidence intervals for the conditional
label means and the context partition.  A second labeled sample learns the
intermediate deterministic score.  A context-only sample then supplies both
the assignment distribution and the final category means.

\subsection{Learning a score for the later assignments}

Write $p_x=P_X(\{x\})$ for the probability mass of a context $x$, and call
$x$ an atom of the context distribution when $p_x>0$.
Use a labeled preprocessing sample to construct confidence intervals $I_x$,
allowed grid values $\Lambda_x$, and a finite set $\widehat B$ of frequently
observed contexts.  Each point in $\widehat B$ forms a private singleton
category and keeps its score $S(x)$, so we call $\widehat B$ the
\emph{fixed-score set}.  It is chosen to contain every atom of mass greater
than $C/K$ and has size $O(K)$.

The same sample produces a finite measurable partition $\mathcal U$:
each observed context is a singleton, and the remaining cells are the gaps
between successive observed contexts in lexicographic order, together with the two outer gaps before the first and after
the last observed context.  A singleton
from $\widehat B$ is excluded from the random assignment; every other cell is
an \emph{assignment cell}.  All contexts in an assignment cell use the same
sampled tuple of tests.  The confidence intervals and partition also provide
a squared-mass bound for controlling all category-sign tests at once.

Call a context \emph{observed} if it appears in the preprocessing sample, and call
a partition cell \emph{unobserved} if it contains no observed context.  The next
lemma records all properties of this preprocessing that the proof
uses.  Its proof, including the construction of $I_x$, $\Lambda_x$, and
$\mathcal U$, is in Appendix~\ref{app:det-upper}.  Let $\rho_x$ be the confidence
radius used to construct $I_x$ before clipping to $[0,1]$, and fix a measurable
version $\mu(x)=\E[Y\mid X=x]$ with values in $[0,1]$.

\begin{lemma}[Finite preprocessing]
\label{lem:det-preprocessing}
Set
\[
  \ell=C_\ell\log\frac{C_\ell(A_\delta+K)}{\alpha\delta},
  \qquad
  N=\left\lceil C_N\frac{(A_\delta+K)\ell}{\alpha^2}\right\rceil
\]
for sufficiently large universal constants $C_\ell,C_N$.
Using $N$ labeled preprocessing examples, with probability at least
$1-\delta/10$ all of the following hold simultaneously.
\begin{enumerate}[label=(\roman*),leftmargin=2em]
\item The confidence intervals contain the conditional mean pointwise:
$\mu(x)\in I_x$ for every $x\in\calX$.  Moreover, every $\Lambda_x$ is a
nonempty subset of one common finite grid $\Lambda\subseteq[0,1]$, every
$u\in I_x$ lies within distance $\gamma$ of $\Lambda_x$, and every
$v\in\Lambda_x$ lies within distance $\gamma$ of $I_x$.
\item The fixed-score set has size
\begin{equation}
  |\widehat B|=O(K),
  \label{eq:det-list-sizes}
\end{equation}
and contains every atom of mass greater than $C/K$.
\item The partition satisfies
\begin{equation}
  \max_{U\text{ assignment cell}}P_X(U)=O(1/K),
  \label{eq:det-unit-caps}
\end{equation}
and its squared-mass budget obeys
\begin{equation}
  V_{\mathcal U}
  :={}
  \sum_{\substack{x\text{ observed}\\p_x>0}}p_x^2\rho_x^2
  +
  \sum_{U\text{ unobserved cell}}P_X(U)^2
  \le
  C\frac{\ell}{N}
  \le C\frac{\alpha^2}{A_\delta+K}.
  \label{eq:det-confidence-budget}
\end{equation}
\end{enumerate}
\end{lemma}

The two terms in $V_{\mathcal U}$ control the effects of fixing assignments.
For a score restricted to $\Lambda_x$, its value and the conditional mean at
an observed singleton are within
$2\rho_x+\gamma$ of one another; on an unobserved cell their difference is
at most one.  The contribution of a cell $U$ to any signed $Y-S$ residual
has absolute value at most $\int_U|\mu-S|\,dP_X$.  Summing the squares of
these bounds gives $O(V_{\mathcal U})$ plus a small grid-tolerance term.
The same budget also
controls the earlier cellwise rounding that makes $S$ deterministic.

The next lemma states the consequence of the deterministic finite-test
algorithm that we use.  It packages two steps from
\citet{NoarovRoth2026Deterministic}: learning a randomized score from
intervals known to contain the conditional mean and then fixing one sampler
seed per partition cell.

\begin{lemma}[Deterministic finite-test score learning]
\label{lem:det-finite-test-score}
Fix a finite grid $\Lambda\subseteq[0,1]$, nonempty closed intervals
$(I_x)_{x\in\calX}$, nonempty allowed sets $\Lambda_x\subseteq\Lambda$, and a
finite partition $\mathcal U$ produced by the preprocessing construction.
Suppose $\mu(x)\in I_x$ for every $x$, every point of $I_x$ lies within
$\gamma$ of $\Lambda_x$, and every point of $\Lambda_x$ lies within $\gamma$
of $I_x$.  Let $V_{\mathcal U}$ be the quantity in
\eqref{eq:det-confidence-budget}.
For every finite nonempty family $\Phi$ of measurable tests
$\phi:\calX\times[0,1]\to[-1,1]$ and every $\beta\in(0,1)$, an
algorithm using $N$ fresh labeled examples and training randomness outputs a
deterministic score $S$ with $S(x)\in\Lambda_x$ for every $x$ and, with
probability at least $1-\beta$,
\begin{equation}
  \max_{\phi\in\Phi}
  \left|\E[\phi(X,S(X))(Y-S(X))]\right|
  \le
  C\left(
    \gamma
    +\sqrt{\frac{\log(2|\Phi|/\beta)}{N}}
    +\sqrt{V_{\mathcal U}\log\frac{2|\Phi|}{\beta}}
  \right).
  \label{eq:det-finite-test-score}
\end{equation}
\end{lemma}

\begin{proof}
Condition on the preprocessing transcript, so the grid, intervals, allowed
sets, partition, and test family are fixed.  Lemma~8.1 of
\citet{NoarovRoth2026Deterministic}, applied to $\Phi$ with failure probability
$\beta/2$, returns a randomized grid predictor $Q$ supported on $\Lambda_x$
whose residuals are at most the first two terms in
\eqref{eq:det-finite-test-score}.  Conditional also on this predictor, apply
their Proposition~6.1 to the same intervals and partition, again with failure
probability $\beta/2$.  Its one-seed-per-cell rounding produces a
deterministic $S$ and adds at most the last square-root term and another
$O(\gamma)$.  A union bound proves the stated probability, and reversing
$S-Y$ to $Y-S$ does not change the absolute residual.
\end{proof}

The cellwise category assignment will be chosen only after $S$ is learned.
We first bound its category-sign residual in expectation over the assignment
randomness, and later use concentration to fix the random choices.  Order the
fixed-score set lexicographically as
$\widehat B=\{x^{\rm priv}_1,\ldots,x^{\rm priv}_{K_{\rm priv}}\}$, where
$K_{\rm priv}=|\widehat B|=O(K)$, and instantiate the full single-tuple map
$J_\tau$ above with $\mathcal B=\widehat B$.  The private categories introduce
no assignment randomness and retain the score $S(x^{\rm priv}_i)$.
Every assignment cell draws its tuple from the same distribution.  For a
fixed context, its category therefore has the same distribution as if one
tuple were drawn and used everywhere.  Averaging over contexts shows that
the expected category-sign residual is an average of the residuals for
these global rules.  It suffices to control each global rule while learning
$S$; concentration will later control the realized cellwise assignments.  For
every $\sigma\in\{-1,1\}^{K+1+K_{\rm priv}}$, include the test
\begin{equation}
  a_{\tau,\sigma}(x,s)
  =
  \sigma_{J_\tau(x,s)}.
  \label{eq:det-category-test}
\end{equation}
The logarithm of the number of these tests is
\[
  m_{\rm sp}\log|\calH|+(K+1+K_{\rm priv})\log2=O(K).
\]
Let
\[
  \Phi_K
  =
  \calG\cup
  \{a_{\tau,\sigma}:\tau\in\calH^{m_{\rm sp}},
      \ \sigma\in\{-1,1\}^{K+1+K_{\rm priv}}\}
\]
be the full finite residual-test family, where each $g\in\calG$ is viewed as
the prediction-independent test $(x,s)\mapsto g(x)$.  Since
$A_\delta\le K$, its
log-cardinality is $O(K)$.  Conditional on the preprocessing
sample, Lemma~\ref{lem:det-finite-test-score} applied to $\Phi_K$ restricts
each $S(x)$ to $\Lambda_x$ and returns a deterministic score
$S:\calX\to[0,1]$ satisfying
\begin{align}
  \max_{g\in\calG}
  |\E[g(X)(Y-S(X))]|&\le C\alpha,
  \label{eq:det-score-ma}\\
  \max_{\tau,\sigma}
  |\E[a_{\tau,\sigma}(X,S(X))(Y-S(X))]|&\le C\alpha.
  \label{eq:det-score-signs}
\end{align}

\subsection{Choosing assignments on the empirical distribution}

Once preprocessing and score learning are complete, the score $S$, partition
$\mathcal U$, and fixed-score set $\widehat B$ are fixed.  Draw a fresh
context-only sample $Z_1,\ldots,Z_N$.  Write $P_X[f]=\E[f(X)]$ and
\[
  \widehat P_X[f]=\frac1N\sum_{i=1}^N f(Z_i).
\]
We apply the population construction directly to this empirical distribution.
For a seed vector $\omega$, let $\widehat r_{\omega,j}$ be the empirical
average of $S$ in nominal category $j$, using zero if no sample point falls
in that category.
Let $\widehat p_\omega$ predict this average on
each nominal category and retain $S$ on $\widehat B$.  Define
\[
  \widehat F_h(\omega)=\widehat P_X[h(S-\widehat p_\omega)],
  \qquad
  \widehat V(\omega)=\widehat P_X[(S-\widehat p_\omega)^2],
\]
and average over independent cell tuples with law $\pi^{m_{\rm sp}}$:
\[
  \widehat d_h(\pi)=\E_\omega\widehat F_h(\omega),
  \qquad
  \widehat{\overline V}(\pi)=\E_\omega\widehat V(\omega).
\]
These are finite sums over all cellwise assignments.  Observations in the
same partition cell share one tuple, just as in the population construction.

\begin{lemma}[Choosing the empirical assignment distribution]
\label{lem:det-law-selection}
For every context sample, the training algorithm can select a distribution
$\widehat\pi$ on $\calH$ such that
\begin{equation}
  \max_{h\in\calH}\widehat d_h(\widehat\pi)\le C\xi,
  \qquad
  \widehat{\overline V}(\widehat\pi)\le C/K^2.
  \label{eq:det-selected-law}
\end{equation}
The selection is information-theoretic.
\end{lemma}

\begin{proof}
Lemma~\ref{lem:det-population-assignment} applies with context distribution
$\widehat P_X$, so some mixture has maximum empirical residual at most
$e_{\rm app}^2/(4\lambda_{\rm cat})$.  Enumerate the rational points of
$\Delta(\calH)$ in a fixed order and accept the first $\widehat\pi$ satisfying
\[
  \max_h\widehat d_h(\widehat\pi)
  <\frac{e_{\rm app}^2}{4\lambda_{\rm cat}}+\xi.
\]
For each fixed assignment, its empirical residual is independent of $\pi$,
while the assignment probability is a polynomial in $\pi$.
Thus $\widehat d$ is continuous, and density of rational points and the
strict slack $\xi$ ensure termination.  Each evaluation is a finite sum;
choosing the first accepted point also makes the selection measurable.

Antisymmetry gives
$\sum_h\widehat\pi(h)\widehat d_h(\widehat\pi)
\ge-\max_h\widehat d_h(\widehat\pi)$.  Write
$\widehat{\overline V}=\widehat{\overline V}(\widehat\pi)$.  Applying
\eqref{eq:det-average-residual-change} under $\widehat P_X$ therefore yields
\[
  \lambda_{\rm cat}\widehat{\overline V}
  \le e_{\rm app}\sqrt{\widehat{\overline V}}
       +\max_h\widehat d_h(\widehat\pi).
\]
Young's inequality gives
$\widehat{\overline V}\le
e_{\rm app}^2/\lambda_{\rm cat}^2+
2\max_h\widehat d_h(\widehat\pi)/\lambda_{\rm cat}=O(K^{-2})$.
\end{proof}

On the preprocessing event, the empirical assignment cells also have small
mass: with probability at least $1-\delta/20$ over the context sample,
\begin{equation}
  \widehat b:=\max_{U\text{ assignment cell}}\widehat P_X(U)=O(1/K).
  \label{eq:det-empirical-cell-cap}
\end{equation}
Appendix~\ref{app:det-upper-transfer} proves this bound together with the
population transfer below.  Conditional on this event, we can apply the
same variance-sensitive concentration argument to the empirical distribution.

\subsection{Fixing the category assignments}

Draw one tuple from $\widehat\pi^{m_{\rm sp}}$ for each assignment cell and
store the resulting seed vector.  We first show that the empirical
multiaccuracy residuals remain small.  Empirical calibration relative to $S$
is exact for every assignment, since each category predicts its empirical
average.  The next subsection transfers both guarantees to the true labels.

The population argument applied to the sample and the assignment concentration bound
have three rate-critical terms:
\begin{align*}
  \frac{e_{\rm app}^2}{\lambda_{\rm cat}}&=O(\xi)
  &&\text{(expected empirical residual)},\\
  \sqrt{\widehat b\,\widehat{\overline V}(\widehat\pi)A_\delta}
    &=O(\xi)
  &&\text{(fixing seeds: variance term)},\\
  \widehat bD_{\rm cat}A_\delta&=O(\xi)
  &&\text{(fixing seeds: scale term)}.
\end{align*}
Here $\widehat b=O(1/K)$ and
$\widehat{\overline V}(\widehat\pi)=O(K^{-2})$.
Consequently,
\[
  \xi=\sqrt{A_\delta/K^3}\lesssim\eps
  \qquad\Longleftrightarrow\qquad
  K^3\gtrsim A_\delta/\eps^2.
\]
Together with the score-learning cost, this gives the error tradeoff
$O(\sqrt{K/N}+\sqrt{A_\delta/K^3})$, suppressing logarithmic factors.
The transfer to the population will fit within the first term.

\begin{lemma}[Fixing the cellwise assignments]
\label{lem:det-fix-assignments}
Condition on successful preprocessing and score learning, and on a context
sample satisfying \eqref{eq:det-empirical-cell-cap}.  Choose $\widehat\pi$
as in Lemma~\ref{lem:det-law-selection}.  With probability at least $1-\delta/10$
over the independent tuples drawn for the assignment cells,
\begin{equation}
  \max_{g\in\calG}|\widehat F_g(\omega)|\le C\xi.
  \label{eq:det-fixed-multiaccuracy-residuals}
\end{equation}
\end{lemma}

The proof is in Appendix~\ref{app:det-upper-seeds}.  We use
Lemma~\ref{lem:det-category-average-concentration} on the empirical
distribution.  Deleting an assignment cell changes the total covariance by
an amount whose square is bounded by a constant times the cell's mass
times the integral of $(S-\widehat p_\omega)^2$ over that cell.
A moment form of the Efron--Stein inequality
gives the two seed-fixing terms above.

\subsection{Transferring the empirical guarantees to the true labels}

The empirical category means use the same sample that selected
$\widehat\pi$, and the categories depend on the assignment seeds.  We need a
transfer bound that allows both dependencies.  The next lemma compares the
final predictor's population residuals against $Y$ directly with its
empirical residuals against $S$.  The auxiliary category-sign tests learned
with $S$ are what make this comparison possible for calibration.

\begin{lemma}[Population transfer for the selected coarsening]
\label{lem:det-coarsening-transfer}
Condition on successful preprocessing and score learning.  Let
$\widehat\pi$ be any measurable choice based on the fresh context sample.
Conditional on this sample, independently draw one tuple from
$\widehat\pi^{m_{\rm sp}}$ per assignment cell, and let $\widehat p_\omega$
use the empirical category means defined above.  With probability at least
$1-\delta/10$ over this sample and these tuples, both
\begin{align}
  \max_{g\in\calG}
    \left|\E[g(X)(Y-\widehat p_\omega(X))]
      -\widehat P_X[g(S-\widehat p_\omega)]\right|
    &\le C\alpha,\nonumber\\
  \max_{\sigma\in\{\pm1\}^{K+1+K_{\rm priv}}}
    \left|\E[\sigma_{J_\omega(X)}(Y-\widehat p_\omega(X))]
      -\widehat P_X[\sigma_{J_\omega}(S-\widehat p_\omega)]\right|
    &\le C\alpha
  \label{eq:det-coarsening-transfer}
\end{align}
hold.
\end{lemma}

The proof in Appendix~\ref{app:det-upper-transfer} uses the same finite-menu
principle as score learning.  For each global tuple, the score-learning
guarantee controls the $Y-S$ residual, and sample concentration controls the
remaining empirical error.  Averaging over global tuples bounds the expected
error for any selected law.  Independent cell assignments then concentrate
around this expectation, using the confidence intervals and squared cell
masses.  For fixed assignments and a fixed test, the residual difference
is affine in the category prediction values.  Its absolute value is
therefore maximized at a vertex of $[0,1]^{K+1}$.  Checking every assignment
of prediction values in $\{0,1\}$ to the nominal categories thus covers all
values in $[0,1]$, including the empirical means.  The context sample needs
no labels: $Y$ appears only in the
population quantity being bounded.

We can now finish both error bounds.  For multiaccuracy, the first transfer
bound and Lemma~\ref{lem:det-fix-assignments} give
\[
  \MA(\widehat p_\omega;\calG)
  \le \max_g|\widehat F_g(\omega)|+C\alpha
  \le C(\alpha+\xi).
\]
For calibration, every category has zero empirical coarsening residual:
\[
  \widehat P_X[\one\{J_\omega=j\}(S-\widehat p_\omega)]=0.
\]
This includes empty empirical categories and private singletons.
The second transfer bound therefore gives
$\operatorname{Cal}_{\mathcal C_\omega}(\widehat p_\omega)\le C\alpha$,
where $\mathcal C_\omega$ is the partition into the nominal categories and
private singletons.
Merging categories with equal numerical predictions can only decrease this
error, so \eqref{eq:det-refinement-dominates-ece} also gives
$\ECE(\widehat p_\omega)\le C\alpha$.

\begin{algorithm}[H]
\caption{The deterministic construction for $A_\delta\le c/\eps$}
\label{alg:deterministic-cma}
\small
\begin{enumerate}[leftmargin=*,label=\textbf{\arabic*.},nosep]
\item Given $\eps$, $\delta$, and $\calG$ with $A_\delta\le c/\eps$, use the
parameter choices above and set $L=A_\delta+K$,
$\ell=C_\ell\log(C_\ell L/(\alpha\delta))$, and
$N=\lceil C_NL\ell/\alpha^2\rceil$.
\item Use one labeled preprocessing sample to construct $I_x$, $\Lambda_x$,
  $\widehat B$ and the partition $\mathcal U$ as in
  Lemma~\ref{lem:det-preprocessing}.
\item Form $\Phi_K$ and apply Lemma~\ref{lem:det-finite-test-score}, with
failure probability $\delta/10$, on a fresh labeled sample to
  learn the deterministic intermediate score $S$.
\item On a fresh context-only selection sample, construct the empirical
  residual-change map $\widehat d$ and use
  Lemma~\ref{lem:det-law-selection} to choose a mixture $\widehat\pi$
  satisfying the empirical residual bound \eqref{eq:det-selected-law}.
\item Independently for each assignment cell $U$, draw and store
  $\tau_U\sim\widehat\pi^{m_{\rm sp}}$, and use $J_{\tau_U}(x,S(x))$ to assign its contexts
  to nominal categories.  Give each point in $\widehat B$ a private category
  and keep its score unchanged.  Lemma~\ref{lem:det-fix-assignments} shows
  that coarsening introduces only small empirical multiaccuracy residuals.
\item Using the same context sample as in step 4, compute the empirical
  mean of $S$ in each nominal category; if no sample point falls in a
  category, use zero.
  Keep the private predictions equal to their scores.
  Output $\widehat p_\omega$, which has at most $K+1+K_{\rm priv}=O(K)$
  distinct prediction values.  Lemma~\ref{lem:det-coarsening-transfer}
  transfers its empirical guarantees relative to $S$ to the population
  guarantees relative to $Y$.
\end{enumerate}
\end{algorithm}

Each sample split has size $N$.  The cell seeds inside
Lemma~\ref{lem:det-finite-test-score} make $S$ deterministic.  After $S$ and
$\widehat\pi$ are fixed, fresh training randomness draws the independent
assignment tuples $\tau_U$.
Each stage succeeds with high probability, so the algorithm uses a
single run.

\begin{proof}[Proof of Theorem~\ref{thm:det-upper}]
In the range $A_\delta\le c/\eps$, choose
$K=\Theta((A_\delta/\eps^2)^{1/3})$ and
$\alpha=\Theta(\eps)$ as above.  Then
\[
  \sqrt{A_\delta/K^3}=O(\eps).
\]
The preceding construction uses a constant number of independent samples,
each of size
\[
  \widetilde O\left(\frac{A_\delta+K}{\eps^2}\right)
  =
  \widetilde O\left(
    \frac{A_\delta}{\eps^2}
    +
    \frac{A_\delta^{1/3}}{\eps^{8/3}}
  \right).
\]
The preprocessing, score-learning, empirical cell-mass,
assignment-fixing, and population-transfer events have failure probabilities
at most $\delta/10,\delta/10,\delta/20,\delta/10,\delta/10$, respectively.
Empirical assignment selection succeeds for every context sample.
A union bound therefore gives success probability at least $1-\delta$.

For $A_\delta>c/\eps$, the finite-test learner suffices without
postprocessing.  Use the common uniform grid
$\Lambda=\{i/N_{\rm grid}:0\le i\le N_{\rm grid}\}$ with
$N_{\rm grid}=\lceil1/\gamma\rceil$ and $\gamma=\alpha/64$.
For each $\sigma\in\{\pm1\}^{\Lambda}$, let
$a_\sigma(x,s)=\sigma(s)$ on this grid, extended by zero elsewhere.
Take
\[
  \Phi=\calG\cup\{a_\sigma:\sigma\in\{\pm1\}^{\Lambda}\},
  \qquad
  \log(2|\Phi|/\delta)=O(A_\delta+1/\eps).
\]
Here a feature test $g\in\calG$ is regarded as the function $(x,s)\mapsto g(x)$.
Construct the same confidence intervals and singleton-and-gap partition,
using $L=A_\delta+|\Lambda|$ in the sample-size formula of
Algorithm~\ref{alg:deterministic-cma}.  The coverage and
squared-mass proof in Appendix~\ref{app:det-upper} does not use the
fixed-score set: it gives
$V_{\mathcal U}=O(\alpha^2/L)$ with probability at least $1-\delta/10$.
Apply Lemma~\ref{lem:det-finite-test-score} to $\Phi$ with failure
probability $\delta/10$ and $N$ fresh labeled examples.
It returns a deterministic score $S$ with every test residual $O(\alpha)$.
The original tests bound MA, while
\[
  \ECE(S)
  =\max_{\sigma\in\{\pm1\}^{\Lambda}}
    |\E[\sigma(S(X))(Y-S(X))]|
\]
bounds ECE by the same amount.  Thus $S$ itself is the final predictor,
and the sample requirement is
\[
  \widetilde O\left(\frac{A_\delta+1/\eps}{\eps^2}\right)
  =\widetilde O(A_\delta/\eps^2)
\]
in this range.  Taking $c_\alpha$ sufficiently small proves the theorem.
\end{proof}

\section{Matching lower bound for randomized predictors}
\label{sec:lower}

The matching randomized lower bound has two independent components.  The
$\log|\calG|/\eps^2$ term is already required by multiaccuracy for a finite
family of tests.  The $\eps^{-5/2}$ term is required by calibration together
with a single multiaccuracy constraint, and therefore persists even when the
bounded test family has constant size.

\begin{theorem}[Matching lower bound for randomized predictors]
\label{thm:lower}
There are universal constants $c,c_0>0$ such that the following holds.  Fix an
integer $N_{\cal G}\ge2$ and a sufficiently small $\eps>0$.  Suppose a
learning algorithm has the following uniform guarantee: for every
finite domain $\calX$, every bounded test family
$\calG\subseteq[-1,1]^{\calX}$ with $|\calG|\le N_{\cal G}$, and every distribution
$\mathcal D$ on $\calX\times[0,1]$, the algorithm receives $n$ i.i.d.
examples and outputs an arbitrary randomized predictor
$Q=(Q_x)_{x\in\calX}$ satisfying
\[
  \Pp\left(
    \CMA_{\mathcal D}(Q;\calG)\le c\eps
  \right)\ge\frac23 .
\]
Then
\[
  n
  \ge
  c_0\left(
    \frac{\log N_{\cal G}}{\eps^2}
    +
    \frac1{\eps^{5/2}}
  \right).
\]
Here the probability is over the sample and all randomness of the learning algorithm.
\end{theorem}

\subsection{The sample cost of recovering hidden offsets}

The two randomized lower-bound terms and the deterministic lower bound in
Section~\ref{sec:det-lower} all use distributions in which independent
hidden bits add small offsets to otherwise fixed label means.  The following lemma
isolates their common sample requirement: changing one bit changes the label
distribution at only one of the $R$ equally likely contexts.

\begin{lemma}[Sample cost of recovering hidden coordinates]
\label{lem:sign-recovery-information}
There is a universal constant $c_{\rm info}>0$ such that the following holds.
Fix either $\mathcal B=\{0,1\}$ or $\mathcal B=\{\pm1\}$.  Let $R\ge1$,
$\eta>0$, and $t_1,\ldots,t_R\in[0,1]$.  For $\theta\in\mathcal B^R$, let
$\mathcal D_\theta$ have uniform marginal on $[R]$ and Bernoulli label means
\[
  \mu_i^\theta=t_i+\eta\theta_i,
\]
all lying in $[1/8,7/8]$.  Suppose $n$ i.i.d. samples and an independent
random seed determine an estimate $\widehat\theta\in\mathcal B^R$ such that,
for every $\theta$,
\[
  \Pp\bigl(d_H(\widehat\theta,\theta)\le0.2R\bigr)\ge\frac23,
\]
where $d_H$ denotes Hamming distance.  Then
\[
  n\ge c_{\rm info}\frac{R}{\eta^2}.
\]
\end{lemma}

The proof compares two hidden vectors that differ in one coordinate.
With fewer than order $R/\eta^2$ samples, these two sample distributions
are too similar to distinguish their bits reliably.  Averaging this
comparison over coordinates proves the lemma; the calculation is in
Appendix~\ref{app:lower-deferred-proofs}.

We establish the two lower-bound terms separately.  The finite-class
multiaccuracy term is the finite-cardinality specialization of the lower
bound of \citet{GibbsTibshirani2025}.

\begin{lemma}[Finite-class multiaccuracy lower bound \citep{GibbsTibshirani2025}]
\label{lem:randomized-group-lower}
There are universal constants $c,c_0>0$ such that the following holds.  For
every integer $N_{\cal G}\ge2$ and every sufficiently small $\eps>0$, there
are a finite domain $\calX_0$, a bounded test family
$\calG_0\subseteq\{\pm1\}^{\calX_0}$ with $|\calG_0|\le N_{\cal G}$, and a finite
family $\mathfrak D_0$ of distributions on $\calX_0\times\{0,1\}$ such that
the following is true.  Let a learning algorithm receive
$n$ i.i.d. examples and output an arbitrary randomized predictor
$Q=(Q_x)_{x\in\calX_0}$.  If, for every $\mathcal D\in\mathfrak D_0$,
\[
  \Pp\left(
    \CMA_{\mathcal D}(Q;\calG_0)\le c\eps
  \right)\ge\frac23,
\]
then
\[
  n\ge c_0\frac{\log N_{\cal G}}{\eps^2}.
\]
\end{lemma}

The proof uses sign tests on a domain of size $\Theta(\log N_{\cal G})$,
chosen so that multiaccuracy bounds the average absolute difference between
$\E[P\mid X=x]$ and $\E[Y\mid X=x]$.  Thresholding the prediction means recovers the hidden
signs.  Lemma~\ref{lem:sign-recovery-information} gives the sample
bound; the details appear in Appendix~\ref{app:lower-deferred-proofs}.

We now turn to the $\eps^{-5/2}$ term, using the one-dimensional hard
family from Proposition~2 of \citet{GibbsTibshirani2025}.  The bounded test
family is the singleton $\calG=\{g\}$ with $g(x)=x$.  We extend their
deterministic-predictor lower bound by recovering hidden bits from an
arbitrary randomized predictor.

The construction uses ordered points with known baseline means, each shifted
by a small offset determined by a hidden bit.  Calibration and the single multiaccuracy constraint
first force realized predictions close to their baselines.  We assign each
prediction to its nearest baseline value; most predictions are then assigned to
the correct input index.  Calibration on these bins, after subtracting the
known baseline contributions, reveals most hidden bits.

Fix an integer $k\ge2$ and let
\[
  x_i=\frac{i}{k},\qquad b_i=\frac14+\frac{x_i}{2},
  \qquad i=1,\ldots,k.
\]
For $\theta\in\{0,1\}^k$ and a parameter $\Delta>0$, define the distribution
$\mathcal D_\theta$ by drawing $I\sim\Unif\{1,\ldots,k\}$, setting $X=x_I$, and
drawing
\[
  Y\mid I=i \sim \Ber(\mu_i^\theta),
  \qquad
  \mu_i^\theta=b_i+\Delta \theta_i.
\]
We will choose $\Delta=a/k^2$ for a sufficiently small universal constant
$a>0$, so all means lie in a fixed subinterval of $(0,1)$.

For a randomized predictor $Q$ and a fixed vector $\theta$, consider the joint
experiment in which $(X,Y)\sim\mathcal D_\theta$ and then $P\sim Q_X$ is drawn
independently of $Y$ conditional on $X$.  All expectations and conditional
expectations in the next equation are with respect to this joint law.  Write
\[
  \CMA_\theta(Q):=\max\left\{
    \left|\E[X(Y-P)]\right|,
    \E\left|\E[Y-P\mid P]\right|
  \right\}.
\]
This is exactly the calibrated multiaccuracy
error under $\mathcal D_\theta$ for the singleton bounded test $g(x)=x$.

The next lemma gives a decoder that depends only on the full predictor
$Q$ and the public construction.

\begin{lemma}[Recovering hidden bits from a randomized predictor]
\label{lem:randomized-decoding}
There are universal constants $a,c>0$ such that, for $\Delta=a/k^2$,
there is a decoder depending only on $Q$, the public baselines, and
$\Delta$ whose estimate $\widehat\theta$ satisfies
\[
  \CMA_\theta(Q)\le c/k^2
  \quad\Longrightarrow\quad
  d_H(\widehat\theta,\theta)\le k/5
\]
for every $\theta\in\{0,1\}^k$ and every randomized predictor $Q$.
\end{lemma}

\begin{proof}
Fix $Q$, and write $\nu=\CMA_\theta(Q)$.  All expectations are over a fresh
test point and a prediction drawn from this fixed predictor.
Let $J(p)$ be the index of the baseline $b_j$ nearest to $p$, breaking ties
by the smaller index.  Define
\[
  z_j=\frac{k}{\Delta}\E[\one\{J(P)=j\}(P-b_I)],
  \qquad
  \widehat\theta_j=\one\{z_j\ge1/2\}.
\]
The joint law of $(I,P)$ is determined by $Q$ and the known uniform context
distribution, so this decoder uses no labels or hidden bits.  It subtracts
the baseline contribution in each bin before estimating the small offset.

\paragraph{Most predictions identify the correct baseline.}
Calibration controls both $\E[P(Y-P)]$ and $\E[Y-P]$ in absolute value by
$\nu$, while multiaccuracy controls $\E[X(Y-P)]$.  Since
$b_I=1/4+X/2$, it follows that
\[
  |\E[(P-b_I)(Y-P)]|\le\frac74\nu.
\]
Put $T=\E[(P-b_I)^2]$.  Conditional independence of $P$ and $Y$ given $I$
and the formula for $\mu_I^\theta$ give
\[
  \E[(P-b_I)(Y-P)]
  =-T+\Delta\E[(P-b_I)\theta_I].
\]
Cauchy--Schwarz and $0\le\theta_I\le1$ imply
\[
  T\le\frac74\nu+\Delta\sqrt T
  \le\frac74\nu+\frac T2+\frac{\Delta^2}{2},
  \qquad
  T\le\frac72\nu+\Delta^2.
\]
Distinct baselines are separated by $1/(2k)$, so $J(P)\ne I$ requires
$|P-b_I|\ge1/(4k)$.  Hence
\[
  q:=\Pp(J(P)\ne I)\le16k^2T.
\]

\paragraph{Calibration reveals the offsets in these bins.}
For the analysis, let
\[
  c_j=\E[\one\{J(P)=j\}(Y-P)].
\]
Each bin is an event determined by $P$, so
$c_j=\E[\one\{J(P)=j\}\E[Y-P\mid P]]$.
The bins partition the prediction values, and the triangle inequality gives
$\sum_j|c_j|\le\E|\E[Y-P\mid P]|\le\nu$.
The label-mean formula gives
\[
  z_j=k\E[\one\{J(P)=j\}\theta_I]-\frac{k}{\Delta}c_j.
\]
If every prediction were assigned correctly, the first term would be
$\theta_j$.  In general, moving a prediction to the wrong bin changes the
vector of bin contributions in at most two coordinates.  Formally,
\[
  \sum_j\left|
    \E[\one\{J(P)=j\}\theta_I]-\frac{\theta_j}{k}
  \right|
  \le
  \E\sum_j
    |(\one\{J(P)=j\}-\one\{I=j\})\theta_I|
  \le2q.
\]
Therefore
\[
  \frac1k\sum_j|z_j-\theta_j|\le2q+\frac{\nu}{\Delta}.
\]
Every incorrectly decoded bit contributes at least $1/2$ to this absolute
error.  Combining the bounds gives
\[
  \frac1k d_H(\widehat\theta,\theta)
  \le4q+\frac{2\nu}{\Delta}
  \le224k^2\nu+64k^2\Delta^2+\frac{2\nu}{\Delta}.
\]
For $\Delta=a/k^2$, $\nu\le c/k^2$, and $k\ge2$, the last expression is at
most $224c+16a^2+2c/a$.  Choose $a\le1/4$ sufficiently small, then $c>0$
sufficiently small, to make this at most $1/5$.
\end{proof}

Combining this decoder with Lemma~\ref{lem:sign-recovery-information} gives
the calibration-driven term in the sample lower bound.

\begin{lemma}[Calibration lower bound for randomized predictors]
\label{lem:randomized-calibration-lower}
There are universal constants $c,c_0>0$ such that the following holds for
every sufficiently small $\eps>0$.  There is a finite domain
$\calX_\eps\subseteq[0,1]$ such that, for the singleton bounded test family
$\calG=\{g\}$ with $g(x)=x$ on $\calX_\eps$, any learning
algorithm with the following guarantee must use at least $c_0\eps^{-5/2}$
samples: for every distribution $\mathcal D$ on $\calX_\eps\times\{0,1\}$, the
algorithm outputs an arbitrary randomized predictor
$Q=(Q_x)_{x\in\calX_\eps}$ satisfying
\[
  \Pp\left(
    \CMA_{\mathcal D}(Q;\calG)\le c\eps
  \right)\ge\frac23 .
\]
\end{lemma}

Take $k=\Theta(\eps^{-1/2})$ and $\Delta=\Theta(k^{-2})$.
Lemma~\ref{lem:randomized-decoding} turns a successful predictor into a
bit vector that is correct on at least $0.8k$ coordinates.  Lemma~\ref{lem:sign-recovery-information}
then requires $\Omega(k/\Delta^2)=\Omega(\eps^{-5/2})$ samples.  The parameter
choices are recorded in Appendix~\ref{app:lower-deferred-proofs}.

The lemma uses the single bounded test $g(x)=x$.  It also yields a version
using only indicator tests, at the cost of a polynomial-size test family.  On
the grid above,
\[
  x_i
  =
  \frac1k\sum_{j=1}^k\one\{x_i\ge x_j\}.
\]
Consequently, multiaccuracy for the $k$ threshold tests on the right controls
multiaccuracy for $g(x)=x$.  Since the lower-bound choice is
$k=\Theta(\eps^{-1/2})$, the same $\eps^{-5/2}$ obstruction holds for a
polynomial-size family of indicator tests, while the constant-size statement
uses the bounded test formulation.

\begin{proof}[Proof of Theorem~\ref{thm:lower}]
Let $c$ be no larger than the accuracy constants in
Lemmas~\ref{lem:randomized-group-lower} and
\ref{lem:randomized-calibration-lower}.  A learning algorithm satisfying the guarantee of
Theorem~\ref{thm:lower} also satisfies the hypotheses of
Lemma~\ref{lem:randomized-calibration-lower}, because the singleton bounded test
family has size at most $N_{\cal G}$.  Therefore
$n\ge c_1\eps^{-5/2}$ for a universal $c_1>0$.

The same learning algorithm also satisfies the hypotheses of
Lemma~\ref{lem:randomized-group-lower}, so
$n\ge c_2\log(N_{\cal G})/\eps^2$ for a universal $c_2>0$.  Hence
\[
  n
  \ge
  c_4\max\left\{\frac{\log N_{\cal G}}{\eps^2},\frac1{\eps^{5/2}}\right\}
  \ge
  \frac{c_4}{2}
  \left(
    \frac{\log N_{\cal G}}{\eps^2}
    +
    \frac1{\eps^{5/2}}
  \right),
\]
which proves the theorem after renaming the constant.
\end{proof}

\section{Lower bound for deterministic predictors}
\label{sec:det-lower}

At the level of the hard distributions, this construction resembles the
randomized lower bound: the domain is ordered, and the conditional mean is a
public baseline plus a small hidden perturbation.  The decoding mechanism is
different.  The randomized proof uses the identity test $g(x)=x$, together with
calibration, to place realized predictions close to the baseline at a
relatively coarse scale.  Here a larger
family of sign tests forces a deterministic predictor $p$ close to the public
baseline at a finer scale.  Because $p$ is a fixed function, its prediction
values create exact level sets $\{x:p(x)=v\}$ that partition the domain.

The proof uses those level sets to recover one hidden sign per domain point.
Multiaccuracy against the sign tests forces $p$ to have small average distance
from the known baseline.  Since the baseline values are separated, most points
then have distinct prediction values and hence singleton level sets.
Calibration on a singleton directly reveals the corresponding hidden sign.  An
application of the shared recovery bound,
Lemma~\ref{lem:sign-recovery-information}, converts recovery of many signs
into the sample lower bound.

\begin{theorem}[Deterministic lower bound with a group budget]
\label{thm:det-budget-lower}
There are universal constants
$c,c_{\rm size},c_{\rm samp},C_0,a_0>0$ such that the following holds for
every sufficiently small $\eps>0$.  Let $A$ be a target budget for
$\log|\calG|$
satisfying
\[
  C_0\log(1/\eps)\le A\le a_0\eps^{-1}.
\]
Then there exist a finite domain $\calX_{\eps,A}$, a finite family
$\mathfrak D_{\eps,A}$ of distributions on
$\calX_{\eps,A}\times\{0,1\}$, and a finite bounded test family
$\calG_{\eps,A}\subseteq\{\pm1\}^{\calX_{\eps,A}}$ with
\[
  c_{\rm size}A\le \log|\calG_{\eps,A}|\le A
\]
with the following property.  If a learning algorithm
receives $n$ i.i.d. examples, outputs a deterministic predictor
$p:\calX_{\eps,A}\to[0,1]$, and, for every
$\mathcal D\in\mathfrak D_{\eps,A}$, guarantees
\[
  \Pp\left(
    \CMA_{\mathcal D}(p;\calG_{\eps,A})\le c\eps
  \right)\ge\frac23,
\]
then
\[
  n\ge c_{\rm samp}A^{1/3}\eps^{-8/3}.
\]
\end{theorem}

\begin{figure}[H]
\centering
\begin{tikzpicture}[x=1.08cm,y=1cm,>=Stealth,font=\small]
  \draw[->,thick] (0.1,0.35)--(10.75,0.35)
    node[right,font=\scriptsize]{context $i$};
  \draw[->,thick] (0.1,0.35)--(0.1,3.55)
    node[above,font=\scriptsize]{value};

  \draw[blue!70!black,line width=1pt] (0.55,0.72)--(10.25,2.92);
  \foreach \x/\y in {0.8/0.82,2.1/1.10,3.4/1.38,4.7/1.66,
                       6.0/1.94,7.3/2.22,8.6/2.50,9.9/2.78}{
    \fill[blue!70!black] (\x,\y) circle (1.5pt);
  }
  \node[blue!70!black,anchor=west,font=\scriptsize] at (10.15,2.9)
    {baseline $t_i$};

  \foreach \x/\yb/\yp in {0.8/0.82/0.88,2.1/1.10/1.15,
                            3.4/1.38/1.32,4.7/1.66/1.72,
                            6.0/1.94/1.88,8.6/2.50/2.56,
                            9.9/2.78/2.72}{
    \draw[gray!65,line width=.65pt] (\x,\yb)--(\x,\yp);
    \fill[orange!85!black] (\x,\yp) circle (2pt);
  }

  \draw[red!70!black,densely dashed,line width=.9pt] (7.3,2.22)--(7.3,1.15);
  \fill[orange!85!black] (7.3,1.15) circle (2pt);
  \draw[red!70!black,line width=.8pt] (7.3,1.15) circle (3.1pt);
  \draw[gray!65!black,densely dashed,line width=.8pt] (2.1,1.15)--(7.3,1.15);
  \draw (2.1,0.4)--(2.1,0.3) node[below,font=\scriptsize] {$2$};
  \draw (7.3,0.4)--(7.3,0.3) node[below,font=\scriptsize] {$6$};
  \node[font=\scriptsize,fill=white,inner sep=1.5pt] at (4.7,0.98)
    {shared prediction value};
  \node[red!70!black,anchor=west,font=\scriptsize] (bad) at (7.75,1.56)
    {baseline-inaccurate point};
  \draw[->,red!70!black,line width=.7pt] (bad.west)--(7.36,1.23);

  \fill[blue!70!black] (0.75,3.3) circle (1.5pt);
  \node[anchor=west,font=\scriptsize] at (0.9,3.3) {public baselines $t_i$};
  \fill[orange!85!black] (3.55,3.3) circle (2pt);
  \node[anchor=west,font=\scriptsize] at (3.7,3.3)
    {deterministic predictions $p_i$};
\end{tikzpicture}
\caption{Most prediction values $p_i$ (orange) lie close to distinct baseline
values $t_i$ (blue), so they form singleton level sets.  The displayed
collision $p_2=p_6$ is possible only because point $6$ is far from its
baseline.  Each collision involving a baseline-accurate point can therefore
be charged to a baseline-inaccurate point.  Calibration on the remaining
singletons makes the sign of $p_i-t_i$ reveal~$\theta_i$.}
\label{fig:det-lower}
\end{figure}
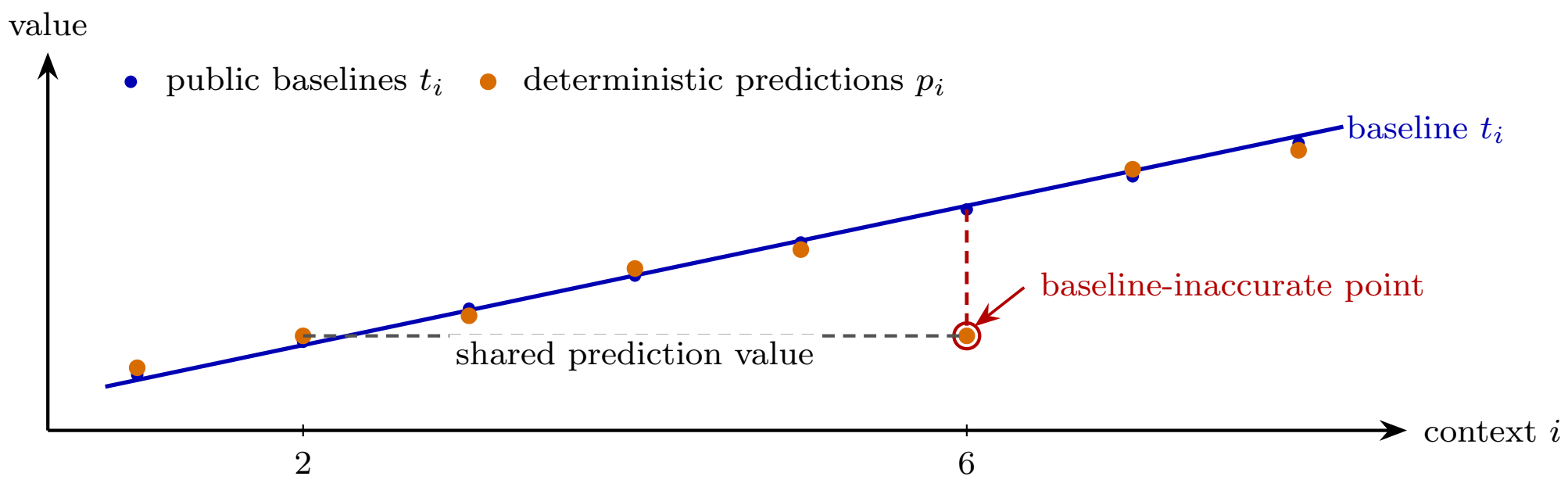

\paragraph{Proof structure.}
The hard family uses $R$ ordered points.  The conditional mean at point $i$ is
a known baseline $t_i$ plus a hidden offset $\eps\theta_i$, where
$\theta_i\in\{\pm1\}$.
Figure~\ref{fig:det-lower} illustrates the counting step that turns closeness
to the separated baseline into singleton prediction values.

The first lemma isolates the deterministic part of the argument.  It converts
average closeness to the separated baseline into many singleton prediction
values, on which calibration reveals the hidden signs directly.

\begin{lemma}[Decoding signs from a predictor close to separated baselines]
\label{lem:deterministic-decoding}
Let $R\ge2$, let $X$ be uniform on $[R]$, and let $\Delta_t>0$.  Suppose
$t_1,\ldots,t_R$ satisfy $|t_i-t_j|\ge\Delta_t$ whenever $i\ne j$.
Fix $\theta\in\{\pm1\}^R$ and $\eta>0$, and suppose
\[
  \mu_i^\theta=t_i+\eta\theta_i
\]
defines valid Bernoulli means.  Let $\mathcal D_\theta$ be the distribution
with this uniform marginal and these label means.  Let
$p:[R]\to[0,1]$ be deterministic, and set
\[
  \overline e(p,t)=\frac1R\sum_{i=1}^R|p(i)-t_i|.
\]
For every $0<r<\Delta_t/2$, the decoder
$\widehat\theta_i=\sgn(p(i)-t_i)$, with ties broken arbitrarily, satisfies
\begin{equation}
  \frac1R d_H(\widehat\theta,\theta)
  \le
  \frac{2\overline e(p,t)}r
  +
  \frac{\ECE_{\mathcal D_\theta}(p)}\eta,
  \label{eq:det-decoding-decomposition}
\end{equation}
where $d_H$ denotes Hamming distance.
\end{lemma}

\begin{proof}
Write $p_i=p(i)$ and call $i$ good if
$|p_i-t_i|<r$.  Markov's inequality shows that at most
$R\overline e(p,t)/r$ indices are bad.  No two good indices have the same
prediction value: if $i\ne j$, then
\[
  |p_i-p_j|
  \ge |t_i-t_j|-|p_i-t_i|-|p_j-t_j|
  >\Delta_t-2r
  >0.
\]
Every good index that shares its prediction value with another index must
therefore share it with a bad index.  Different good indices have different
prediction values, so they must share with different bad indices.  Thus
there are at most $R\overline e(p,t)/r$ such good indices.  All but at most
$2R\overline e(p,t)/r$ indices are therefore good and have a prediction
value shared by no other index.

If a good index $i$ whose prediction value is unique is decoded incorrectly, then
$|\mu_i^\theta-p_i|\ge\eta$: for example, when $\theta_i=1$, an incorrect
sign means $p_i\le t_i$ while $\mu_i^\theta=t_i+\eta$, and the other case is
symmetric.  Because no other point receives prediction $p_i$, this point contributes
$|\mu_i^\theta-p_i|/R$ to ECE.  Hence the number of incorrectly decoded
singleton good indices is at most
\[
  \frac{R\ECE_{\mathcal D_\theta}(p)}{\eta}.
\]
Adding this count to the bound on the remaining indices proves
\eqref{eq:det-decoding-decomposition}.
\end{proof}

Lemma~\ref{lem:sign-recovery-information} now converts this decoding
guarantee into a sample lower bound of order $R/\eta^2$.

It remains to choose a prescribed number of sign tests that force the
predictor close to the baseline.
The next lemma uses a parameter $K_{\rm test}$ to construct about
$\exp(\Theta(K_{\rm test}))$ sign tests such that
every vector $d\in\R^N$ has correlation at least a constant times
$\sqrt{K_{\rm test}/N}\|d\|_1$ with one of them.  Therefore, if the residual vector
$d_i=\mu_i^\theta-p_i$ had large average absolute value, some test in the
family would detect it.  Multiaccuracy therefore bounds the distance from
the conditional means, which are themselves close to the public baselines.

\begin{lemma}[A small family of sign tests detects $\ell_1$ error]
\label{lem:random-sign-l1}
There are universal constants $a_*,A_*,c_*,C_*>0$
such that the following holds.  Let $N\ge2$ and let $K_{\rm test}$ satisfy
\[
  A_*\log N\le K_{\rm test}\le a_*N .
\]
Then there is a family $\calG\subseteq\{\pm1\}^{[N]}$ with
\[
  \exp(c_*K_{\rm test})\le |\calG|\le \exp(C_*K_{\rm test})
\]
such that, for every $d\in\R^N$,
\begin{equation}
  \max_{g\in\calG}\left|\sum_{i=1}^N g(i)d_i\right|
  \ge
  c_*\sqrt{\frac{K_{\rm test}}N}\sum_{i=1}^N |d_i|.
  \label{eq:random-sign-l1}
\end{equation}
\end{lemma}

The construction splits the coordinates into about $K_{\rm test}$ blocks.
A distribution over a small set of sign patterns gives, in each block, an expected
absolute inner product at least a constant times the Euclidean norm of
that block's residual vector.  We include every choice of sign flips on
the blocks.  For each pattern, suitable flips make all block contributions
nonnegative, so averaging over patterns shows that some test attains the
sum of these lower bounds.  Each block has $O(N/K_{\rm test})$ coordinates, so its Euclidean norm
is at least a constant times $\sqrt{K_{\rm test}/N}$ times its $\ell_1$ norm.
The proof is deferred to
Appendix~\ref{app:det-lower-proofs}.

\begin{proof}[Proof of Theorem~\ref{thm:det-budget-lower}]
Let $a_*,A_*,c_*,C_*$ be the constants from
Lemma~\ref{lem:random-sign-l1}, increasing $C_*$ if necessary so that
$C_*\ge1$.  Set $K_{\rm test}=A/C_*$.

\paragraph{Choosing the domain size.}
With $R$ domain points, the sign-test lemma and multiaccuracy will bound
the average distance from the conditional means by
$O(\eps\sqrt{R/K_{\rm test}})$.  The means themselves are within $\eps$
of the public baselines.  The decoding lemma requires average distance
from the baselines to be a small fraction of their spacing, $1/R$.
Balancing the first error with this spacing gives
\[
  \eps\sqrt{\frac{R}{K_{\rm test}}}\asymp\frac1R,
  \qquad\text{so we set}\qquad
  R=\left\lfloor K_{\rm test}^{1/3}\eps^{-2/3}\right\rfloor.
\]
The second error, $\eps$, is small relative to $1/R$ because
$R\eps\le(K_{\rm test}\eps)^{1/3}\le(a_0/C_*)^{1/3}$.

For sufficiently small $\eps$, $R\ge2$ and the floor loses at most a
factor two.  Also $\log R=O(\log(1/\eps))$ and
\[
  \frac{K_{\rm test}}R\le2(K_{\rm test}\eps)^{2/3}
  \le2(a_0/C_*)^{2/3}.
\]
Thus taking $C_0$ sufficiently large and $a_0$ sufficiently small
ensures $A_*\log R\le K_{\rm test}\le a_*R$, as required by the
sign-test lemma.

\paragraph{Constructing the hard family.}
Apply Lemma~\ref{lem:random-sign-l1} with $N=R$ to obtain
$\calG\subseteq\{\pm1\}^{[R]}$ satisfying \eqref{eq:random-sign-l1} and
\[
  \frac{c_*}{C_*}A\le\log|\calG|\le A.
\]
Set $\calX_{\eps,A}=[R]$ and $\calG_{\eps,A}=\calG$.  Let $X$ be uniform
on $[R]$, and define
\[
  t_i=\frac13+\frac{i}{10R},
  \qquad
  \mu_i^\theta=t_i+\eps\theta_i,
  \qquad i=1,\ldots,R.
\]
For each $\theta\in\{\pm1\}^R$, let $\mathcal D_\theta$ have conditional
labels $Y\mid X=i\sim\Ber(\mu_i^\theta)$, and put
$\mathfrak D_{\eps,A}=\{\mathcal D_\theta:\theta\in\{\pm1\}^R\}$.
All means lie in $[1/4,3/4]$ for sufficiently small $\eps$.

\paragraph{Multiaccuracy forces closeness to the baseline.}
Fix $\theta$ and a deterministic predictor $p$ with
$\MA_{\mathcal D_\theta}(p;\calG)\le c\eps$, and write $p_i=p(i)$.
Apply the sign-test lemma directly to the residual vector
$(\mu_i^\theta-p_i)_{i=1}^R$:
\[
  c_*\sqrt{\frac{K_{\rm test}}R}\sum_{i=1}^R|\mu_i^\theta-p_i|
  \le
  \max_{g\in\calG}\left|\sum_{i=1}^R g(i)(\mu_i^\theta-p_i)\right|
  \le Rc\eps.
\]
Since $|\mu_i^\theta-t_i|=\eps$, the triangle inequality gives
\[
  \overline e(p,t)
  \le\frac{c}{c_*}\eps\sqrt{\frac{R}{K_{\rm test}}}+\eps.
\]

\paragraph{Calibration reveals the signs.}
On the event $\CMA_{\mathcal D_\theta}(p;\calG)\le c\eps$, we also have
$\ECE_{\mathcal D_\theta}(p)\le c\eps$.  Apply
Lemma~\ref{lem:deterministic-decoding} with $r=1/(40R)$ and $\eta=\eps$.
The baselines have spacing $1/(10R)>2r$, and
\begin{align*}
  \frac1R d_H(\widehat\theta,\theta)
  &\le80R\,\overline e(p,t)+c\\
  &\le\frac{80c}{c_*}
       \frac{\eps R^{3/2}}{\sqrt{K_{\rm test}}}
       +80R\eps+c\\
  &\le\left(\frac{80}{c_*}+1\right)c+80(a_0/C_*)^{1/3}.
\end{align*}
Choose $a_0$ sufficiently small that the second term is at most $1/10$
and $c$ sufficiently small that the first is at most $1/10$.
The decoder therefore recovers at least $0.8R$ hidden signs on the
success event.  It uses only $p$ and the public baselines, not the
conditional means used in the preceding analysis.

The learning guarantee makes this event occur with probability at least
$2/3$ for every $\theta$.  Lemma~\ref{lem:sign-recovery-information},
with $\eta=\eps$, gives
\[
  n\ge c_{\rm info}\frac R{\eps^2}
  \ge c'A^{1/3}\eps^{-8/3},
\]
using $K_{\rm test}=A/C_*$ and the floor estimate above.
For constant confidence, the upper bound constructs
$\Theta(A^{1/3}\eps^{-2/3})$ categories, essentially the same number
$R$ of singleton prediction values forced by this lower bound.
Taking $c_{\rm size}\le c_*/C_*$ and $c_{\rm samp}\le c'$ proves the theorem.
\end{proof}

\section{Discussion}

We have resolved the sample complexity of calibrated multiaccuracy for both 
randomized predictors and deterministic predictors and have shown that requiring the output to be deterministic
comes with a polynomial statistical cost.

Two questions remain.  First, the learning algorithm that we give achieving the tight deterministic
upper bound runs in exponential time; finding a polynomial-time algorithm with
the same rate is open.  Second, our  lower bound for deterministic learners  requires
$\log|\calG|=\Omega(\log(1/\eps))$.  Thus the deterministic rate remains open
when $\log|\calG|=o(\log(1/\eps))$, including in particular for constant-size
families of real-valued bounded tests.  For these constant-size families, the known bounds are
$\Omega(\eps^{-5/2})$ and
$\widetilde O(\eps^{-8/3})$. Note that easy improvements are possible in the special case of a constant number of  binary valued indicator tests (i.e. ``groups''): if we have $k$ groups for some constant $k$, partitioning contexts by
their $2^k$ group-membership patterns yields only a constant number of cells despite the exponential blow-up.
Predicting the empirical label mean on each cell gives a deterministic
predictor that satisfies calibrated multi-accuracy with sample complexity $O(\eps^{-2})$. This construction fails for real-valued test functions.

\section*{AI methodology}

We used GPT models ranging from GPT-5.5 to GPT-6 in the development of this
paper to explore proof strategies, simplify arguments, and draft and revise
mathematical proofs. In particular, after we had the randomized upper bound and both lower bounds, the construction of the matching deterministic
upper bound from Section \ref{sec:det-upper} was found by GPT-5.6 Sol during an approximately 24-hour
goal-directed run in the Codex environment using Ultra mode. The authors
have verified all proofs and are responsible for the contents of the paper.

\section*{Acknowledgments}

This work was partially supported by the Simons Collaboration on Algorithmic
Fairness, NSF grants RI-2551608 and CCF-2217062, and a gift from Apple.

\bibliographystyle{plainnat}
\bibliography{references,refs}

\appendix

\section{Deferred proofs for the deterministic upper bound}
\label{app:det-upper}

This appendix supplies the preprocessing and concentration arguments used in
Section~\ref{sec:det-upper}.  Each stage succeeds with high probability,
so one run suffices for the theorem's guarantee.

Throughout this appendix, assume $A_\delta\le c/\eps$ and use the parameters
from the proof of Theorem~\ref{thm:det-upper}.  Thus
\[
  K=\left\lceil c_K(A_\delta/\eps^2)^{1/3}\right\rceil,
  \qquad
  \alpha=c_\alpha\eps,
  \qquad
  L=A_\delta+K,
\]
with $A_\delta\le K$ and
$\sqrt{A_\delta/K^3}=O(\eps)$.  Let
\begin{equation}
  \ell=C_{\ell}\log\frac{C_{\ell}L}{\alpha\delta},
  \qquad
  N=\left\lceil C_N\frac{L\ell}{\alpha^2}\right\rceil.
  \label{eq:det-app-parameters}
\end{equation}
Here the universal constants are chosen so that the inequalities below hold.
Each sample split has size $N$.

\subsection{Confidence intervals and the context partition}

Let $O_0$ be the contexts observed in the preprocessing sample, and let $N_x$
be the number of appearances of $x$.  For $x\in O_0$, let
$\widehat\mu_x$ be the average label and set
\[
  \rho_x=\min\{1,\sqrt{\ell/N_x}\},
  \qquad
  I_x=[\widehat\mu_x-\rho_x,\widehat\mu_x+\rho_x]\cap[0,1].
\]
For $x\notin O_0$, set $I_x=[0,1]$.  With
$\gamma=\alpha/64$ and $N_{\rm grid}=\lceil1/\gamma\rceil$, define
\[
  \Lambda=\{i/N_{\rm grid}:0\le i\le N_{\rm grid}\},
  \qquad
  \Lambda_x=\{v\in\Lambda:\dist(v,I_x)\le\gamma\}.
\]

Fix a constant $c_{\rm fix}>0$ and define
\begin{equation}
  \widehat B=\{x:N_x\ge c_{\rm fix}N/K\}.
  \label{eq:det-app-lists}
\end{equation}
Each point in this fixed-score set keeps its score and forms a private
singleton category.

The preprocessing sample also induces the context partition.  Sort its
distinct observed contexts in lexicographic order as
$z_1<\cdots<z_q$.  Partition $\calX\setminus O_0$ into the gap cells
\[
  \{x:x<z_1\},\qquad
  \{x:z_i<x<z_{i+1}\}\ (1\le i<q),\qquad
  \{x:x>z_q\},
\]
where every set is understood to lie in $\calX\setminus O_0$, and add the
singleton $\{x\}$ for every $x\in O_0$.  Empty gap cells may be
retained by the implementation and are ignored when calling the resulting
family a partition.  Denote this finite measurable partition by
$\mathcal U$.  A singleton $\{x\}$ with $x\in\widehat B$ is excluded from the
random assignment; all other cells are assignment cells.  If no assignment
cells remain, their maximum mass is taken to be zero and all coarsening
errors vanish.

\begin{proof}[Proof of Lemma~\ref{lem:det-preprocessing}]
The counts sum to $N$, so $|\widehat B|\le K/c_{\rm fix}$.
At the parameters above, $N/K\ge c\ell/\eps^2$, so the inclusion threshold
exceeds two.  For every atom of mass $p>C/K$, its expected count is a
sufficiently large constant factor above this threshold.  A lower-tail
Chernoff bound gives miss probability at most $\exp(-cN/K)$.
There are at most $K/C$ such atoms, so a union bound detects all of them
with failure probability at most $\delta/100$.
For coverage, condition on the full feature sample $(X_1,\ldots,X_N)$.
Almost surely in this sample, the labels are conditionally independent and
the labels at an observed context $x$ have mean $\mu(x)$.  Whenever
$\rho_x<1$, Hoeffding's inequality gives
\[
  \Pp\left(
    |\widehat\mu_x-\mu(x)|>\sqrt{\ell/N_x}
    \,\middle|\,X_1,\ldots,X_N
  \right)\le2e^{-2\ell}.
\]
There are at most $N$ observed contexts.  A union bound, followed by averaging
over the feature sample, proves simultaneous coverage of them; contexts with
$\rho_x=1$ and unobserved contexts receive $[0,1]$.  Thus, on the coverage
event, $\mu(x)\in I_x$ for every $x\in\calX$.  Taking $C_\ell$ sufficiently
large makes the failure probability $2Ne^{-2\ell}$ at most $\delta/100$.
For any $u\in I_x$, the $\gamma$-net property of $\Lambda$ gives a grid value
$v$ with $|u-v|\le\gamma$.  Then $\dist(v,I_x)\le\gamma$, so
$v\in\Lambda_x$.  This also proves that every $\Lambda_x$ is nonempty.  By
definition, every $v\in\Lambda_x$ satisfies $\dist(v,I_x)\le\gamma$, and all
allowed sets are subsets of the common finite grid $\Lambda$.

For the radius term, consider atoms with $p_x\ge8\ell/N$.
There are at most $N$ of them.  Chernoff's inequality and a union bound give
$N_x\ge Np_x/2$ simultaneously for these atoms, except with probability
at most $Ne^{-\ell}\le\delta/100$.  On this event,
\begin{align*}
  \sum_{\substack{x\text{ observed}\\p_x>0}}p_x^2\rho_x^2
  &\le
  \sum_{p_x\ge8\ell/N}p_x^2\frac{2\ell}{Np_x}
  +\sum_{0<p_x<8\ell/N}p_x^2\\
  &\le\frac{2\ell}{N}\sum_xp_x+\frac{8\ell}{N}\sum_xp_x
  \le\frac{10\ell}{N}.
\end{align*}

For the gaps, write $m_U=P_X(U)$ and introduce $r$ fresh independent
contexts only for the analysis.  Conditional on the preprocessing sample,
$\sum_Um_U^r$ is the probability that all $r$ lie in the same gap.
Attach independent uniform $[0,1]$ tie-breakers to all $N+r$ contexts,
sorting first by context and using these numbers only between equal
contexts.  If the fresh contexts lie in one unobserved gap, they occupy
$r$ consecutive positions.  This also covers equal fresh contexts,
since none of the preprocessing contexts can lie in that gap.
Exchangeability makes their positions a uniform $r$-subset, and there are
$N+1$ possible consecutive blocks.  Thus
\[
  \E\sum_{U\text{ unobserved}}m_U^r
  \le\frac{N+1}{\binom{N+r}{r}}
  \le N\left(\frac rN\right)^r.
\]
Consequently, Markov's inequality applied to $\sum_Um_U^r$ gives
\[
  \Pp\left(\max_{U\text{ unobserved}}m_U>\frac{er}{N}\right)
  \le Ne^{-r}.
\]
Take $r=\lceil\log(100N/\delta)\rceil$.  The parameters in
\eqref{eq:det-app-parameters} ensure $r=O(\ell)$, so, with failure
probability at most $\delta/100$,
\[
  \sum_{U\text{ unobserved}}m_U^2
  \le\max_Um_U\sum_Um_U
  \le\frac{C\ell}{N}.
\]
The tie-breakers are not part of the algorithm.

Combining these events proves
$V_{\mathcal U}\le C\ell/N\le C\alpha^2/L$
with failure probability at most $\delta/10$.
The gap bound also gives
\[
  \max_{U\text{ unobserved}}P_X(U)
  \le C\ell/N=O(\alpha^2/L)=O(1/K).
\]
An observed atom outside the fixed-score set has mass $O(1/K)$ by the
detection guarantee, and an observed non-atom has mass zero.  This proves
the assignment-cell mass bound.
\end{proof}

The coverage and squared-mass arguments do not use the fixed-score threshold.
For any $N$ and $\ell\ge C\log(N/\delta)$, the same intervals, allowed grid
values, and singleton-and-gap partition therefore give
pointwise coverage and $V_{\mathcal U}\le C\ell/N$ with probability at least $1-\delta/10$,
without forming $\widehat B$.  We use this observation in
the large-budget case.

Conditional on the event in Lemma~\ref{lem:det-preprocessing}, the auxiliary
class in \eqref{eq:det-category-test} is fixed before the labeled score sample
is drawn.  Its log-cardinality is $O(K)$.  Applying
Lemma~\ref{lem:det-finite-test-score}, with failure probability $\delta/10$, gives
a deterministic $S$ supported on
$\Lambda_x$ for which
\eqref{eq:det-score-ma}--\eqref{eq:det-score-signs} hold.  Indeed, their
conditional bound is
\[
  C\gamma
  +C\sqrt{\frac{\log(20|\Phi_K|/\delta)}{N}}
  +C\sqrt{
    \frac{\alpha^2}{L}\log(20|\Phi_K|/\delta)}
  =O(\alpha).
\]

\subsection{Transferring empirical residuals to the true labels}
\label{app:det-upper-transfer}

\begin{proof}[Proof of Lemma~\ref{lem:det-coarsening-transfer}]
We also establish the empirical mass bound
\eqref{eq:det-empirical-cell-cap}.  Condition throughout on successful
preprocessing and score learning.  The partition, fixed-score set, and score
are therefore fixed before the fresh context sample is drawn.  Write
$\Delta=P_X-\widehat P_X$ and $\mu(x)=\E[Y\mid X=x]$.

The proof first bounds, for each cell, the difference between population
and sample averages of bounded functions supported on that cell.  We then
combine these bounds with the score residuals $\mu-S$ to control the
effect of fixing the independent assignment choices.  The expected
residual difference is an average over rules that use one tuple everywhere;
score learning and a finite-class sample bound control each of those rules.

\smallskip\noindent\emph{Squared cell discrepancies.}
Let $m_U=P_X(U)$, $\widehat m_U=\widehat P_X(U)$, and define
\[
  a_U=
  \begin{cases}
    |m_U-\widehat m_U|,&U\text{ is an observed singleton},\\
    m_U+\widehat m_U,&U\text{ is an unobserved gap}.
  \end{cases}
\]
For every function $f$ with $|f|\le1$ on $U$,
$|\Delta[\one_U f]|\le a_U$.  For a singleton $U=\{x\}$, the difference
is $(m_U-\widehat m_U)f(x)$; for a gap, the population and empirical
integrals have absolute values at most $m_U$ and $\widehat m_U$.
Since $(m_U+\widehat m_U)^2\le
8m_U^2+2(\widehat m_U-m_U)^2$, we have
\[
  T_\Delta:=\sum_Ua_U^2
  \le
  2\|\widehat{\mathbf m}-\mathbf m\|_2^2
  +8\sum_{U\text{ unobserved}}m_U^2,
\]
where $\mathbf m,\widehat{\mathbf m}$ are the vectors of cell masses.
Conditional on the partition, and taking expectation over the fresh
context sample, the empirical mass vector satisfies
\[
  \E\|\widehat{\mathbf m}-\mathbf m\|_2^2
  =\frac{1-\sum_Um_U^2}{N}\le\frac1N.
\]
Changing one observation moves the empirical mass vector by at most
$\sqrt2/N$ in Euclidean norm, and hence changes
$\|\widehat{\mathbf m}-\mathbf m\|_2$ by at most the same amount.
Jensen's inequality and bounded differences therefore give,
with failure probability at most $\delta/20$,
\[
  \|\widehat{\mathbf m}-\mathbf m\|_2
  \le \frac1{\sqrt N}+\sqrt{\frac{\log(20/\delta)}N}.
\]
The preprocessing bound on squared gap masses now implies
\begin{equation}
  T_\Delta\le C\ell/N.
  \label{eq:det-cell-discrepancies}
\end{equation}
On the same event,
\[
  \widehat b
  \le \max_{U\text{ assignment cell}}m_U
       +\|\widehat{\mathbf m}-\mathbf m\|_2
  \le C/K+C\sqrt{\ell/N}=O(1/K).
\]
The last step uses $N\ge C K^2\ell$, which follows from the parameter choices
after taking $C_N$ sufficiently large: $N=\Theta(K^4\ell/A_\delta)$ and
$A_\delta\le K$.  This proves \eqref{eq:det-empirical-cell-cap}.

\smallskip\noindent\emph{Residual differences for global tuples.}
For any nominal prediction vector $r\in[0,1]^{K+1}$, let $p_{\omega,r}$
predict $r_j$ on nominal category $j$ and retain $S$ on private categories.
Write $p_{\tau,r}$ when the same tuple $\tau$ is used in every assignment
cell.  Initially restrict to $r\in\{0,1\}^{K+1}$.
For each global tuple, consider either a group test or a full category-sign
test,
\[
  \chi_{\tau,v}(x)=g(x)
  \quad\text{or}\quad
  \chi_{\tau,v}(x)=\sigma_{J_\tau(x,S(x))},
  \qquad \sigma\in\{\pm1\}^{K+1+K_{\rm priv}},
\]
where $v$ records the chosen test and prediction vector $r$.  Put
$f_{\tau,v}=\chi_{\tau,v}(S-p_{\tau,r})$.  The residual difference we must
control is, by Lemma~\ref{lem:postprocessing-decomposition},
\begin{align*}
  Z_{\tau,v}
  &=\E[\chi_{\tau,v}(X)(Y-p_{\tau,r}(X))]
    -\widehat P_X[f_{\tau,v}]\\
  &=P_X[\chi_{\tau,v}(\mu-S)]+\Delta[f_{\tau,v}].
\end{align*}
The score residual $P_X[\chi_{\tau,v}(\mu-S)]$ is at most $C\alpha$ in
absolute value by \eqref{eq:det-score-ma}--\eqref{eq:det-score-signs}.  The functions
$f_{\tau,v}$ vanish on private categories and are bounded by one.  Their
number is at most
\[
  |\calH|^{m_{\rm sp}}
  \bigl(|\calG|\,2^{K+1}+2^{2(K+1)+K_{\rm priv}}\bigr),
\]
whose logarithm is $O(A_\delta+K)$.  This class is fixed before the context
sample.  Hoeffding's inequality and a union bound give, with failure
probability at most $\delta/40$,
\[
  \max_{\tau,v}|\Delta[f_{\tau,v}]|
  \le C\sqrt{\frac{A_\delta+K}{N}}\le C\alpha.
\]
Consequently, on this event,
\begin{equation}
  \max_{\tau,v}|Z_{\tau,v}|\le C\alpha.
  \label{eq:det-global-transfer}
\end{equation}

\smallskip\noindent\emph{Independent cell assignments.}
Now fix a context sample satisfying this event and
\eqref{eq:det-cell-discrepancies}, and let $\widehat\pi$ be its selected
distribution.  Define $\chi_{\omega,v}$ and $Z_{\omega,v}$ by replacing the
global tuple with the realized cell tuples:
\[
  Z_{\omega,v}
  =P_X[\chi_{\omega,v}(\mu-S)]
     +\Delta[\chi_{\omega,v}(S-p_{\omega,r})].
\]
For an assignment cell $U$, put
$c_U=P_X[\one_U|\mu-S|]$.  Coverage and $S(x)\in\Lambda_x$ imply
$c_{\{x\}}\le p_x(2\rho_x+\gamma)$ at observed singletons, while
$c_U\le P_X(U)$ on unobserved gaps.  Let $b=O(1/K)$ be the largest
population assignment-cell mass.  Then
\[
  \sum_{U\text{ assignment cell}}c_U^2
  \le8V_{\mathcal U}
     +2\gamma^2\sum_{x\text{ observed, nonprivate}}p_x^2
  \le8V_{\mathcal U}+2\gamma^2b.
\]
The contribution of $U$ to $Z_{\omega,v}$ has absolute value at most
$c_U+a_U$.  Therefore
\[
  \sum_{U\text{ assignment cell}}(c_U+a_U)^2
  \le C\bigl(V_{\mathcal U}+\gamma^2b+T_\Delta\bigr)
  \le C\frac{\alpha^2}{A_\delta+K}.
\]
Here $V_{\mathcal U},T_\Delta=O(\ell/N)$, $\gamma=\alpha/64$, and
$A_\delta\le K$.  Conditional on the sample, the assignment-cell
contributions are independent.  The private contribution is deterministic
and disappears after centering.  Hoeffding's inequality and a union bound
over $v$ thus give, with conditional failure probability at most $\delta/40$,
\[
  \max_v|Z_{\omega,v}-\E_\omega Z_{\omega,v}|\le C\alpha.
\]
Every assignment cell has the same marginal tuple law; private categories
are unchanged under every tuple.  Hence
\[
  \E_\omega Z_{\omega,v}
  =\E_{\tau\sim\widehat\pi^{m_{\rm sp}}}Z_{\tau,v}.
\]
Equation~\eqref{eq:det-global-transfer} bounds this expectation even when
$\widehat\pi$ depends on the sample.  Combining the bounds gives
$\max_v|Z_{\omega,v}|\le C\alpha$.

For a fixed assignment and test, $Z_{\omega,v}$ is affine in $r$.
Its absolute value on $[0,1]^{K+1}$ is maximized at a vertex, so the same
event covers every nominal prediction vector, including the empirical means
computed from this sample and assignment, with zero for categories
containing no sample points.
For these means, $Z_{\omega,v}$ is exactly the difference of population and
empirical residuals in
\eqref{eq:det-coarsening-transfer}.  The three failure probabilities sum to
$\delta/10$, completing the proof of
Lemma~\ref{lem:det-coarsening-transfer}.
\end{proof}

\subsection{Fixing the category assignments}
\label{app:det-upper-seeds}

We first isolate the concentration estimate used for the multiaccuracy
residuals.

\begin{lemma}[Concentration after category averaging]
\label{lem:det-category-average-concentration}
Suppose every assignment cell has mass at most $b$, and every category has
score diameter at most $D_{\rm cat}$.  Then, for every
$h:\calX\to[-1,1]$ and $t\ge0$,
\[
  \Pp_\omega\left(
    |F_h(\omega)-\E_\omega F_h(\omega)|
    >C\left(
      \sqrt{b\,\E_\omega V(\omega)t}+bD_{\rm cat}t
    \right)
  \right)
  \le2e^{-t}.
\]
\end{lemma}

The change caused by deleting an assignment cell is controlled by the
integral of $(S-p_\omega)^2$ over that cell.  Summing over cells gives the two
bounds needed for a moment form of the Efron--Stein inequality.

We use the following leave-one-coordinate-out form of the generalized
Efron--Stein inequality
\citep[Corollary~D.1, equation~(D.4)]{CelisseMaryHuard2018CrossValidation};
see also \citet[Theorem~15.5]{BoucheronLugosiMassart2013} for the
coordinate-resampling form.  If $Z$ is a measurable
function of independent coordinates and $Z_{-i}$ is any measurable function
of all coordinates except the $i$th, then, for every real $q\ge2$,
writing $\|Z\|_q=(\E|Z|^q)^{1/q}$,
\begin{equation}
  \|Z-\E Z\|_q
  \le
  C\sqrt q\left\|
    \left(\sum_i(Z-Z_{-i})^2\right)^{1/2}
  \right\|_q.
  \label{eq:det-moment-efron-stein}
\end{equation}
For a measurable set $A$ of positive mass, write
\[
  \overline S_A=\frac1{P_X(A)}\int_A S\,dP_X,
  \qquad
  \overline h_A=\frac1{P_X(A)}\int_A h\,dP_X,
\]
and define the mass-weighted variance and covariance
\[
  V(A)=\int_A(S-\overline S_A)^2\,dP_X,
  \qquad
  F_h(A)=\int_Ah(S-\overline S_A)\,dP_X.
\]
These are $P_X(A)$ times the conditional variance of $S$ and the conditional
covariance of $h$ and $S$ on $A$, respectively.  Both are defined to be
zero when $P_X(A)=0$.  The following identity
controls the effect of deleting part of a positive-mass category
$C=A\sqcup B$:
\begin{equation}
  F_h(C)-F_h(B)
  =\int_A(h-\overline h_B)(S-\overline S_C)\,dP_X.
  \label{eq:det-covariance-deletion}
\end{equation}
Indeed, integrating the product on the right over $C$ gives $F_h(C)$,
because $S-\overline S_C$ has mean zero there.  Integrating it over $B$
gives $F_h(B)$, because $h-\overline h_B$ has mean zero there.  Subtracting
proves the identity.  If $P_X(B)=0$, choose any
$\overline h_B\in[-1,1]$.

\begin{lemma}[Squared changes from deleting assignment cells]
\label{lem:det-deletion-calculus}
Fix a realized assignment and a test $h:\calX\to[-1,1]$.  Let
$V_{\rm tot}$ and $F_{\rm tot}$ be the
sums of $V(C_j)$ and $F_h(C_j)$ over its nominal categories.
Delete assignment cell $U_i$ from every category, leaving the other
assignments fixed and recomputing each category average on the remaining
contexts.  Denote the resulting totals by $V_{-i}$ and
$F_{h,-i}$.  If every assignment cell has mass at most $b$ and every category
has score diameter at most $D_{\rm cat}$, then
\begin{equation}
  \sum_i(F_{\rm tot}-F_{h,-i})^2\le4bV_{\rm tot},
  \qquad
  \sum_i(V_{\rm tot}-V_{-i})^2\le bD_{\rm cat}^2V_{\rm tot}.
  \label{eq:det-deletion-bounds}
\end{equation}
\end{lemma}

\begin{proof}
Write $p_\omega$ for the category-averaged score and set
\[
  c_i=P_X(U_i),\qquad
  E_i=\int_{U_i}(S-p_\omega)^2\,dP_X.
\]
The assignment cells partition the nonprivate region, so
$\sum_iE_i=V_{\rm tot}$.  Apply \eqref{eq:det-covariance-deletion} to
$A=C_j\cap U_i$, $B=C_j\setminus U_i$, and sum over positive-mass
categories; zero-mass categories contribute zero.
Since $|h-\overline h_B|\le2$, Cauchy--Schwarz gives
\[
  |F_{\rm tot}-F_{h,-i}|^2\le4c_iE_i.
\]
The same identity with $h=S$ controls variance, because $F_S(C)=V(C)$.
Both $S$ and $\overline S_B$ lie in the interval spanned by the scores in
$C_j$; when $P_X(B)=0$, choose $\overline S_B$ anywhere in that interval.
This auxiliary choice does not affect the zero-mass covariance.  Thus
\[
  |V_{\rm tot}-V_{-i}|^2\le D_{\rm cat}^2c_iE_i.
\]
Summing these inequalities and using $c_i\le b$ proves the claim.
\end{proof}

\begin{proof}[Proof of Lemma~\ref{lem:det-category-average-concentration}]
All objects in the lemma other than the assignment seeds are fixed.  In the
application, this means conditioning on the preprocessing transcript, the
learned score $S$, the selection sample, and the selected distribution
$\widehat\pi$.  Under this conditioning, the cell seeds are independent.
All expectations and moment norms below are over these seeds.  Write $V_{\rm tot}=V(\omega)$,
$F_{\rm tot}=F_h(\omega)$, and $\overline V=\E_\omega V_{\rm tot}$.

First control $W=\sqrt{V_{\rm tot}}$, the root-mean-square change in the
score.  Taking the square root gives a deletion bound that no longer
depends on the realized $V_{\rm tot}$.  Write
$W_{-i}=\sqrt{V_{-i}}$.  When $V_{\rm tot}>0$,
\eqref{eq:det-deletion-bounds} gives
\[
  \sum_i(W-W_{-i})^2
  =\sum_i\frac{(V_{\rm tot}-V_{-i})^2}
                  {(\sqrt{V_{\rm tot}}+\sqrt{V_{-i}})^2}
  \le bD_{\rm cat}^2.
\]
When $V_{\rm tot}=0$, the same deletion bound forces every $V_{-i}=0$.
Applying \eqref{eq:det-moment-efron-stein} to $W$, then using
$\E W\le\sqrt{\overline V}$, shows that for $q\ge2$,
\[
  \|W\|_q\le\sqrt{\overline V}+CD_{\rm cat}\sqrt{bq}.
\]
Now apply \eqref{eq:det-moment-efron-stein} to $F_{\rm tot}$.
The first bound in \eqref{eq:det-deletion-bounds} gives
\[
  \|F_{\rm tot}-\E F_{\rm tot}\|_q
  \le C\sqrt{bq}\,\|W\|_q
  \le C\left(\sqrt{b\overline Vq}+bD_{\rm cat}q\right).
\]
For $t\ge2$, take $q=t$ and apply Markov's inequality at a sufficiently
large constant multiple of the right side to obtain failure probability at
most $2e^{-t}$.  When $t\le\log2$, the asserted probability bound is trivial;
for $\log2<t<2$, the $q=2$ bound implies it after enlarging the same universal
constant.  This proves
Lemma~\ref{lem:det-category-average-concentration}.
\end{proof}

\begin{proof}[Proof of Lemma~\ref{lem:det-fix-assignments}]
Condition on the preprocessing, score, and context sample satisfying
\eqref{eq:det-empirical-cell-cap}.  Apply
Lemma~\ref{lem:det-category-average-concentration} to the empirical
distribution $\widehat P_X$, with its category averages $\widehat p_\omega$.
Its assignment-cell mass bound is $\widehat b=O(1/K)$, and
Lemma~\ref{lem:det-law-selection} gives
$\E_\omega\widehat V(\omega)=O(K^{-2})$ and
$|\E_\omega\widehat F_g(\omega)|=O(\xi)$.
Taking $t=CA_\delta$ and union bounding over the signed class gives
\eqref{eq:det-fixed-multiaccuracy-residuals} with failure probability at most
$\delta/10$, since
\[
  \sqrt{\widehat b\,\E_\omega\widehat V\,t}
  =O\!\left(\sqrt{\frac{A_\delta}{K^3}}\right),
  \qquad
  \widehat bD_{\rm cat}t
  =O\!\left(\sqrt{\frac{A_\delta}{K^3}}\right).
\]

\end{proof}

\section{Deferred proofs from the randomized lower bound}
\label{app:lower-deferred-proofs}

This appendix proves the shared recovery bound, followed by the two
applications from Section~\ref{sec:lower}.

\begin{proof}[Proof of Lemma~\ref{lem:sign-recovery-information}]
Pair a hidden vector $\theta$ with $\theta^{(i)}$, obtained by flipping only
coordinate $i$.  Write $a=t_i+\eta\theta_i$ and
$b=t_i+\eta\theta_i^{(i)}$ for the two label means at that coordinate.
They lie in $[1/8,7/8]$ and satisfy $|a-b|\le2\eta$.
For one sample, the likelihood ratio
$L_1=d\mathcal D_\theta/d\mathcal D_{\theta^{(i)}}$ is $1$ unless the
sampled context is $i$.  Here $\E_{\theta^{(i)}}$ denotes expectation over
samples from $\mathcal D_{\theta^{(i)}}$, with the hidden vector fixed.
Consequently,
\[
  \E_{\theta^{(i)}}L_1^2
  =1-\frac1R+\frac1R\left(\frac{a^2}{b}
                  +\frac{(1-a)^2}{1-b}\right)
  =1+\frac{(a-b)^2}{Rb(1-b)}
  \le1+\frac{C\eta^2}{R}.
\]
For $n$ independent samples the likelihood ratio $L_n$ is a product, so
$\E_{\theta^{(i)}}L_n^2\le\exp(Cn\eta^2/R)$.  Including the estimator's
independent seed does not change this ratio.  If $\mathbb P_\theta$
denotes the law of this full transcript, Cauchy--Schwarz gives
\[
  \operatorname{TV}(\mathbb P_\theta,\mathbb P_{\theta^{(i)}})
  =\frac12\E_{\theta^{(i)}}|L_n-1|
  \le\frac12\sqrt{\exp(Cn\eta^2/R)-1}.
\]
Here total variation is the supremum of the absolute difference in
probabilities assigned to the same event.  Thus, when
$n\le c_{\rm info}R/\eta^2$ for a sufficiently small universal constant
$c_{\rm info}$, it is at most $1/30$.

The two values of coordinate $i$ are different.  Applying the total-variation
bound to the event $\{\widehat\theta_i=\theta_i\}$ therefore gives
\[
  \mathbb P_\theta(\widehat\theta_i\ne\theta_i)
  +\mathbb P_{\theta^{(i)}}(\widehat\theta_i\ne\theta_i^{(i)})
  \ge1-\operatorname{TV}(\mathbb P_\theta,\mathbb P_{\theta^{(i)}})
  \ge\frac{29}{30}.
\]
Average over all pairs in $\mathcal B^R$ and sum over $i$.  For uniform
$\Theta\in\mathcal B^R$, this implies
$\E d_H(\widehat\Theta,\Theta)\ge29R/60$.
On the other hand, the Hamming distance is at most $R/5$ on success and
at most $R$ on failure, so the assumed recovery guarantee gives
\[
  \E d_H(\widehat\Theta,\Theta)
  \le\frac23\frac R5+\frac13R=\frac{7R}{15}<\frac{29R}{60}.
\]
This contradiction proves the sample lower bound.
\end{proof}

\begin{proof}[Proof of Lemma~\ref{lem:randomized-group-lower}]
Fix $N_{\cal G}\ge2$ and let $d=\lfloor\log_2N_{\cal G}\rfloor$.
Take $\calX_0=[d]$ with the uniform distribution and
$\calG_0=\{\pm1\}^{[d]}$.  Then $|\calG_0|=2^d\le N_{\cal G}$, and for
every $z\in\R^d$,
\[
  \max_{g\in\calG_0}|\langle g,z\rangle|=\|z\|_1.
\]

For each $\theta\in\{\pm1\}^d$, let $\mathcal D_\theta$ have this uniform
marginal and label means $\mu_i^\theta=1/2+4\eps\theta_i$, and let
$\mathfrak D_0$ consist of all these distributions.  The means lie in
$[1/4,3/4]$ for $\eps\le1/16$.
Fix an output predictor $Q$, and write
$\bar p_i=\E_{P\sim Q_i}P$.  The sign-test identity gives
\[
  \MA_{\mathcal D_\theta}(Q;\calG_0)
  =\frac1d\sum_i|\mu_i^\theta-\bar p_i|.
\]
Decode $\widehat\theta_i=\sgn(\bar p_i-1/2)$, breaking ties arbitrarily.
Each mistake contributes at least $4\eps$ to the sum, so
\[
  \frac1d d_H(\widehat\theta,\theta)
  \le\frac{\MA_{\mathcal D_\theta}(Q;\calG_0)}{4\eps}.
\]
Choose the accuracy constant $c\le 4/5$.  Whenever the learning algorithm
succeeds, this decoder recovers at least $0.8d$ signs.
Lemma~\ref{lem:sign-recovery-information}, with $\eta=4\eps$, now gives
$n=\Omega(d/\eps^2)=\Omega(\log N_{\cal G}/\eps^2)$.
\end{proof}

\begin{proof}[Proof of Lemma~\ref{lem:randomized-calibration-lower}]
Let $a,c_{\rm dec}>0$ be the constants in
Lemma~\ref{lem:randomized-decoding}.  For sufficiently small $\eps$, set
\[
  k=\lfloor\eps^{-1/2}\rfloor\ge2,\qquad \Delta=a/k^2.
\]
Use the hard family $\mathcal D_\theta$, $\theta\in\{0,1\}^k$, defined in
Section~\ref{sec:lower}.  Since $a\le1/4$, all label means lie in
$[1/8,7/8]$.  Choose $c\le c_{\rm dec}$.  On every successful training run,
\[
  \CMA_\theta(Q)\le c\eps\le c_{\rm dec}/k^2,
\]
so Lemma~\ref{lem:randomized-decoding} returns a bit vector correct on at
least $0.8k$ coordinates.  This estimate is a function of the sample and the learning algorithm's
independent seed.  Applying Lemma~\ref{lem:sign-recovery-information} with
$\mathcal B=\{0,1\}$ and $\eta=\Delta$ yields
\[
  n\ge c_{\rm info}\frac{k}{\Delta^2}
  =\frac{c_{\rm info}}{a^2}k^5
  =\Omega(\eps^{-5/2}),
\]
where the last step uses $k\ge\eps^{-1/2}/2$.
\end{proof}

\section{Deferred proofs from the deterministic lower bound}
\label{app:det-lower-proofs}

\begin{proof}[Proof of Lemma~\ref{lem:random-sign-l1}]
Partition $[N]$ into $k=\lceil K_{\rm test}\rceil$ nonempty blocks
$B_1,\ldots,B_k$, each of size at most $s=\lceil N/k\rceil$.
The assumed parameter range, with suitable universal constants, ensures
$1\le k\le N$.  We construct a small family of sign patterns and then allow
each block to flip its sign independently.

First, there is a distribution on sign vectors
$\zeta\in\{\pm1\}^N$, supported on at most
\[
  D_4=\sum_{\ell=0}^{\min\{4,N\}}\binom N\ell=O(N^4)
\]
vectors, whose moments of degree at most four agree with independent
uniform signs.  To see this, associate to each sign vector its vector of
products $\prod_{i\in T}\zeta_i$ with $T\subseteq[N]$ and $|T|\le4$,
including the empty product $1$.  Start with the uniform distribution on
all sign vectors.
If its support has more than $D_4$ points, the corresponding moment
vectors are linearly dependent.  Adjust the probabilities along a
nonzero dependence until one becomes zero.  The constant coordinate
preserves total mass; all probabilities remain nonnegative, and all
moments are unchanged.  Repeating gives the claimed support bound.

For any $v\in\R^N$, put $Z=\sum_i v_i\zeta_i$.  Since $\zeta_i^2=1$,
these moments imply
\[
  \E Z^2=\|v\|_2^2,\qquad
  \E Z^4=3\|v\|_2^4-2\sum_i v_i^4\le3\|v\|_2^4.
\]
Hölder's inequality in the form
$\E Z^2\le(\E|Z|)^{2/3}(\E Z^4)^{1/3}$ therefore gives
\[
  \E|Z|\ge\frac1{\sqrt3}\|v\|_2,
\]
with the zero vector handled trivially.

Let $\mathcal S$ be the support of this distribution.  For
$\zeta\in\mathcal S$ and $\sigma\in\{\pm1\}^k$, include the test
\[
  g_{\zeta,\sigma}(i)=\sigma_b\zeta_i
  \quad\text{for }i\in B_b
\]
in $\calG$.  For a fixed pattern $\zeta$, the block signs can make all
block contributions nonnegative.  Hence, for every $d\in\R^N$,
\begin{align*}
  \max_{g\in\calG}|\langle g,d\rangle|
  &=\max_{\zeta\in\mathcal S}
      \sum_{b=1}^k\left|\sum_{i\in B_b}\zeta_i d_i\right|\\
  &\ge\E_\zeta\sum_{b=1}^k\left|\sum_{i\in B_b}\zeta_i d_i\right|\\
  &\ge\frac1{\sqrt3}\sum_{b=1}^k\|d_{B_b}\|_2
  \ge\frac1{\sqrt{3s}}\|d\|_1
  \ge\sqrt{\frac{k}{6N}}\|d\|_1.
\end{align*}
The last inequality uses $s\le2N/k$.  The family was fixed independently
of $d$, so the guarantee holds simultaneously for every vector.

For any one pattern, the $2^k$ block flips are distinct.  Thus
\[
  2^k\le|\calG|\le D_4\,2^k,
  \qquad
  \log|\calG|=\Theta(K_{\rm test}),
\]
where the upper bound uses $K_{\rm test}\ge A_*\log N$.
This proves both the cardinality and detection guarantees.
\end{proof}

\section{Proof of the finite-test online lemma}
\label{app:online-proof}

\begin{proof}[Proof of Lemma~\ref{lem:online-finite-tests}]
We first describe the online prediction rule.  Let
\[
  \calF=\Phi\times\{-1,1\}.
\]
For $f=(\phi,\sigma)$, define $\psi_f(x,p)=\sigma\phi(x,p)$.  Let
$D=|\calF|=2|\Phi|$.

On round $t$, before observing $Y_t$, maintain the residual sums
\[
  R_{t-1}(f)
  =
  \sum_{s<t}\psi_f(X_s,S_s)(Y_s-S_s),
  \qquad f\in\calF.
\]
Set
\[
  \eta=\min\left\{1,\sqrt{\frac{\log D}{N}}\right\}.
\]
After observing $X_t=x$, put $w_f=\exp(\eta R_{t-1}(f))$ and consider
\[
  H_t(p)=\sum_{f\in\calF}w_f\psi_f(x,p).
\]
This is continuous in $p$.  If $H_t$ has a zero in $[0,1]$, choose its smallest
zero as $S_t$.  If $H_t(p)>0$ for all $p$, choose $S_t=1$.  If $H_t(p)<0$ for
all $p$, choose $S_t=0$.  In all cases, for every label $y\in[0,1]$,
\begin{equation}
  (y-S_t)H_t(S_t)\le 0.
  \label{eq:root-condition}
\end{equation}
This is immediate at a zero.  If $H_t$ is everywhere positive then $S_t=1$,
and if $H_t$ is everywhere negative then $S_t=0$.

Let
\[
  z_t(f)=\psi_f(X_t,S_t)(Y_t-S_t).
\]
Then $|z_t(f)|\le 1$.  For the exponential potential
\[
  W_t=\sum_{f\in\calF}\exp(\eta R_t(f)),
\]
the inequality $e^u\le 1+u+u^2$ for $|u|\le1$ gives
\begin{align*}
  W_t
  &=
  \sum_f\exp(\eta R_{t-1}(f))\exp(\eta z_t(f))  \\
  &\le
  W_{t-1}
  +
  \eta\sum_f w_f z_t(f)
  +
  \eta^2\sum_f w_f z_t(f)^2 \\
  &\le
  W_{t-1}(1+\eta^2),
\end{align*}
because
\[
  \sum_f w_f z_t(f)=(Y_t-S_t)H_t(S_t)\le 0
\]
by \eqref{eq:root-condition}.  Therefore $W_N\le D\exp(\eta^2N)$.  If
$\log D\le N$, every coordinate satisfies
\[
  R_N(f)
  \le
  \frac{\log D}{\eta}+\eta N
  \le
  C\sqrt{N\log D}.
\]
If $\log D>N$, the trivial bound $|R_N(f)|\le N$ gives the same conclusion.
Since both signs of every test are included,
\begin{equation}
  \sup_{\phi\in\Phi}
  \left|\sum_{t=1}^N \phi(X_t,S_t)(Y_t-S_t)\right|
  \le
  C\sqrt{N\log D}.
  \label{eq:online-regret}
\end{equation}

For each history, let $s_t(x)$ be the prediction selected by the same rule
after seeing a context $x$.  Then $S_t=s_t(X_t)$, and
\eqref{eq:online-regret} is the claimed
bound.
\end{proof}

\section{Polynomial-time implementation in the real-arithmetic model}
\label{sec:algorithmic}

The preceding appendix defines and analyzes the online score rules $s_t$.
This section shows that these rules, and
hence the full construction from Theorem~\ref{thm:main}, can be implemented in
polynomial time in an idealized real-arithmetic model.  Throughout this section
we assume that $\calG$ is given
explicitly and that, on input $x$, each value $g(x)$ can be evaluated in unit
time.  Arithmetic on real numbers, including exponentials, is also counted at
unit cost. 

There is one small bookkeeping point.  For implementation it is cleaner to work
with \emph{representations} of piecewise-linear tests rather than distinct
functions.  Let $\widetilde{\calA}_{M,\rho}$ be the set of tuples
\[
  (0=i_0<i_1<\cdots<i_L=N_{\rm grid};\sigma_0,\ldots,\sigma_L),
  \qquad
  1\le L\le M,\quad N_{\rm grid}=1/\rho,\quad \sigma_j\in\{-1,1\},
\]
interpreted as the linear interpolation through the selected grid points
$i_j\rho$ with values $\sigma_j$.  Different tuples may represent the same
function, but this causes no problem: the exponential-potential proof is valid
for a finite multiset of represented tests, and every test in $\calA_{M,\rho}$ has at
least one representation.  Moreover
\[
  |\widetilde{\calA}_{M,\rho}|
  \le
  \sum_{L\le M}\binom{N_{\rm grid}-1}{L-1}2^{L+1}
  \le
  \exp\{C M\log(eN_{\rm grid})\}.
\]

The running time in the next proposition depends polynomially on the fine grid size
$N_{\rm grid}=1/\rho$; with the parameter choice in Theorem~\ref{thm:main}, this is
$N_{\rm grid}=m^2=O(n^{2/5})$.

\begin{proposition}[Unit-cost arithmetic implementation]
\label{prop:efficient-implementation}
Fix $M$ and $\rho$ with $N_{\rm grid}=1/\rho$ an integer.  If $\calG$ is explicit, then
the online intermediate-prediction algorithm from Lemma~\ref{lem:online-finite-tests}, run with the represented
test family $\widetilde{\calA}_{M,\rho}$, can be implemented using
$\poly(N,|\calG|,M,N_{\rm grid})$ arithmetic operations.  Consequently, with the parameter
choice used in Theorem~\ref{thm:main}, the full calibrated multiaccuracy
algorithm can be implemented in time polynomial in $n$ and $|\calG|$ in this
arithmetic model.
\end{proposition}

\begin{proof}
We describe how to implement one online prediction step from the stored
history of earlier rounds.  This suffices, because the batch predictor evaluates
$s_J(x)$ by applying the same rule to one of the stored online states.
The grid-selection step uses the scores already recorded during training.

Let the previous predictions be $S_1,\ldots,S_{t-1}$ and write
$e_s=Y_s-S_s$.  For grid indices $0\le i<j\le N_{\rm grid}$, let
$z_i=i\rho$ and define the half-open interval
\[
  I_{ij}=
  \begin{cases}
    [z_i,z_j), & j<N_{\rm grid},\\
    [z_i,1], & j=N_{\rm grid}.
  \end{cases}
\]
Define two edge residuals
\[
  A_{ij}
  =
  \sum_{s:S_s\in I_{ij}}
  e_s\frac{z_j-S_s}{z_j-z_i},
  \qquad
  B_{ij}
  =
  \sum_{s:S_s\in I_{ij}}
  e_s\frac{S_s-z_i}{z_j-z_i}.
\]
These quantities can be computed for all $i<j$ by direct summation in
$O(tN_{\rm grid}^2)$ arithmetic operations, or faster using sorted prefix sums.  The
direct bound is already polynomial.

Consider a represented one-dimensional test
\[
  a=(0=i_0<i_1<\cdots<i_L=N_{\rm grid};\sigma_0,\ldots,\sigma_L).
\]
With the half-open convention above, each previous prediction belongs to a
unique selected edge.  The cumulative residual of the test $a$ is therefore
the sum of its edge contributions:
\begin{equation}
  R_{t-1}(a)
  =
  \sum_{\ell=1}^L
  \left(
    \sigma_{\ell-1}A_{i_{\ell-1},i_\ell}
    +
    \sigma_\ell B_{i_{\ell-1},i_\ell}
  \right).
  \label{eq:edge-residual-decomposition}
\end{equation}
Thus the exponential weight of a represented test factors along a path in the
directed acyclic graph whose vertices are pairs $(i,\sigma)$ and whose edges go
from $(i,\sigma)$ to $(j,\tau)$ for $i<j$.

Fix $\theta\in\{\eta,-\eta\}$ and set
\[
  T_\theta(i,\sigma,j,\tau)
  =
  \exp\left(
    \theta(\sigma A_{ij}+\tau B_{ij})
  \right).
\]
For a candidate prediction $p$ and an edge
$e=((i,\sigma),(j,\tau))$, define
\[
  c_e(p)=\one\{p\in I_{ij}\}
  \left(\sigma\frac{z_j-p}{z_j-z_i}
       +\tau\frac{p-z_i}{z_j-z_i}\right).
\]
Every represented path contains exactly one edge whose interval contains
$p$, so its test value is the sum of these edge contributions.

A forward recurrence can accumulate both path weights and test values.
Let $W^\theta_\ell(v)$ be the total weight of prefixes with $\ell$ edges
ending at vertex $v$, and let $B^\theta_\ell(v;p)$ be their weighted sum of
accumulated edge contributions at $p$.  Initialize
\[
  W^\theta_0((0,\sigma))=1,\qquad
  B^\theta_0((0,\sigma);p)=0,
\]
and set both quantities to zero at every other vertex.  Writing
$T_\theta(e)=T_\theta(i,\sigma,j,\tau)$, extend the prefixes by
\[
  W^\theta_{\ell+1}(v)
  =\sum_{e:u\to v}T_\theta(e)W^\theta_\ell(u),
\]
\[
  B^\theta_{\ell+1}(v;p)
  =\sum_{e:u\to v}T_\theta(e)
    \bigl(B^\theta_\ell(u;p)+W^\theta_\ell(u)c_e(p)\bigr).
\]
The first recurrence multiplies each prefix weight by its new edge weight;
the second also adds the new edge contribution.  Summing over complete
paths gives
\begin{equation}
  S_\theta(p)
  :=\sum_{a\in\widetilde{\calA}_{M,\rho}}
       \exp(\theta R_{t-1}(a))a(p)
  =\sum_{\ell=1}^M\sum_{\sigma\in\{\pm1\}}
       B^\theta_\ell((N_{\rm grid},\sigma);p).
  \label{eq:weighted-score-dp}
\end{equation}
There are $O(N_{\rm grid})$ vertices and $O(N_{\rm grid}^2)$ edges.
Computing the tables for $\ell\le M$ therefore takes
$O(MN_{\rm grid}^2)$ operations for each candidate $p$.

It remains to choose the online prediction.  The group contribution is explicit:
\[
  G_t(x)
  =
  \sum_{g\in\calG}
  \left(
    \exp(\eta R_{t-1}(g))-\exp(-\eta R_{t-1}(g))
  \right)g(x),
\]
where $R_{t-1}(g)=\sum_{s<t}g(X_s)e_s$ is maintained directly.  The
function used to choose the next prediction for the represented test family is therefore
\[
  H_t(p)=G_t(x)+S_\eta(p)-S_{-\eta}(p).
\]
Every represented one-dimensional test is linear on each fine grid interval
$[z_q,z_{q+1}]$, and hence so is $H_t$.  We evaluate $H_t(z_q)$ for all
$q=0,\ldots,N_{\rm grid}$ using \eqref{eq:weighted-score-dp}.
Zero-valued gridpoints and zeros interpolated between adjacent values of
opposite signs are the candidates; choose the smallest candidate, if any.
If all grid values are
strictly positive, then $H_t$ is positive on all of $[0,1]$ and we choose
$S_t=1$; if all are strictly negative, we choose $S_t=0$.  This exactly
implements the prediction rule in Lemma~\ref{lem:online-finite-tests}.

After $Y_t$ is observed, the algorithm appends $(S_t,e_t)$ to the history and
updates the $|\calG|$ multiaccuracy residuals.  Thus every online step, every later
evaluation of a stored online state, and hence the full intermediate-prediction
subroutine use only polynomially many arithmetic operations in $N$, $|\calG|$,
$M$, and $N_{\rm grid}$.

Finally, choose the grid from the recorded scores $S_1,\ldots,S_N$.
After sorting, prefix sums of $1$, $S_i$, and $S_i^2$ evaluate the empirical
interval contribution in Lemma~\ref{lem:population-quantizer} for every
pair of fine-grid endpoints.  Indeed, its numerator is
$(S_i-a)(b-S_i)=-S_i^2+(a+b)S_i-ab$ for $a<S_i<b$.
There are $O(N_{\rm grid}^2)$ candidate intervals, so grid selection is
also polynomial.
\end{proof}

\end{document}